\pdfoutput=1
\documentclass{article}
\PassOptionsToPackage{numbers, sort&compress}{natbib}
\usepackage[final]{neurips_2022}

\usepackage{wrapfig}
\usepackage[utf8]{inputenc} % allow utf-8 input
\usepackage[T1]{fontenc}    % use 8-bit T1 fonts
\usepackage{url}            % simple URL typesetting
\usepackage{booktabs}       % professional-quality tables
\usepackage{amsfonts}       % blackboard math symbols
\usepackage{nicefrac}       % compact symbols for 1/2, etc.
\usepackage{xcolor}         % colors
\usepackage{graphicx}
\usepackage{bbm}
\usepackage{paralist}
\usepackage{xspace}
\usepackage{bm}
\usepackage{multirow}
\usepackage{amsmath}
\usepackage{subcaption}
\usepackage{makecell}

\usepackage{algorithm}     % algorithm envir
\usepackage{algpseudocode} % algorithm envir

\usepackage{amsthm}

\usepackage{hyperref} 
\definecolor{mydarkblue}{rgb}{0,0.1,0.6}
\hypersetup{
    colorlinks=true,
    citecolor=mydarkblue
          }
          
\def\rvx{{\mathbf{x}}}

\newcommand{\tps}[1]{\overrightarrow{\boldsymbol{#1}}}
\newcommand{\mms}[1]{\{\mskip-5mu\{#1\}\mskip-5mu\}}

\newcommand{\bbmms}[1]{\big\{\mskip-10mu \big\{  #1 \big\} \mskip-10mu \big\}}
\newcommand{\bmms}[1]{\Big\{\mskip-10mu \Big\{  #1 \Big\} \mskip-10mu \Big\}}
\newcommand{\Bmms}[1]{\bigg\{\mskip-10mu \bigg\{  #1 \bigg\} \mskip-10mu \bigg\}}
\newcommand{\ms}[1]{\tilde{\boldsymbol{#1}}}
\newcommand{\s}[1]{\hat{\boldsymbol{#1}}}

\newcommand{\cbit}{\begin{compactitem}}
\newcommand{\ceit}{\end{compactitem}}
\newcommand{\cben}{\begin{compactenum}}
\newcommand{\ceen}{\end{compactenum}}

\newtheorem{lemma}{Lemma}
\newtheorem{theorem}{Theorem}

\newcommand{\amethod}{{\sc $(\leq)$}\xspace}

\newcommand{\method}{{\sc $(k,c)(\leq)$-SetGNN}\xspace}
\newcommand{\methods}{{\sc $(k,c)(\leq)$-SetGNN$^*$}\xspace}

\newcommand{\kcswl}{{\sc $(k,c)(\leq)$-SetWL}\xspace}
\newcommand{\kswl}{{\sc $k(\leq)$-SetWL}\xspace}
\newcommand{\ksfwl}{{\sc $k(\leq)$-SetFWL}\xspace}
\newcommand{\kmwl}{{\sc $k$-MultisetWL}\xspace}

\makeatletter
\newcommand\footnoteref[1]{\protected@xdef\@thefnmark{\ref{#1}}\@footnotemark}
\makeatother

\newcommand{\beq}{\begin{equation}}
	\newcommand{\eeq}{\end{equation}}

\newcommand{\bit}{\begin{itemize}}
	\newcommand{\eit}{\end{itemize}}
\newcommand{\ben}{\begin{enumerate}}
	\newcommand{\een}{\end{enumerate}}

\newcounter{x}

\newcommand{\calK}{\mathcal{K}}
\newcommand{\calX}{\mathcal{X}}

\newcommand{\R}{\mathbb{R}}

\usepackage{url}            % simple URL typesetting
\usepackage{xcolor}
\usepackage{titlesec}
\titlespacing*{\subsection}{0pt}{1pt}{1pt}
\titlespacing*{\section}{0pt}{2pt}{2pt}
\newtheorem{corollary}{Corollary}[theorem]

\newtheorem{innercustomthm}{Theorem}
\newenvironment{customthm}[1]
  {\renewcommand\theinnercustomthm{#1}\innercustomthm}
  {\endinnercustomthm}

\title{A Practical, Progressively-Expressive GNN}

\author{%
  Lingxiao Zhao\\
  Carnegie Mellon University\\
  \texttt{lingxiao@cmu.edu} \\
  \And
 Louis Härtel\\
 RWTH Aachen University\\
 \texttt{haertel@informatik.rwth-aachen.de}\\
  \And
  Neil Shah\\
  Snap Inc.\\
  \texttt{nshah@snap.com } \\
  \And
  Leman Akoglu\\
  Carnegie Mellon University\\
  \texttt{lakoglu@andrew.cmu.edu} 
}

\begin{document}

\maketitle

\begin{abstract}
    Message passing neural networks (MPNNs) have become a dominant flavor of graph neural networks (GNNs) in recent years. Yet, MPNNs come with notable limitations; namely, they are at most as powerful as the 1-dimensional Weisfeiler-Leman (1-WL) test in distinguishing graphs in a graph isomorphism %(GI)
    testing framework. To this end, researchers have drawn inspiration from the $k$-WL hierarchy to develop more expressive GNNs.  However, current $k$-WL-equivalent GNNs are not practical for even small values of $k$, as $k$-WL becomes combinatorially more complex as $k$ grows. At the same time, several works have found great empirical success in graph learning tasks without highly expressive models, implying that chasing expressiveness with a \textit{``coarse-grained ruler''} of expressivity like $k$-WL is often unneeded in practical tasks.  To truly understand the expressiveness-complexity tradeoff, one desires a more \textit{``fine-grained ruler,''} which can more gradually increase expressiveness.  Our work puts forth such a proposal: Namely, we first propose the \kcswl hierarchy with greatly reduced complexity from $k$-WL, achieved by moving from $k$-tuples of nodes to sets with ${\leq}k$ nodes defined over ${\leq}c$ connected components in the induced original graph.  We show favorable theoretical results for this model in relation to $k$-WL, and concretize it via \method, which is as expressive as \kcswl.  Our model is \textit{practical} and \textit{progressively-expressive}, increasing in power with $k$ and $c$.  We demonstrate effectiveness on several benchmark datasets, achieving several state-of-the-art results with runtime and memory usage applicable to practical graphs. We open source our implementation at \url{https://github.com/LingxiaoShawn/KCSetGNN}.
    %showing \ns{?}.
\end{abstract}

\section{Introduction}

% Graphs, GNNs and advantages
In recent years, graph neural networks (GNNs) have gained considerable attention \cite{wu2020graph, wu2020comprehensive} for their ability to tackle various node-level \cite{kipf2016semi, velivckovic2017graph}, link-level \cite{zhang2020revisiting, sankar2021graph} and graph-level \cite{bodnar2021weisfeiler, maron2019provably} learning tasks, given their ability to learn rich representations for complex graph-structured data.  The common template for designing GNNs follows the  {message passing} paradigm; these so-called message-passing neural networks (MPNNs) are built by stacking layers which encompass feature transformation and aggregation operations over the input graph \cite{ma2021unified, gilmer2017neural}.  
% GNNs: shortcomings
Despite their advantages, MPNNs have several limitations including oversmoothing \cite{zhao2020pairnorm, chen2020measuring, oono2019graph},
oversquashing \cite{alon2021on, topping2021understanding}, inability to distinguish node identities \cite{you2021identity} and positions \cite{you2019position}, and \textit{expressive power} \citep{Xu:2019ty}.% -- the last of which is the focus of our work.

%Graphs provide a powerful abstraction for many relational phenomena in the real world, and therefore have been used to tackle various graph based learning tasks, such as \la{example with citations}. With the recent advances in deep learning, graph neural networks (GNNs) have gained a lot of attention \la{cite surveys\cite{}}, owing their popularity to a variety of factors including latent representation learning, end-to-end training, being able to readily admit complex (e.g., node-attributed, directed) graphs, to name a few.

% GNN shortcomings, specifically expressiveness
Since Xu et al.'s \cite{Xu:2019ty} seminal work showed that MPNNs are at most as powerful as the first-order Weisfeiler-Leman (1-WL) test in the graph isomorphism (GI) testing framework, there have been several follow-up works on improving the understanding of GNN expressiveness \cite{arvind2020weisfeiler, chen2020can}. %For example, several works showed that 1-WL equivalent GNNs are not able to capture graph properties such as
%cycles or triangle counts \cite{arvind2020weisfeiler,chen2020can}.
In response, the community proposed many GNN models to overcome such limitations \cite{journals/corr/abs-2003-04078,azizian2020expressive, zhao2021stars}.  Several of these 
%Furthermore, since \cite{chen2019equivalence} showed the equivalence between graph isomorphism testing and universal permutation-invariant function approximation, many 
 aim to design powerful higher-order GNNs which are increasingly expressive \cite{morris2019weisfeiler, maron2019universality, keriven2019universal, bodnar2021weisfeiler} by inspiration from the $k$-WL hierarchy \cite{shervashidze2011weisfeiler}.

% As the equivalence has been shown
% between WL tests and universal permutation invariant function approximation \cite{chen2019equivalence},
% many recent work are inspired by the $k$-WL hierarchy \la{old citation?} in their design toward more powerful GNNs 
% \cite{morris2019weisfeiler,keriven2019universal,maron2019provably,balcilar2021breaking}
% \la{cite k-WL inspired representative powerful recent GNNs:?}

% LIMITATION: expressive models do not scale.
% ALSO: do we really need them?
A key limitation of such higher-order models that reach beyond 1-WL is their poor scalability; in fact, these models can only be applied to small graphs with small $k$ in practice \cite{maron2019provably, maron2019universality, morris2019weisfeiler} due to combinatorial growth in complexity.
On the other hand lies an open question on whether practical graph learning tasks indeed need such complex, and extremely expressive GNN models.
Historically, Babai et al. \cite{babai1980random} showed that almost all graphs on $n$ nodes
%(i.e., $O(2^{n \choose 2})$ graphs) 
can be distinguished by 1-WL.
%, which challenges  the growing trend toward highly expressive models. 
In other contexts like node classification, researchers have encountered superior generalization performance with graph-augmented multi-layer perceptron (GA-MLP) models \cite{rossi2020sign, klicpera2018predict} compared to MPNNs, despite the former being strictly less expressive than the latter \cite{chen2020graph}.  %Also 
Considering that increased model complexity has negative implications in terms of overfitting and generalization \cite{hawkins2004problem}, %one may justifiably
it is worth re-evaluating continued efforts to pursue maximally expressive GNNs in practice. Ideally, one could study these models' generalization performance on various real-world tasks by increasing $k$ (expressiveness) in the $k$-WL hierarchy.
%Unfortunately, current $k$-WL equivalent GNNs, such as $k$-IGNs \cite{maron2019provably}, are not implementable beyond $k>3$ due to combinatorial \la{?} growth in computational complexity. 
Yet, given the impracticality of this too-coarse ``ruler'' of expressiveness, which becomes infeasible beyond $k$$>$$3$, one desires a more fine-grained ``ruler'', that is, a new hierarchy whose expressiveness grows more gradually. Such a hierarchy could enable us to gradually build more expressive models which admit improved scaling, and avoid unnecessary leaps in model complexity for tasks which do not require them, guarding against overfitting.

\begin{figure}[t!]
	\vspace{-0.15in}
	\centering
	\includegraphics[width=0.99\textwidth]{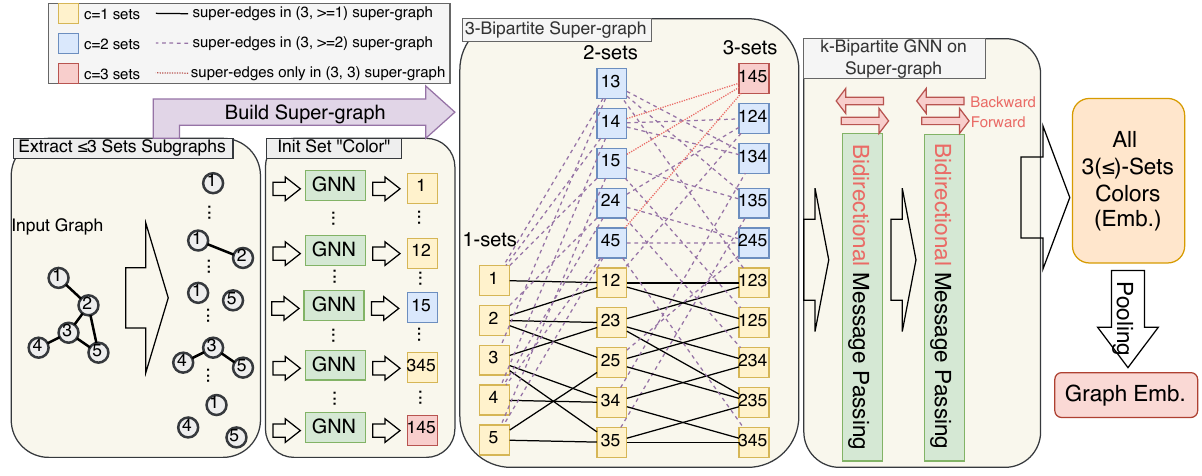}
	\vspace{-0.1in}
	\caption{Main steps of \method. 
	Given a graph and ($k,c$), we build the $(k,c)$-bipartite super-graph (in middle) containing sets with up to $k$ nodes and $c$ connected components in the induced subgraph, on which  a base GNN assigns initial ``colors''. Bidirectional bipartite GNN layers with frw.-bckw. message passing learn set embeddings, pooled into graph embedding. The size of super-graph, and accordingly its expressiveness,  grows progressively with increasing $k$ and $c$. The $2$-bipartite message passing generalizes normal GNN,   edge GNN, and line graph (see Appendix.\ref{apdx:k_bipartite_discussion}).  }
	 \label{fig:chart}
	 \vspace{-0.3in}
\end{figure}

% This paper:
\textbf{Present Work.} In this paper, we propose such a hierarchy, and an associated \textit{practical progressively-expressive GNN model}, called \method, %\la{name may be hard to refer by others}, 
whose expressiveness can be modulated through $k$ and $c$.  In a nutshell, we take inspiration from $k$-WL, yet achieve practicality through three design ideas which simplify key bottlenecks of scaling $k$-WL: \textbf{First}, we move away from $k$-tuples of the original $k$-WL 
%(i.e., ordered multisets with $k$ nodes) 
to $k$-multisets (unordered), and then to $k(\leq)$-sets. %(i.e., unordered sets with $\leq k$ nodes)
 We demonstrate that these steps drastically reduce the number nodes in the $k$-WL graph while retaining significant expressiveness. (Sec.s \ref{ssec:remove_ordering}$-$\ref{ssec:remove_repetition})
\textbf{Second}, by considering the underlying sparsity of an input graph, %which are often sparse \la{citation}, 
we reduce scope to $k, c(\leq)$-sets that consist of ${\leq}k$ nodes whose induced subgraph is comprised of ${\leq}{c}{\leq}{k}$ connected components. 
This also yields massive reduction in the number of nodes while improving practicality; i.e. small values of $c$ on sparse graphs can allow one to increase $k$ well beyond $3$ in practice. (Sec. \ref{ssec:consider_sparsity})
%We also theoretically show that \method's expressiveness grows progressively by increasing $k$ and $c$, enabling its use as an expressiveness hierarchy.
\textbf{Third}, we design a super-graph architecture that consists of a sequence of $k$$-$$1$ \textit{bipartite} graphs
over which we learn embeddings for our $k, c(\leq)$-sets, using bidirectional message passing.  % (i.e., induced subgraphs).  %\la{abstract does not say anything about $k$-bi, or fast ``colors'' -- elaborate here?} \ns{leaving for LZ as i also don't know} 
These embeddings can be pooled to yield a final graph embedding.. (Sec.s \ref{ssec:to_gnn}$-$\ref{ssec:bidrectional}) We also speed up initializing ``colors'' for the $k, c(\leq)$-sets, for $c>1$, substantially reducing computation and memory. (Sec. \ref{ssec:supernode_init}) Fig. \ref{fig:chart} overviews of our proposed framework. Experimentally,  our \method outperforms existing state-of-the-art GNNs on simulated expressiveness as well as real-world graph-level tasks, achieving new bests on substructure counting and ZINC-12K, respectively. We show that generalization performance reflects increasing expressiveness by $k$ and $c$.
Our proposed scalable designs allow us to
train models with e.g. $k$$=$$10$ or $c$$=$$3$ with practical running time and memory requirements.   
% Our code and datasets are open-sourced at
% \url{https://anonymous.4open.science/r/KCSetGNN}.

%\la{include/scatter formal sentences about what else we are able to show/state.} \ns{some things which would be nice to show (but not sure if we can say/show them all): (a) improves expressiveness by increasing k and c, (b) is faster than k-WL, (c) can be as expressive as X (something the rest of the community can understand, and (d) is accurate and practically useful for larger graphs.}

%Collectively, \la{what to say about the savings? pick from complexity analysis.}

%%%%We  

%  tuple -> multiset -> set -> bigraph tweaks "designs for scalability."   If he wants to write the wishlist version and fill in later on after appendix, then it is even easier to write these tweaks as no-brainers, if nothing gets sacrificed at all from k-WL.  But it is the difference between saying we make a practical version of k-WL and we make a practically more expressive GNN (by deviating from k-WL but only taking some inspiration from it)  

%\la{i probably missed various essential citations above. Please scatter them as fit.}

\section{Related Work}

\textbf{MPNN limitations.} Given the understanding that MPNNs have expressiveness upper-bounded by 1-WL \cite{Xu:2019ty}, several researchers investigated what else MPNNs \textit{cannot} learn.  To this end, Chen et al. \cite{chen2020can} showed that MPNNs cannot count induced connected substructures of 3 or more nodes, while along with \cite{chen2020can}, Arvind et al. \cite{arvind2020weisfeiler} showed that MPNNs can only count star-shaped patterns.  Loukas \cite{loukas2019graph} further proved several results regarding decision problems on graphs (e.g. subgraph verification, cycle detection), finding that MPNNs cannot solve these problems unless strict conditions of depth and width are met. Moreover, Garg et al. \cite{garg2020generalization} showed that many standard MPNNs cannot compute properties such as longest or shortest cycles, diameters or existence of cliques.

\textbf{Improving expressivity.} Several works aim to improve  expressiveness limitations of MPNNs.  One approach is to inject features into the MPNN aggregation, motivated by Loukas \cite{loukas2019graph} who showed that MPNNs can be universal approximators when nodes are sufficiently distinguishable.  Sato et al.
\cite{sato2021random} show that injecting random features can better enable MPNNs to solve problems like minimum dominating set and maximum matching.  You et al. \cite{you2021identity} inject cycle counts as node features, motivated by the limitation of MPNNs not being able to count cycles.  Others proposed utilizing subgraph isomorphism counting to empower MPNNs with substructure information they cannot learn \cite{bouritsas2022improving, bodnar2021weisfeiler}.  Earlier work by You et al. \cite{you2019position} adds positional features to distinguish node which na\"ive MPNN embeddings would not.  Recently, several works also propose utilizing subgraph aggregation; Zhang et al. \cite{zhang2021nested}, Zhao et al. \cite{zhao2021stars} and Bevilacqua et al.  \cite{bevilacqua2021equivariant} propose subgraph variants of the WL test which are no less powerful than 3-WL. Recently a following up work \cite{frasca2022understanding} shows that rooted subgraph based extension with 1-WL as kernel is bounded by 3-WL.  The community is yet exploring several distinct avenues in overcoming limitations:  Murphy et al. \cite{murphy2019relational} propose a relational pooling mechanism which sums over all permutations of a permutation-sensitive function to achieve above-1-WL expressiveness.  Balcilar et al. \cite{balcilar2021breaking} generalizes spatial and spectral MPNNs and shows that instances of spectral MPNNs are more powerful than 1-WL. Azizian et al. \cite{azizian2020expressive} and Geerts et al. \cite{geerts2022expressiveness} unify expressivity and approximation ability results for existing GNNs.

\textbf{$k$-WL-inspired GNNs.} %(Mainly Morris)
% go sparse paper
The $k$-WL test captures higher-order interactions in graph data by considering all $k$-tuples, i.e., size $k$ ordered multisets defined over the set of nodes. While highly expressive, it does not scale to practical graphs beyond a very small $k$ ($k=3$ pushes the practical limit even for small graphs). However, several works propose designs which can achieve $k$-WL in theory and strive to make them practical.  Maron et al. \cite{maron2019universality} proposes a general class of invariant graph networks $k$-IGNs having exact $k$-WL expressivity \citep{geerts2020expressive}, while being not scalable. 
Morris et al. have several work on $k$-WL and its variants like $k$-LWL \cite{morris2017glocalized}, $k$-GNN \cite{morris2019weisfeiler}, and $\delta$-$k$-WL-GNN \cite{morris2020weisfeiler}. Both $k$-LWL and $k$-GNN use a variant of $k$-WL that only considers $k$-sets and are strictly weaker than $k$-WL but much more practical. In our paper we claim that $k(\leq)$-sets should be used with additional designs to keep the best expressivity while remaining practical. The $\delta$-$k$-WL-GNN works with $k$-tuples but sparsifies the connections among tuples by considering locality. Qian et al. \cite{qian2022ordered} extends the k-tuple to subgraph network but has same scalability bottleneck as k-WL.  
A recent concurrent work SpeqNet \cite{morris2022speqnets} proposes to reduce number of tuples by restricting the number of connected components of tuples' induced subgraphs. The idea is independently explored in our paper in Sec.\ref{ssec:consider_sparsity}. 
All four variants proposed by Morris et al. do not have realizations beyond $k >3$ in their experiments while we manage to reach $k=10$. Interestingly, a recent work \cite{kim2021transformers} links graph transformer with $k$-IGN that having better (or equal) expressivity, however is still not scalable. 

%Additional 

% \section{$k$-WL equivalent isomorphism test: $k(\leq)$-SetWL}
% \section{A practical isomorphism test: $(k,c)(\leq)$-SetWL}

\section{A practical progressively-expressive isomorphism test: \kcswl}

We first motivate our method \method from the GI testing perspective by introducing \kcswl. 
%To achieve progressively-expressive design with being highly expressive and practical, we start from $k$-WL algorithm and practicalize it greatly via simplifications. 
We first introduce notation and background of $k$-WL (Sec. \ref{ssec:preliminaries}). Next, we show how to increase the practicality of $k$-WL without reducing much of its expressivity, by removing node ordering (Sec. \ref{ssec:remove_ordering}), node repetitions (Sec. \ref{ssec:remove_repetition}), and leveraging graph sparsity (Sec. \ref{ssec:consider_sparsity}). We then give the complexity analysis (Sec. \ref{ssec:complexity_analysis}). We close the section with extending the idea to $k$-FWL which is as expressive as $(k$+$1)$-WL (Sec. \ref{ssec:fwl}).  All proofs are included in Appendix \ref{apdx:proofs}.  

\textbf{Notation:} Let $G=(V(G), E(G), l_G)$ be an undirected,
%(edge being symmetric is necessary) 
colored graph with nodes $V(G)$, edges $E(G)$, and a color-labeling function $l_G: V(G)\rightarrow C$ where $C=\{c_1,...,c_d\}$ denotes a set of $d$ distinct colors. Let $[n]=\{1,2,3,...,n\}$.
Let $(\cdot)$ denote a \textit{tuple} (ordered multiset), $\mms{\cdot}$ denote a \textit{multiset} (set which allows repetition), and $\{\cdot\}$ denote a  \textit{set}.  
We define $\tps{v}=(v_1,...,v_k)$ as a $k$-tuple, $\ms{v}= \mms{v_1,...,v_k}$ as a $k$-multiset, and $\s{v}= \{v_1,...,v_k\}$ as a $k$-set. Let $\tps{v}[x/i] = (v_1,...,v_{i-1},x,v_{i+1}, ..., v_k)$ denote replacing the $i$-th element in $\tps{v}$ with $x$, and analogously for $\ms{v}[x/i]$ and $\s{v}[x/i]$ (assume mutlisets and sets are represented with the canonical ordering). When $v_i \in V(G)$, let $G[\tps{v}]$, $G[\ms{v}]$, $G[\s{v}]$ denote the induced subgraph on $G$ with nodes inside $\tps{v}, \ms{v},\s{v}$ respectively (keeping repetitions). An isomorphism between two graphs $G$ and $G'$ is a bijective mapping $p:V(G)\rightarrow V(G')$ which satisfies $\forall u,v \in V(G), (u,v)\in E(G) \Longleftrightarrow (p(u),p(v)) \in E(G')$ and $\forall u \in V(G), l_G(u) = l_{G'}(p(u))$. Two graphs $G, G'$ are isomorphic if there exists an isomorphism between them, which we denote as $G \cong G'$. 

% To measure expressivity, we define the separation power of graph functions based on  \citep{geerts2022expressiveness}, which measures the ability to separate node sets. 

% \begin{definition} [Separation Power] Let $\mathcal{G}$ denotes all undirected graphs, $\mathcal{V}$ denotes the set of all combinations of nodes.  Let $\mathcal{F}$ be a set of functions with domain on $(\mathcal{G}, \mathcal{V})$. Define equivalence relation $\rho(\mathcal{F})$ as:
% $((G, \bm{v}) , (G',\bm{v}')) \in \rho(\cal{F})$ $ \Longleftrightarrow \forall f \in \mathcal{F}, f(G, \bm{v}) = f(G', \bm{v}') $.
% \end{definition}

\subsection{Preliminaries: the $k$-Weisfeiler-Leman ($k$-WL) Graph Isomorphism Test}
\label{ssec:preliminaries}
%The GI problem has been investigated for many decades and still draws attention due to its relationship to the concept of NP-completeness. GI is neither known to be solvable in polynomial time, nor to be NP-complete, and is thus called NP-intermediate. Many algorithms have been developed to partially solve the GI problem; 
The 1-dimensional Weisfeiler-Leman ($1$-WL) test, also known as color refinement \cite{read1977graph} algorithm, is a widely celebrated approach to (approximately) test GI. Although extremely fast and effective for most graphs ($1$-WL can provide canonical forms for all but $n^{-1/7}$ fraction of $n$-vertex graphs \cite{babai1980random}), it fails to distinguish members of many important graph families, such as regular graphs. The more powerful $k$-WL algorithm first appeared in \cite{weisfeiler1976construction}, and  extends coloring of vertices (or 1-tuples) to that of $k$-tuples. $k$-WL is progressively expressive with increasing $k$, and can distinguish any finite set of graphs given a sufficiently large $k$. $k$-WL has many interesting connections to logic, games, and linear equations \cite{cai1992optimal,grohe2015pebble}. Another variant of $k$-WL is called $k$-FWL (Folklore WL), such that $(k+1)$-WL is equally expressive as $k$-FWL \cite{cai1992optimal, grohe2021logic}. Our work focuses on $k$-WL. % instead of $k$-FWL. 

$k$-WL iteratively recolors all $n^k$ $k$-tuples defined on a graph $G$ with $n$ nodes. At iteration 0, each $k$-tuple $\tps{v}=(v_1,...,v_k) \in V(G)^k$ is initialized with a color as its \textit{atomic type} $\bm{at_k}(G, \tps{v})$. Assume $G$ has $d$ colors, then $\bm{at_k}(G, \tps{v}) \in \{0,1\}^{2\binom{k}{2} + kd}$ is an ordered vector encoding. The first $\binom{k}{2}$ entries indicate whether $v_i = v_j$,  $\forall i,j, 1{\leq}i{<}j{\leq}k$, which is the node repetition information. The second $\binom{k}{2}$ entries indicate whether $(v_i, v_j) \in E(G)$. The last $kd$ entries one-hot encode the initial color of $v_i$, $\forall i, 1\leq i  \leq k$. Importantly,  $\bm{at_k}(G, \tps{v}) =  \bm{at_k}(G', \tps{v'})$ if and only if $v_i \mapsto v'_i$ is an isomorphism from $G[\tps{v}]$ to $G'[\tps{v'}]$.
Let $\bm{wl}_k^{(t)}(G, \tps{v})$ denote the color of $k$-tuple $\tps{v}$ on graph $G$ at $t$-th iteration of $k$-WL, where colors are initialized with $\bm{wl}_k^{(0)}(G, \tps{v})= \bm{at_k}(G, \tps{v})$. 

At the $t$-th iteration, $k$-WL updates the color of each $\tps{v} \in V(G)^k$ according to
% {\footnotesize
\begin{align}
 \scriptstyle
    \bm{wl}_k^{(t+1)}(G, \tps{v}) = \text{HASH} \Big( 
    \bm{wl}_k^{(t)}(G, \tps{v}),  
     \bmms{\bm{wl}_k^{(t)}(G, \tps{v}[x/1]) \Big| x \in V(G)}, \ldots, \bmms{\bm{wl}_k^{(t)}(G, \tps{v}[x/k]) \Big| x \in V(G)}
    \Big) \label{eq:kwl}  
\end{align}
% }
\vspace{-0.1in}

Let $\bm{gwl}_k^{(t)}(G)$ denote the encoding of $G$ at $t$-th iteration of $k$-WL.  Then,
{\small
\begin{align}
    \bm{gwl}_k^{(t)}(G) = \text{HASH}\bigg( \Bmms{ \bm{wl}_k^{(t)}(G, \tps{v}) \Big | \tps{v}\in V(G)^k}  \bigg) \label{eq:kgwl}
\end{align}
}
\vspace{-0.1in}

Two $k$-tuples $\tps{v}$ and $\tps{u}$ are connected by an edge if $|\{i\in[k]|v_i = u_i\}|={k-1}$, or informally if they share $(k$$-$$1)$ entries. 
Then, $k$-WL defines a super-graph $S_{k\text{-wl}}(G)$ with its nodes being all $k$-tuples in $G$, and edges defined as above.
Eq. \eqref{eq:kwl} defines the rule of color refinement on $S_{k\text{-wl}}(G)$. Intuitively, $k$-WL is akin to (but more powerful as it orders subgroups of neighbors)  running 1-WL algorithm on the supergraph $S_{k\text{-wl}}(G)$. 
% Then eq.\ref{eq:kwl} shows that at each iteration $\tps{v}$'s new color is determined by its own color and its neighbors' colors. 
As $t\rightarrow \infty$, the color $\bm{wl}_k^{(t+1)}(G, \tps{v})$  converges to a stable value, denoted as $\bm{wl}_k^{(\infty)}(G, \tps{v})$ and the corresponding stable graph color denoted as  $\bm{gwl}_k^{(\infty)}(G)$. For two non-isomorphic graphs $G, G'$, $k$-WL can successfully distinguish them if $\bm{gwl}_k^{(\infty)}(G) \not =$ $\bm{gwl}_k^{(\infty)}(G')$. The expressivity of $\bm{wl}_k^{(t)}$ can be exactly characterized by first-order logic with counting quantifiers. Let $\mathbf{C}_k^t$ denote all first-order formulas with at most $k$ variables and $t$-depth counting quantifiers, then $\bm{wl}^{(t)}_k(G, \tps{v}) = \bm{wl}^{(t)}_k(G', \tps{v'})$ $\Longleftrightarrow$ $\forall \phi \in \mathbf{C}^t_k , \phi(G, \tps{v}) = \phi(G', \tps{v'})$. Additionally, there is a $t$-step bijective pebble game that are equivalent to $t$-iteration $k$-WL in expressivity. See Appendix.\ref{apdx:wl-pebble} for the pebble game characterization of $k$-WL. 

% $\rho(wl_k^{(t)}) = \mathbf{C}_k^t$ \citep{cai1992optimal}. 
% Hence the expressivity of $wl_k^{(t)}$ is progressively increased by $k$ and $t$, based on the property of $\mathbf{C}_k^t$.

Despite its power, $k$-WL uses all $n^k$ tuples and has $O(kn^k)$ complexity at each iteration.
%While being extremely powerful, $k$-WL uses all $n^k$ tuples and has $O(kn^k)$ complexity at each iteration, which disallows it from being practically used for even small $k>3$. We next describe several steps that we undertake to greatly simplify and practicalize $k$-WL.

%To greatly reduce its complexity, we take the direction of reducing number of tuples, we propose to change $k$-tuples to $k(\leq)$-sets (sets with $\leq k$ elements), and we conjecture that it doesn't lose any expressivity in terms of separating set of nodes. To further reduce number of $k(\leq)$-sets, we further propose to only consider a $k(\leq)$-set $\s{v}$ if its induced subgraph $G[\s{v}]$ has number of connected components $\leq c$.

\subsection{From $k$-WL to \kmwl: Removing Ordering}
\label{ssec:remove_ordering}

Our first proposed adaptation to $k$-WL is to \textit{remove ordering} in each $k$-tuple, i.e. changing $k$-tuples to $k$-multisets.  This greatly reduces the number of supernodes to consider 
by $O(k!)$ times. 

Let $\ms{v}=\mms{v_1,...,v_k}$ be the corresponding multiset of tuple $\tps{v}=(v_1,...,v_k)$. We introduce a canonical order function on $G$,  $o_G:V(G)\rightarrow [n]$, and a corresponding indexing function $o^{-1}_G(\ms{v},i)$, which returns the $i$-th element of $v_1,...,v_k$ sorted according to $o_G$. Let $\bm{mwl}_k^{(t)}(G, \ms{v})$ denote the color of the $k$-multiset $\ms{v}$ at $t$-th iteration of \kmwl, formally defined next.

At $t=0$, we initialize the color of $\ms{v}$ as $\bm{mwl}_k^{(0)}(G, \ms{v}) =\text{HASH}\big( \mms{\bm{at}_k(G, p(\ms{v})) | p \in \text{perm[k]} } \big)$ where $\text{perm[k]}$\footnote{This function should also consider repeated elements; we omit this for clarity of presentation.} denotes the set of all permutation mappings of $k$ elements. It can be shown that $\bm{mwl}_k^{(0)}(G, \ms{v}) = \bm{mwl}_k^{(0)}(G', \ms{v}')$ if and only if $G[\ms{v}]$ and $G'[\ms{v}']$ are isomorphic. 

At $t$-th iteration, \kmwl updates the color of every $k$-multiset by 
{\small\begin{align}
% \scriptstyle
    \bm{mwl}_k^{(t+1)} (G, \ms{v}) = \text{HASH} \bigg( 
    \bm{mwl}_k^{(t)}(G, \ms{v}),   \bigg\{\mskip-10mu \bigg\{  \label{eq:kmwl}
    & \mms{\bm{mwl}_k^{(t)}(G, \ms{v}[x/o^{-1}_G(\ms{v},1)]) \big| x \in V(G)}, \\..., \nonumber
    & \mms{\bm{mwl}_k^{(t)}(G, \ms{v}[x/o^{-1}_G(\ms{v},k)]) \big| x \in V(G)}
    \bigg\}\mskip-10mu \bigg\}
    \bigg)   
\end{align}}
Where $\ms{v}[x/o^{-1}_G(\ms{v},i)])$ denotes replacing the $i$-th (ordered by $o_G$) element of the multiset with $x\in V(G)$. Let $S_{k\text{-mwl}}(G)$ denote the super-graph defined by \kmwl.
Similar to Eq. \eqref{eq:kgwl}, the graph level encoding is
$\bm{gmwl}_k^{(t)}(G) = \text{HASH}( \mms{ \bm{mwl}_k^{(t)}\big(G, \ms{v}\big) \big|\forall \ms{v}\in V(S_{k\text{-mwl}}(G))   } \big)$. 

Interestingly, although \kmwl has significantly fewer number of node groups  than $k$-WL, we show it is no less powerful than $k$-$1$-WL in terms of distinguishing graphs, while being upper bounded by $k$-WL in distingushing both node groups and graphs.

\begin{theorem}\label{thm:kwl=kmwl} 
 Let $k\ge 1$ and $\bm{wl}_k^{(t)}(G, \ms{v}) := \mms{\bm{wl}_k^{(t)}(G, p(\ms{v})) | p \in \text{perm[k]}}$.
 For all $t\in \mathbb{N}$ and all graphs $G, G'$:
 %\cbit
%\item[(1)] 
 \kmwl is upper bounded by $k$-WL in distinguishing multisets $G, \ms{v}$ and $G', \ms{v}'$ at $t$-th iteration, i.e.
$\bm{wl}_k^{(t)}(G, \ms{v}) = \bm{wl}_k^{(t)}(G', \ms{v}')$  $\Longrightarrow$ $\bm{mwl}_k^{(t)}(G, \ms{v}) = \bm{mwl}_k^{(t)}(G', \ms{v}')$. 
%\item[(2)] $k$-WL and \kmwl are equivalent in distinguishing graphs $G$ and $G'$ at $t$-th iteration, i.e. $\bm{gwl}_k^{(t)}(G) = \bm{gwl}_k^{(t)}(G') \Longleftrightarrow$  $\bm{gmwl}_k^{(t)}(G) = \bm{gmwl}_k^{(t)}(G')$ \lz{not  sure}.  
%\ceit
\end{theorem}

\begin{theorem}\label{thm:cfi-mwl}
    \kmwl is no less powerful than  ($k$-$1$)-WL in distinguishing graphs: for any $k\ge 3$ there exists graphs that can
    be distinguished by \kmwl but not by ($k$-$1$)-WL.
\end{theorem}

Theorem \ref{thm:cfi-mwl} is proved by using a variant of a series of CFI \cite{cai1992optimal} graphs which cannot be distinguished by $k$-WL. This theorem shows that \kmwl is indeed very powerful and finding counter examples of $k$-WL distinguishable graphs that cannot be distinguished by \kmwl is very hard. Hence we conjecture that \kmwl may have the same expressivity as $k$-WL in distinguishing undirected graphs, with an attempt of proving it in Appendix.\ref{apdx:conjecture}. Additionally, the theorem also implies that \kmwl is strictly more powerful than {{\sc ($k-1$)-MultisetWL}\xspace}.

We next give a pebble game characterization of \kmwl, which is named doubly bijective $k$-pebble game presented in Appendix.\ref{apdx:mwl-pebble}. The game is used in the proof of Theorem \ref{thm:cfi-mwl}. 

\begin{theorem}\label{dbp_mwl}
 \kmwl has the same expressivity as the doubly bijective $k$-pebble game.
\end{theorem}
 
\subsection{From \kmwl to \kswl: Removing Repetition}
\label{ssec:remove_repetition}
Next, we propose further \textit{removing repetition} inside any $k$-multiset, i.e. transforming $k$-multisets to $k(\leq)$-sets. We assume elements of $k$-multiset $\ms{v}$ and $k(\leq)$-set $\s{v}$ in $G$ are sorted based on the ordering function $o_G$, and omit $o_G$ for clarity. Let $s(\cdot)$ transform a multiset to set by removing repeats, and let $r(\cdot)$ return a tuple with the number of repeats for each distinct element in a multiset. Specifically, let $\s{v} =  s(\ms{v})$ and $\s{n} =  r(\ms{v})$, then $m:=|\s{v}|=|\s{n}| \in [k]$ denotes the number of distinct elements in $k$-multiset $\ms{v}$, and $\forall i \in [m]$, $\s{v}_i$ is the $i$-th distinct element with $\s{n}_i$ repetitions. Clearly there is an injective mapping between $\ms{v}$ and $(\s{v}, \s{n} )$; let $f$ be the inverse mapping such that $\ms{v} = f(s(\ms{v}), r(\ms{v}))$. Equivalently, each $m$-set $\s{v}$ can be mapped with a multiset of $k$-multisets:   $\s{v} \leftrightarrow \mms{\ms{v}=f(\s{v}, \s{n})\ |\  \sum_{i=1}^m \s{n}_i  = k, \forall i \ \s{n}_i \geq 1}$. Based on this relationship, we extend the connection among $k$-multisets to $k(\leq)$-sets: given $m_1,m_2\in[k]$, a $m_1$-set $\s{v}$ is connected to a $m_2$-set $\s{u}$ if and only if $\exists \s{n}_v, \s{n}_u$, $f(\s{v}, \s{n}_v)$ is connected with $f(\s{u}, \s{n}_u)$ in \kmwl. Let $S_{k\text{-swl}}(G)$ denote the defined super-graph on $G$ by \kswl. 
It can be shown that this is equivalent to either (1) $(|m_1-m_2|=1)\land (|\s{v} \cap \s{u}|=\min(m_1,m_2))$  or (2) $(m_1=m_2)\land (|\s{v} \cap \s{u}|=m_1-1)$ is true. Notice that $S_{k\text{-swl}}(G)$ contains a sequence of $k$$-$$1$ bipartite graphs with each reflecting the connections among the ($m$$-$$1$)-sets and the $m$-sets. It also contains $k$$-$$1$ subgraphs, i.e. the connections among the $m$-sets for $m=2,...,k$. Later on we will show that these $k$$-$$1$ subgraphs can be ignored without affecting \kswl. 

Let $\bm{swl}_k^{(t)}(G, \s{v})$ denote the color of $m$-set $\s{v}$ at $t$-th iteration of \kswl. Now we formally define \kswl. At $t=0$, we initialize the color of a $m$-set $\s{v}$ ($m\in[k]$) as: 
{\small\begin{align}
\bm{swl}_k^{(0)}(G, \s{v})= \text{HASH}\big( \mms{ \bm{mwl}_k^{(0)}(G, f(\s{v}, \s{n}) ) \ \big|\  \s{n}_1+...+\s{n}_m=k, \forall i \ \s{n}_i \geq 1  }\big) \label{eq:kswl-init}
\end{align}}
Clearly $\bm{swl}_k^{(0)}(G, \s{v})=\bm{swl}_k^{(0)}(G', \s{v}')$ if and only if $G[\s{v}]$ and $G'[\s{v}']$ are isomorphic. At $t$-th iteration, \kswl updates the color of every $m$-set $\s{v}$  by 
\begin{footnotesize}
\begin{align}
   &\bm{swl}_k^{(t+1)}(G, \s{v})= \text{HASH} \bigg( 
    \bm{swl}_k^{(t)}(G, \s{v}),   \mms{ \bm{swl}_k^{(t)}(G, \s{v}\cup\{x\} ) \ |\  x \in V(G)\setminus \s{v}}, \mms{ \bm{swl}_k^{(t)}(G, \s{v}\setminus {x} ) \ |\  x \in \s{v}},\nonumber
    \\
    \hspace{-0.3in}& \Bmms{\mms{ \bm{swl}_k^{(t)}(G, \s{v}[x/o^{-1}_G(\s{v},1)] ) \ |\  x \in V(G) \setminus \s{v}},...,\mms{ \bm{swl}_k^{(t)}(G, \s{v}[x/o^{-1}_G(\s{v},m)] ) \ |\  x \in V(G)\setminus \s{v} } }  \bigg)  \label{eq:kswl}
\end{align}
%\vspace{-0.2in}
\end{footnotesize}
Notice that when $m=1$ and $m=k$, the third and second part of the hashing input is an empty multiset, respectively. Similar to Eq. \eqref{eq:kgwl}, we formulate the graph level encoding as $\bm{gswl}_k^{(t)}(G) = \text{HASH}( \mms{ \bm{swl}_k^{(t)}\big(G, \s{v} \big) \ \big| \ \forall \s{v} \in V(S_{k\text{-swl}}(G))  } \big)$. 

To characterize the relationship of their expressivity, we first extend \kmwl on sets by defining the color of a $m$-set $\s{v}$ on \kmwl as $\bm{mwl}_k^{(t)}(G, \s{v}) := \mms{\bm{mwl}_k^{(t)}(G, f(\s{v}, \s{n}) ) \ \big|\  \sum_{i=1}^m \s{v}_i = k, \forall i \ \s{n}_i \geq 1}$. We prove that \kmwl is at least as expressive as \kswl in terms of separating node sets and graphs. 

\vspace{0mm}
\begin{theorem}
Let $k\ge 1$, then $\forall t\in \mathbb{N}$ and all graphs $G, G'$:
%\cbit
%\item[(1)] 
$\bm{mwl}_k^{(t)}(G, \s{v}) = \bm{mwl}_k^{(t)}(G', \s{v}') \Longrightarrow \bm{swl}_k^{(t)}(G, \s{v}) = \bm{swl}_k^{(t)}(G', \s{v}')$. 
%\item[(2)] $\bm{gmwl}_k^{(t)}(G) = \bm{gmwl}_k^{(t)}(G') \Longrightarrow \bm{gswl}_k^{(t)}(G) = \bm{gswl}_k^{(t)}(G')$
%\ceit
\end{theorem}

\vspace{-2mm}
We also conjecture that \kmwl and \kswl could be equally expressive, and leave it to future work. As a single $m$-set corresponds to $\binom{k-1}{m-1}$ $k$-multisets, moving from \kmwl to \kswl further reduces the computational cost greatly. 

% \bigg\{\mskip-10mu\
\subsection{From \kswl to \kcswl: Accounting for Sparsity}
\label{ssec:consider_sparsity}
Notice that for two arbitrary graphs $G$ and $G'$ with equal number of nodes, the number of $k(\leq)$-sets and the connections among all $k(\leq)$-sets in \kswl are exactly the same, regardless of whether they are dense or sparse. We next propose to \textit{account for the sparsity} of a  graph $G$ to further reduce the complexity of \kswl. As the graph structure is encoded inside every $m$-set, when the graph becomes sparser, there would be more sparse $m$-sets with a potentially large number of disconnected components. Based on the hypothesis that the induced subgraph over a set (of nodes) with fewer disconnected components naturally contains more structural information, we propose to restrict the $k(\leq)$-sets to be $(k,c)(\leq)$-sets: all sets with at most $k$ nodes and at most $c$ connected components in its induced subgraph. Let $S_{k,c\text{-swl}}(G)$ denote the super-graph defined by \kcswl, then $S_{k,c\text{-swl}}(G) = S_{k\text{-swl}}(G)[\{\s{v}| \text{\#components} (G[\s{v}])\le c\}]$, which is the induced subgraph on the super-graph defined by \kswl. Fortunately, $S_{k,c\text{-swl}}(G)$ can be efficiently and recursively constructed based on $S_{(k-1,c)\text{-swl}}(G)$, and we include the construction algorithm in the Appendix \ref{apdx:construct_supergraph}.   
\kcswl can be defined similarly to \kswl (Eq. \eqref{eq:kswl-init} and Eq. \eqref{eq:kswl}), however while removing all colors of sets that do not exist on $S_{k,c\text{-swl}}(G)$. 

\kcswl is progressively expressive with increasing $k$ and $c$, and when $c=k$, \kcswl becomes the same as \kswl, as all $k(\leq)$-sets are then considered. Let $\bm{swl}_{k,c}^{(t)}(G, \s{v})$ denote the color of a $(k,c)(\leq)$-set $\s{v}$ on $t$-th iteration of \kcswl, then  $\bm{gswl}_{k,c}^{(t)}(G) = \text{HASH}( \mms{ \bm{swl}_{k,c}^{(t)}\big(G, \s{v} \big) \big|\forall \s{v} \in S_{k,c\text{-swl}}(G) \big)}$.

\vspace{0mm}
\begin{theorem}\label{thm:kswl}
Let $k\ge 1$, then $\forall t\in \mathbb{N}$ and all graphs $G, G'$:

\vspace{-0.05in}
\cbit
\item[(1)] when $1\leq c_1 <c_2 \leq k$,
if $G,G'$ cannot be distinguished by  {\sc $(k,c_2)(\leq)$-SetWL}, they cannot be distinguished by {\sc $(k,c_1)(\leq)$-SetWL} 
\item[(2)] when $k_1 < k_2$, $\forall c\leq k_1$, if $G,G'$ cannot be distinguished by  {\sc $(k_2,c)(\leq)$-SetWL}, they cannot be distinguished by {\sc $(k_1,c)(\leq)$-SetWL}
\ceit
\end{theorem}

\vspace{-2mm}
\subsection{Complexity Analysis}
\label{ssec:complexity_analysis}
All color refinement algorithms described above run on a super-graph; thus, their complexity at each iteration is linear to the number of supernodes and number of edges of the super-graph. Instead of using big$\mathcal{O}$ notation that ignores constant factors, we compare the exact number of supernodes and edges. Let $G$ be the input graph with $n$ nodes and average degree $d$. 
For $k$-WL, there are $n^k$ supernodes and each has $n*k$ number of neighbors, hence $S_{k\text{-wl}}(G)$ has $n^k$ supernodes and $kn^{k+1}/2$ edges. For $m\in [k]$, there are $\binom{n}{m}$ $m$-sets and each connects to $m$ number of $(m-1)$-sets. So $S_{k\text{-swl}}$ has $\sum_{i=1}^k \binom{n}{i} \le \binom{n}{k}\frac{n-k+1}{n-2k+1}$ supernodes and $\sum_{i=2}^k i\binom{n}{i}=n\sum_{i=1}^{k-1}\binom{n-1}{i} \le n\binom{n-1}{k-1}\frac{n-k+1}{n-2k+2} $ edges (derivation in Appendix). Here we ignore edges within $m$-sets for any $m\in [k]$ as they can be reconstructed from the bipartite connections among $(m-1)$-sets and $m$-sets, described in detail in Sec. \ref{ssec:to_gnn}. Consider e.g., $n=30,k=5$; we get $\frac{|V(S_{k\text{-wl}}(G))|}{|V(S_{k\text{-swl}}(G))|}=139$, $\frac{|E(S_{k\text{-wl}}(G))|}{|E(S_{k\text{-swl}}(G))|}=2182$. Directly analyzing the savings by restricting number of components is not possible without assuming a graph family. In Sec. \ref{ssec:runtime} we measured the scalability for \method with different number of components directly on sparse graphs, where \kcswl shares similar scalability. 

\subsection{Set version of $k$-FWL} 
\label{ssec:fwl}
$k$-FWL is a stronger GI algorithm and it has the same expressivity as $k$+$1$-WL \cite{grohe2021logic}, in this section we also demonstrate how to extend the set to $k$-FWL to get \ksfwl. Let $\bm{fwl}_k^{(t)}(G, \tps{v})$ denote the color of $k$-tuple $\tps{v}$ at $t$-th iteration of $k$-FWL. Then $k$-FWL is initialized the same as the $k$-WL, i.e.  $\bm{fwl}_k^{(0)}(G, \tps{v})$ =  $\bm{wl}_k^{(0)}(G, \tps{v})$. At $t$-th iteration, $k$-FWL updates colors with
\begin{align}
 \scriptstyle
    \bm{fwl}_k^{(t+1)}(G, \tps{v}) = \text{HASH} \Big( 
    \bm{fwl}_k^{(t)}(G, \tps{v}),  
    \bmms{ \big(\bm{fwl}_k^{(t)}(G, \tps{v}[x/1]), ..., \bm{fwl}_k^{(t)}(G, \tps{v}[x/k]) \big) \Big| x \in V(G)
    }
    %  \Bmms{\bm{fwl}_k^{(t)}(G, \tps{v}[x/1]) \Big| x \in V(G)}, \ldots, \Bmms{\bm{fwl}_k^{(t)}(G, \tps{v}[x/k]) \Big| x \in V(G)}
    \Big) \label{eq:kfwl}  
\end{align}
Let $\bm{sfwl}_k^{(t)}(G, \s{v})$ denote the color of $m$-set $\s{v}$ at $t$-th iteration of \ksfwl. Then at $t$-th iteration it updates with 
\begin{align}
 \scriptstyle
    \bm{sfwl}_k^{(t+1)}(G, \s{v}) = \text{HASH} \Big( 
    \bm{sfwl}_k^{(t)}(G, \s{v}),  
    \bmms{ \bbmms{\bm{sfwl}_k^{(t)}(G, \s{v}[x/1]), ..., \bm{sfwl}_k^{(t)}(G, \s{v}[x/m]) } \Big| x \in V(G)
    }
    \Big) \label{eq:ksfwl}  
\end{align}
The \ksfwl should have better expressivity than \kswl. We show in the next section that \kswl can be further improved with less computation through an intermediate step while this is nontrivial for \ksfwl. We leave it to future work of studying \ksfwl.

\section{A practical progressively-expressive GNN: \method}
In this section we transform \kcswl to a GNN model by replacing the HASH function in Eq. \eqref{eq:kswl} with a combination of MLP and DeepSet \cite{zaheer2017deep}, given they are universal function approximators for vectors and sets, respectively \cite{hornik1989multilayer, zaheer2017deep}. After the transformation, we propose two additional improvements (Sec. \ref{ssec:bidrectional} and Sec. \ref{ssec:supernode_init}) to further improve scalability. We work on vector-attributed graphs. Let $G=(V(G), E(G), X)$ be an undirected graph with node features $\rvx_i\in \mathbb{R}^d, \forall i \in V(G)$.

\subsection{From $(k,c)(\leq)$-SetWL to $(k,c)(\leq)$-SetGNN}
\label{ssec:to_gnn}
\kcswl defines a super-graph $S_{k,c\text{-swl}}$, which  %and Eq. \eqref{eq:kswl} 
aligns with Eq. \eqref{eq:kswl}.
%$S_{k,c\text{-swl}}$. 
We first rewrite Eq. \eqref{eq:kswl} to reflect its connection to $S_{k,c\text{-swl}}$. For a supernode $\s{v}$ in $S_{k,c\text{-swl}}$, let $\mathcal{N}^G_{\text{left}}(\s{v})=\{\s{u} \ |\ \s{u}\in S_{k,c\text{-swl}}, \s{u}\leftrightarrow \s{v} \text{ and } |\s{u}| = |\s{v}| - 1  \}$, and  $\mathcal{N}^G_{\text{right}}(\s{v})=\{\s{u} \ |\ \s{u}\in S_{k,c\text{-swl}}, \s{u}\leftrightarrow \s{v} \text{ and } |\s{u}| = |\s{v}| + 1  \}$. Then we can rewrite Eq. \eqref{eq:kswl} for \kcswl as 
\vspace{-0.05in}
\begin{scriptsize}
\begin{align}
    \bm{swl}_{k,c}^{(t+1)}(G, \s{v}) = \bigg( 
    \bm{swl}_{k,c}^{(t)}(G, \s{v}), 
    \bm{swl}_{k,c}^{(t+\frac{1}{2})}(G, \s{v}), 
    \mms{\bm{swl}_{k,c}^{(t)}(G, \s{u}) \ |\ \s{u} \in \mathcal{N}^G_{\text{left}}(\s{v})},
    \mms{\bm{swl}_{k,c}^{(t+\frac{1}{2})}(G, \s{u}) \ |\ \s{u} \in \mathcal{N}^G_{\text{left}}(\s{v}) } 
    \bigg)   \label{eq:kcswl}
% \vspace{-0.3in}
\end{align}
\end{scriptsize}
%\vspace{-0.2in}
where $ \bm{swl}_{k,c}^{(t+\frac{1}{2})}(G, \s{v}):= \mms{\bm{swl}_{k,c}^{(t)}(G, \s{u}) \ |\ \s{u} \in \mathcal{N}^G_{\text{right}}(\s{v})}$. Notice that we omit $\text{HASH}$ and apply it implicitly. Eq. \eqref{eq:kcswl} essentially splits the computation of 
Eq. \eqref{eq:kswl} into two steps and %reduces  overhead by 
avoids repeated computation via caching the explicit $t$$+$$\frac{1}{2}$ step. It also implies that the connection among $m$-sets for any $m\in[k]$ can be reconstructed from the bipartite graph among $m$-sets and $(m-1)$-sets.

Next we formulate \method formally. Let $h^{(t)}(\s{v}) \in \mathbb{R}^{d_t}$ denote the vector representation of supernode $\s{v}$ on the $t$-th iteration of \method. For any input graph $G$, it initializes representations of supernodes by 

\vspace{-0.2in}
\begin{align}
    h^{(0)}(\s{v}) = \text{BaseGNN} (G[\s{v}]) \label{eq:setgnn-init}
\end{align}
where the BaseGNN can be any GNN model. Theoretically the BaseGNN should be chosen to encode non-isomorphic induced subgraphs distinctly, mimicking HASH. Based on   empirical tests of several GNNs on all (11,117) possible $8$-node non-isomorphic graphs \cite{balcilar2021breaking}, and given GIN \cite{Xu:2019ty} is simple and fast with nearly $100\%$ separation rate, we use GIN as our BaseGNN in experiments. 

\method iteratively updates representations of all supernodes by
\vspace{-0.1in}
\begin{footnotesize}
\begin{align}
\scriptstyle
    h^{(t+\frac{1}{2})}(\s{v}) &= \sum_{\s{u}\in \mathcal{N}^G_{\text{right}}(\s{v})} \text{MLP}^{(t+\frac{1}{2})}(h^{(t)}(\s{u})) \label{eq:SetGNN1}\\
    h^{(t+1)}(\s{v}) &= \text{MLP}^{(t)}\Big(h^{(t)}(\s{v}),h^{(t+\frac{1}{2})}(\s{v}), 
    \sum_{\s{u}\in \mathcal{N}^G_{\text{left}}(\s{v})} \text{MLP}^{(t)}_A(h^{(t)}(\s{u})),
    \sum_{\s{u}\in \mathcal{N}^G_{\text{left}}(\s{v})} \text{MLP}^{(t)}_B(h^{(t+\frac{1}{2})}(\s{u}))
    \Big)\label{eq:SetGNN2}
\end{align}
\end{footnotesize}
\vspace{-0.1in}
Then after $T$ iterations, we compute the graph level encoding as 
\begin{align}
    h^{(T)}(G) = \text{POOL}(  \mms{h^{(T)}( \s{v}) \ |\  \s{v} \in  V(S_{k,c\text{-swl}}) }  )
\end{align}
where POOL can be chosen as summation. We visualize the steps in Figure \ref{fig:chart}. Under mild conditions, \method has the same expressivity as \kcswl. 

% \begingroup
% \setlength{\topsep}{0pt}
\vspace{0mm}
\begin{theorem}\label{thm:setgnn}
When ($i$) BaseGNN can distinguish any non-isomorhpic graphs with at most $k$ nodes, ($ii$) all MLPs have sufficient depth and width, and ($iii$) POOL is an injective function, then for any $t\in \mathbb{N}$, $t$-layer \method is as expressive as \kcswl at the $t$-th iteration.     
\end{theorem}
% \endgroup

\vspace{-2mm}
The following facts can be derived easily from Theorem \ref{thm:kswl} and Theorem \ref{thm:setgnn}. 

\vspace{-1mm}
\begin{corollary}
\method is progressively-expressive with increasing $k$ and $c$, that is,

\vspace{-0.05in}
\cbit
\item[(1)] when $c_1 > c_2$, {\sc $(k,c_1)(\leq)$-SetGNN} is more expressive than  {\sc $(k,c_2)(\leq)$-SetGNN}, and
\item[(2)] when $k_1 > k_2$, {\sc $(k_1,c)(\leq)$-SetGNN} is more expressive than  {\sc $(k_2,c)(\leq)$-SetGNN}.
\ceit
\end{corollary}

\subsection{Bidirectional Sequential Message Passing}\label{ssec:bidrectional}
The $t$-th layer of \method (Eq. \eqref{eq:SetGNN1} and Eq. \eqref{eq:SetGNN2}) are essentially propagating information back and forth on the super-graph $S_{k,c\text{-swl}}(G)$, which is a sequence of $k$$-$$1$ bipartite graphs (see the middle of Figure \ref{fig:chart}), in \textit{parallel} for all supernodes. We propose to change it to bidirectional blockwise \textit{sequential} message passing, which we call \methods, defined as follows. 

\vspace{-0.2in}
\begin{footnotesize}
% \vspace{-0.05in}
\begin{align}
   & m=k-1 \text{ to } 1, \forall m\text{-set } \s{v},   h^{(t+\frac{1}{2})}(\s{v}) =  \text{MLP}_{m,1}^{(t)}\Big(
    h^{(t)}(\s{v}),
    \sum_{\s{u}\in \mathcal{N}^G_{\text{right}}(\s{v})} \text{MLP}_{m,2}^{(t)}(h^{(t+\frac{1}{2})}(\s{u})) 
    \Big)  \label{eq:bmp1}\\ 
   & m=2 \text{ to } k, \forall m\text{-set } \s{v},   h^{(t+1)}(\s{v}) =  \text{MLP}_{m,1}^{(t+\frac{1}{2})}\Big(
    h^{(t+\frac{1}{2})}(\s{v}),
    \sum_{\s{u}\in \mathcal{N}^G_{\text{left}}(\s{v})} \text{MLP}_{m,2}^{(t+\frac{1}{2})}(h^{(t+1)}(\s{u})) 
    \Big) \label{eq:bmp2}
\end{align}
% \vspace{-0.05in}
\end{footnotesize}
\vspace{-0.2in}

Notice that \methods has lower memory usage, as \method load the complete supergraph ($k$$-$$1$ bipartites) while \methods loads 1 out of $k$$-$$1$ bipartites at a time, which is beneficial for limited-size GPUs. What is  more, %we proved the following theorem.
for a small, finite $t$ it is even more expressive than \method. We provide both implementation of parallel and sequential message passing in the official github repository, while only report the performance of \methods given its efficiency and better expressivity. 

\vspace{0in}
\begin{theorem}
For any $t\in\mathbb{N}$, the $t$-layer \methods is more expressive than the $t$-layer \method. As $\lim_{t\to\infty}$, \method is as expressive as \methods.
\end{theorem}

\subsection{Improving Supernode Initialization}
\label{ssec:supernode_init}
Next we describe how to improve the supernode initialization (Eq. \eqref{eq:setgnn-init}) and extensively reduce computational and memory overhead for $c>1$ without losing any expressivity. We achieve this by the fact that a graph with $c$ components can be viewed as a set of $c$ connected components. Formally,

\vspace{0mm}
\begin{theorem}
\label{thm:c}
 Let $G$ be a graph with $c$ connected components $C_1,...,C_c$, and $G'$ be a graph also with $c$ connected components $C'_1,...,C'_c$, then $G$ and $G'$ are isomorphic if and only if $\exists p: [c]\rightarrow [c]$, s.t.  $\forall i\in [c]$, $C_i$ and $C'_{p(i)}$ are isomorphic. 
\end{theorem}

\vspace{-2.5mm}
The theorem implies that we only need to apply Eq. \eqref{eq:setgnn-init} to all supernodes with a single component, and for any $c$-components $\s{v}$ with components $\s{u}_1,..., \s{u}_c$, we can get the encoding by $h^{(0)}(\s{v}) = \text{DeepSet}(\{h^{(0)}(\s{u}_1),..., h^{(0)}(\s{u}_c)\}) $ without passing its induced subgraph to BaseGNN. This eliminates the heavy computation of passing a large number of induced subgraphs. We  give the algorithm of building connection between $\s{v}$ to its single components in Appendix \ref{apdx:inference_graph}.

\section{Experiments}

We design experiments to answer the following questions. \textbf{Q1. Performance:} How does \methods compare to SOTA expressive GNNs?
\textbf{Q2. Varying $k$ and $c$}: How does the progressively increasing expressiveness reflect on generalization performance?  
\textbf{Q3. Computational requirements:} Is \method feasible on practical graphs w.r.t. running time and memory usage? %-- can report wallclock time. is it hours or minutes or days, people will wonder.}

\subsection{Setup}
\label{ssec:setup}

{\bf Datasets.~} To inspect the expressive power, we use four different types of simulation datasets: \textbf{1)} EXP \citep{abboud2020surprising} contains 600 pairs of $1\&2$-WL failed graphs which we split into two where graphs in each pair is assigned to two different classes; \textbf{2)} SR25 \citep{balcilar2021breaking} has 15 strongly regular graphs ($3$-WL failed) with 25 nodes each, which we transform to a 15-way classification task; % with the goal of mapping each graph into a different class. 
\textbf{3)} Substructure counting (i.e. triangle, tailed triangle, star and 4-cycle) tasks on random graph dataset \citep{chen2020can}; \textbf{4)} Graph property regression (i.e. connectedness, diameter, radius) tasks on random graph dataset \citep{pna}.
We also evaluate performance on two real world graph learning tasks: \textbf{5)} ZINC-12K \citep{dwivedi2020benchmarking}, and 
\textbf{6)} QM9 \cite{wu2018moleculenet} for molecular property regression.
See Table \ref{tab:data-semantics} in  Appendix \ref{ssec:data} for detailed dataset statistics.

{\bf Baselines.~}
We use GCN \citep{gcn}, GIN \citep{Xu:2019ty}, PNA$^*$ \citep{pna}, PPGN \citep{maron2019provably}, PF-GNN \citep{dupty2022pfgnn}, and GNN-AK \citep{zhao2021stars} as baselines on the simulation datasets.
On ZINC-12K we also reference CIN \citep{bodnar2021weisfeiler} directly from literature. Most baselines results are taken from \cite{zhao2021stars}.
Finally, we compare to GINE \cite{hu2020pretraining} on QM9. \textbf{Hyperparameter and model configurations} are described in Appendix \ref{apdx:experimental-setup}.

%TODO for Lingxiao: mention that we use published results from prior work in lieu of optimizing the performance of baseline methods where appropriate (e.g. for Table 4).

\subsection{Results}
\label{ssec:results}
Theorem \ref{thm:cfi-mwl} shows that \kmwl is able to distinguish CFI(k) graphs.
It also holds for \kswl as the proof doesn't use repetitions. 
We implemented the construction of CFI(k) graphs for any k. Empirically we found that \textit{\methods is able to distinguish CFI(k) for k = 3, 4, 5 (k>6 out of memory)}. This empirically verifies its theoretical expressivity.

Table \ref{tab:simulation} shows the performance results on all the simulation datasets. The low expressive models such as GCN, GIN and PNA$^*$ underperform on EXP and SR25, while 3-WL equivalent PPGN excels only on EXP.
Notably, \methods achieves 100\% discriminating power with any $k$ larger than 2 and 3, and any $c$ larger than 1 and 0, resp. for EXP and SR25.
\methods also outperforms the baselines on all substructure counting tasks, as well as on two out of three graph property prediction tasks (except Radius) with significant gap, for relatively small values of $k$ and $c$.

\begin{table}[t!]
\setlength{\tabcolsep}{4pt}
%\setlength\extrarowheight{-5pt}
%\vspace{-0.15in}
\fontsize{8}{8}\selectfont
    \caption{Simulation data performances.
    %Accuracy on EXP\&SR25, error on rest.
    %{\method \la{take aways}}. 
    For \methods, $(k,c)$ values that achieve reported performance  in parenthesis.
    (ACC: accuracy, MA[S]E: mean abs.[sq.] error) %, OOM: out of memory) 
    }
    \label{tab:simulation}
   % \vspace{-0.1in}
    \centering
    \begin{tabular}{l r r cccc  ccc}
    \toprule
     \multirow{2}{*}{Method} & \multirow{2}{*}[-0mm]{\makecell{EXP \\ (ACC)}} & \multirow{2}{*}[-0mm]{\makecell{SR25 \\ (ACC)}} & \multicolumn{4}{c}{Counting Substructures (MAE)} & \multicolumn{3}{c}{Graph Properties ($\log_{10}$(MSE))} \\
     \cmidrule(l{2pt}r{2pt}){4-7} \cmidrule(l{2pt}r{2pt}){8-10}
                             &      &       & Triangle & Tailed Tri. & Star & 4-Cycle          &  IsConnected & Diameter & Radius  \\ 
    \midrule  
     GCN      &  50\%& 6.67\%& 0.4186 & 0.3248 & 0.1798 & 0.2822 & -1.7057 & -2.4705 & -3.9316 \\
   % \textsf{GCN-AK$^+$} & \BFSERIES100\%& 6.67\%& 0.0137 & 0.0134 & 0.0174 & 0.0183 & -2.6705 & -3.9102 & -5.1136\\
  %  \midrule
     GIN       &  50\%& 6.67\%& 0.3569 & 0.2373 & 0.0224 & 0.2185 & -1.9239 & -3.3079 & -4.7584 \\
  %  \revise{\textsf{GIN-AK}}&  \revise{\BFSERIES100\%} & \revise{6.67\%} & 0.0934 & 0.0751 & 0.0216 & 0.0726 \\
   % \textsf{GIN-AK$^+$}  & \BFSERIES100\%& 6.67\%& 0.0123 & 0.0112 & 0.0150 & 0.0126 & -2.7513 & -3.9687 & -5.1846\\
  %  \midrule

     PNA$^*$        &  50\%& 6.67\%& 0.3532 & 0.2648 & 0.1278 & 0.2430 &-1.9395 &-3.4382 &-4.9470 \\
   % \textsf{PNA$^*$-AK$^+$}   & \BFSERIES100\%& 6.67\%& 0.0118 & 0.0138 & 0.0166 & 0.0132 &-2.6189 & -3.9011& -5.2026\\
   % \midrule
     PPGN        & \textbf{100\%}& 6.67\%&0.0089 &0.0096  &0.0148  &0.0090  & -1.9804 & -3.6147 & -5.0878 \\
    GIN-AK$^+$  & \textbf{100\%}& 6.67\%& 0.0123 & 0.0112 & 0.0150 & 0.0126 & -2.7513 & -3.9687 & -5.1846\\
     {PNA$^*$-AK$^+$}   & \textbf{100\%}& 6.67\%& 0.0118 & 0.0138 & 0.0166 & 0.0132 &-2.6189 & -3.9011& \textbf{-5.2026}\\
 %    {PPGN-AK$^+$}   & \textbf{100\%}&  \textbf{100\%}& OOM & OOM& OOM& OOM& OOM& OOM& OOM\\
  %  \textsf{PPGN-AK$^+$}   & 100\%&  \BFSERIES100\%& OOM & OOM& OOM& OOM& OOM& OOM& OOM\\
  \midrule 
  $(k,c)$\amethod & \textbf{100\%} & \textbf{100\%} & \textbf{0.0073}  & \textbf{0.0075}  & \textbf{0.0134}  & \textbf{0.0075}  & \textbf{-5.4667} & \textbf{-4.0800} & {-5.1603}  \\
                & \multicolumn{1}{c}{($\geq$$3$, $\geq$$2$)} & 
                \multicolumn{1}{c}{($\geq$$4$, $\geq$$1$)}  & 
                \multicolumn{1}{c}{($3$, $2$)}&  
                \multicolumn{1}{c}{($4$, $1$)} &  
                \multicolumn{1}{c}{($3$, $2$)}& 
                \multicolumn{1}{c}{($4$, $1$)}&  
                \multicolumn{1}{c}{($4$, $2$)} & 
                \multicolumn{1}{c}{($4$, $1$)} & 
                \multicolumn{1}{c}{($2$, $2$)}\\
    \bottomrule         
    \end{tabular}
    \vspace{-0.3in}
\end{table}
\begin{table}[t!]
\caption{Train and Test performances on substructure counting tasks by varying $k$ and $c$. Notice the orders of magnitude drop in Test MAE between bolded entries per task.}
    \label{tab:simbykc}
    {\scalebox{0.6}{
    \begin{tabular}{rr|llllllll}
    \toprule
\multicolumn{1}{l}{}  & \multicolumn{1}{l}{}  & \multicolumn{8}{c}{Counting Substructures (MAE)}                                                                                                                                                                          \\\midrule
\multicolumn{1}{c}{}  & \multicolumn{1}{c}{}  & \multicolumn{2}{c}{Triangle}                         & \multicolumn{2}{c}{Tailed Tri.}                      & \multicolumn{2}{c}{Star}                             & \multicolumn{2}{c}{4-Cycle}                          \\
\midrule
\multicolumn{1}{l}{k} & \multicolumn{1}{l}{c} & \multicolumn{1}{c}{Train} & \multicolumn{1}{c}{Test} & \multicolumn{1}{c}{Train} & \multicolumn{1}{c}{Test} & \multicolumn{1}{c}{Train} & \multicolumn{1}{c}{Test} & \multicolumn{1}{c}{Train} & \multicolumn{1}{c}{Test} \\
\midrule
2                     & 1                     & 0.9941 ± 0.2623           & \textbf{1.1409 ± 0.1224}          & 1.1506 ± 0.2542           & \textbf{0.8695 ± 0.0781}          & 1.5348 ± 2.0697           & \textbf{2.3454 ± 0.8198}          & 1.2159 ± 0.0292           & 0.8361 ± 0.1171          \\
3                     & 1                     & 0.0311 ± 0.0025           & \textbf{0.0088 ± 0.0001}          & 0.0303 ± 0.0108           & \textbf{0.0085 ± 0.0018}          & 0.0559 ± 0.0019           & \textbf{0.0151 ± 0.0006}          & 0.1351 ± 0.0058           & \textbf{0.1893 ± 0.0030}          \\
4                     & 1                     & 0.0321 ± 0.0008           & 0.0151 ± 0.0074          & 0.0307 ± 0.0085           & 0.0075 ± 0.0012          & 0.0687 ± 0.0104           & 0.0339 ± 0.0009          & 0.0349 ± 0.0007           & \textbf{0.0075 ± 0.0002}          \\
5                     & 1                     & 0.0302 ± 0.0070           & 0.0208 ± 0.0042          & 0.0553 ± 0.0009           & 0.0189 ± 0.0024          & 0.0565 ± 0.0078           & 0.0263 ± 0.0023          & 0.0377 ± 0.0057           & 0.0175 ± 0.0036          \\
6                     & 1                     & 0.0344 ± 0.0024           & 0.0247 ± 0.0085          & 0.0357 ± 0.0017           & 0.0171 ± 0.0000          & 0.0560 ± 0.0000           & 0.0168 ± 0.0022          & 0.0356 ± 0.0014           & 0.0163 ± 0.0064          \\
2                     & 2                     & 0.3452 ± 0.0329           & 0.4029 ± 0.0053          & 0.2723 ± 0.0157           & 0.2898 ± 0.0055          & 0.0466 ± 0.0025           & 0.0242 ± 0.0006          & 0.2369 ± 0.0123           & 0.2512 ± 0.0029          \\
3                     & 2                     & 0.0234 ± 0.0030           & 0.0073 ± 0.0009          & 0.0296 ± 0.0074           & 0.0100 ± 0.0009          & 0.0640 ± 0.0003           & 0.0134 ± 0.0006          & 0.0484 ± 0.0135           & 0.0194 ± 0.0065          \\
4                     & 2                     & 0.0587 ± 0.0356           & 0.0131 ± 0.0010          & 0.0438 ± 0.0140           & 0.0094 ± 0.0002          & 0.0488 ± 0.0008           & 0.0209 ± 0.0063          & 0.0464 ± 0.0037           & 0.0110 ± 0.0020         \\
\bottomrule
\end{tabular}
}}
\vspace{-0.2in}
\end{table}

In Table \ref{tab:simbykc}, we show the train and test MAEs on substructure counting tasks for individual values of $k$ and $c$.
As expected, performances improve for increasing $k$ when $c$ is fixed, and vice versa. It is notable that orders of magnitude improvements on test error occur 
moving from $k$$=$$2$ to $3$ for the triangle tasks as well as the star task, while a similarly large magnitude drop is obtained 
at $k$$=$$4$ for the 4-cycle task, which is expected as triangle and 4-cycle have 3 and 4 nodes respectively.
Similar observations hold for graph property tasks as well. (See Appendix \ref{apdx:graph-property})

\begin{wraptable}{r}{4cm}
\vspace{-0.2in}
\fontsize{8}{8}\selectfont
\caption{SetGNN$^*$ achieves SOTA on ZINC-12K.}
\vspace{-0.1in}
\label{tab:zinc_mae}
{\scalebox{0.9}{
\begin{tabular}{lc}
    \toprule
    Method & MAE \\
    \midrule
    GatedGCN  & 0.363 $\pm$ 0.009 \\
    GCN & 0.321 $\pm$ 0.009 \\
    PNA  & 0.188 $\pm$ 0.004 \\
    DGN  & 0.168 $\pm$ 0.003 \\
    GIN & 0.163 $\pm$ 0.004 \\
    GINE & 0.157 $\pm$ 0.004 \\
    HIMP & 0.151 $\pm$ 0.006 \\
    PNA$^*$ & 0.140 $\pm$ 0.006 \\
    GSN & 0.115 $\pm$ 0.012 \\
    PF-GNN & 0.122 $\pm$ 0.010 \\
    GIN-AK$^+$ & 0.080 $\pm$ 0.001 \\
    CIN & 0.079 $\pm$ 0.006 \\
    \midrule
    $(k,c)$\amethod & \textbf{0.0750 $\pm$ 0.0027} \\
    \bottomrule
\end{tabular}
}}
\vspace{-0.25in}
\end{wraptable}

In addition to simulation datasets, we evaluate our \methods on real-world data; Table \ref{tab:zinc_mae} shows our performance on  ZINC-12K.   Our method achieves a new state-of-the-art performance, with a mean absolute error (MAE) of 0.0750, using $k$$=$$5$ and $c$$=$$2$.

In Table \ref{tab:zincbykc} we show the test and validation MAE along with training loss for varying $k$ and $c$ for ZINC-12K.
For a fixed $c$, validation MAE and training loss both follow a first decaying and later increasing trend with increasing $k$, potentially owing to the difficulty in fitting with too many sets (i.e. supernodes) and edges in the super-graph. 

Similar results are shown for QM9 in Table \ref{tab:qm9bykc}. For comparison we also show GINE performances using both 4 and 6 layers, both of which are significantly lower than \methods.

\begin{table}[h]
\vspace{-0.2in}
\begin{minipage}{0.46\linewidth}
    \centering
    \caption{\methods performances on ZINC-12K  by varying ($k$,$c$). %Test MAE at lowest Val. MAE highlighted in \textbf{bold}.
    \textbf{Test MAE} at lowest Val. MAE, and lowest \underline{Test MAE}.
    }
    \label{tab:zincbykc}
    {
        \scalebox{0.7}{
        \begin{tabular}{cclll}
        \toprule
        \multicolumn{1}{c}{$k$} & \multicolumn{1}{c}{$c$} & {Train loss}      & {Val.} MAE         & {Test MAE}        \\
        \midrule
        2                     & 1                     & 0.1381 ± 0.0240 & 0.2429 ± 0.0071 & 0.2345 ± 0.0131 \\
        
        3                     & 1                     & 0.1172 ± 0.0063 & 0.2298 ± 0.0060 & 0.2252 ± 0.0030 \\

        4                     & 1                     & 0.0693 ± 0.0111 & 0.1645 ± 0.0052 & 0.1636 ± 0.0052 \\
        5                     & 1                     & 0.0643 ± 0.0019 & 0.1593 ± 0.0051 & 0.1447 ± 0.0013 \\
        6                     & 1                     & 0.0519 ± 0.0064 & 0.0994 ± 0.0093 & 0.0843 ± 0.0048 \\
        7                     & 1                     & 0.0543 ± 0.0048 & 0.0965 ± 0.0061 & 0.0747 ± 0.0022 \\
        8                     & 1                     & 0.0564 ± 0.0152 & 0.0961 ± 0.0043 & \underline{0.0732 ± 0.0037} \\
        9                     & 1                     & 0.0817 ± 0.0274 & 0.0909 ± 0.0094 & 0.0824 ± 0.0056 \\
        10                    & 1                     & 0.0894 ± 0.0266 & 0.1060 ± 0.0157 & 0.0950 ± 0.0102 \\
        \midrule
        2                     & 2                     & 0.1783 ± 0.0602 & 0.2913 ± 0.0102 & 0.2948 ± 0.0210 \\
        3                     & 2                     & 0.0640 ± 0.0072 & 0.1668 ± 0.0078 & 0.1391 ± 0.0102 \\
        4                     & 2                     & 0.0499 ± 0.0043 & 0.1029 ± 0.0033 & 0.0836 ± 0.0010 \\
        5                     & 2                     & 0.0483 ± 0.0017 & {0.0899 ± 0.0056} & \textbf{0.0750 ± 0.0027} \\
        6                     & 2                     & 0.0530 ± 0.0064 & 0.0927 ± 0.0050 & 0.0737 ± 0.0006 \\
        7                     & 2                     & 0.0547 ± 0.0036 & 0.0984 ± 0.0047 & 0.0784 ± 0.0043 \\
        \midrule
        3                     & 3                     & 0.0798 ± 0.0062 & 0.1881 ± 0.0076 & 0.1722 ± 0.0086 \\
        4                     & 3                     & 0.0565 ± 0.0059 & 0.1121 ± 0.0066 & 0.0869 ± 0.0026 \\
        5                     & 3                     & 0.0671 ± 0.0156 & 0.1091 ± 0.0097 & 0.0920 ± 0.0054 \\
        \bottomrule
        \end{tabular}
        }
    }
\end{minipage}\qquad
\begin{minipage}{0.49\linewidth}
    \caption{\methods performances on QM9  by varying ($k$,$c$). \textbf{Test MAE} at lowest Val. MAE, and lowest \underline{Test MAE}. All variances are ${\leq}0.002$ and thus omitted.}
    \label{tab:qm9bykc}
    \setlength{\tabcolsep}{3pt}

    \fontsize{9}{10.6}\selectfont
	{
    	\scalebox{0.8}{
    	
        \begin{tabular}{cclll}
        \toprule
        \multicolumn{1}{c}{$k$} & \multicolumn{1}{c}{$c$} &  {Train loss}      & {Val.} MAE         & {Test MAE}         \\\midrule
        % {$(k$} & {$c)$} & $\leq${\sc SetGNN}     &          &       \\
        % \midrule
        2 & 1 & 0.0376 ± 0.0005               & 0.0387 ± 0.0007               & 0.0389 ± 0.0008               \\
        3 & 1 & 0.0308 ± 0.0010               & 0.0386 ± 0.0017               & 0.0379 ± 0.0010               \\
        4 & 1 & 0.0338 ± 0.0003               & 0.0371 ± 0.0005               & 0.0370 ± 0.0006               \\
        5 & 1 & 0.0299 ± 0.0017               & 0.0343 ± 0.0008               & 0.0341 ± 0.0009               \\
        6 & 1 & 0.0226 ± 0.0004               & 0.0296 ± 0.0007               & 0.0293 ± 0.0007               \\
        7 & 1 & 0.0208 ± 0.0005               & {0.0289 ± 0.0007}               & \underline{0.0269 ± 0.0003}               \\
        \midrule
        2 & 2 & 0.0367 ± 0.0007               & 0.0398 ± 0.0004               & 0.0398 ± 0.0004               \\
        3 & 2 & 0.0282 ± 0.0013               & 0.0358 ± 0.0009               & 0.0356 ± 0.0007               \\
        4 & 2 & 0.0219 ± 0.0004               & 0.0280 ± 0.0008               & 0.0278 ± 0.0008               \\
        5 & 2 & 0.0175 ± 0.0003               & {0.0267± 0.0005}               & \textbf{0.0251± 0.0006}                     \\
        \midrule
        3 & 3 & 0.0391 ± 0.0107              & 0.0428 ± 0.0057              & 0.0425 ± 0.0052             \\
        4 & 3 & 0.0219 ± 0.0011               & 0.0301 ± 0.0010               & 0.0286 ± 0.0004   \\
        \hline \hline
        \multicolumn{2}{l}{GINE} ($L$$=$$4$) &  0.0507 ± 0.0014 &	0.0478 ± 0.0003 &	0.0479 ± 0.0004 \\
        \multicolumn{2}{l}{GINE} ($L$$=$$6$) &  0.0440 ± 0.0009 &	0.0440 ± 0.0009 &	0.0451 ± 0.0009 \\
        \bottomrule
        \end{tabular}
        }
    }
\end{minipage}
\vspace{-0.15in} 
\end{table}

\subsection{Computational requirements}
\label{ssec:runtime}
\begin{wrapfigure}{rh}{0.4\textwidth}
	\vspace{-0.45in}
	\centering
	\begin{subfigure}{0.18\textwidth}
	    \centering
	    \includegraphics[width=3.1cm]{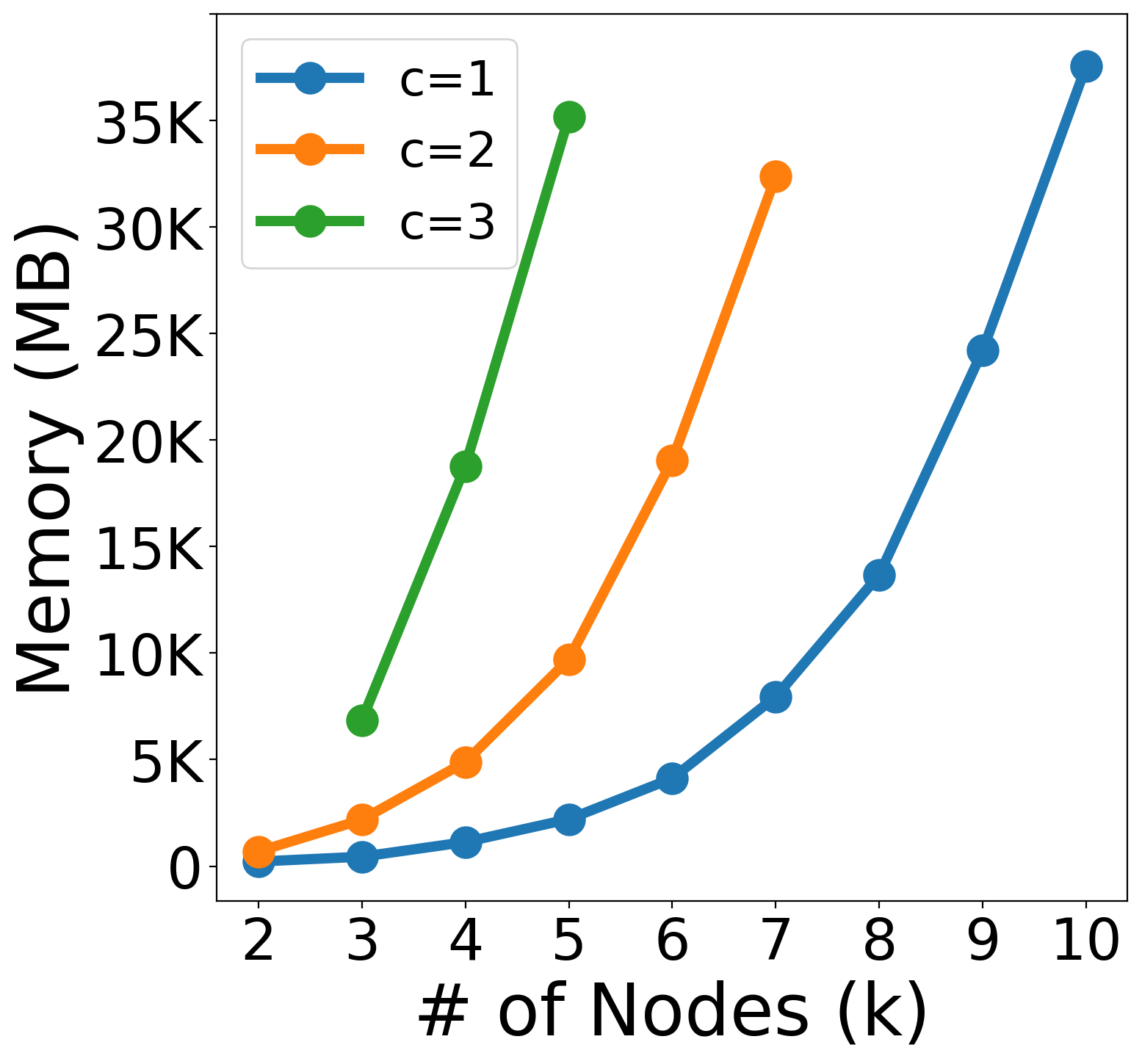}
	    \caption{Memory usage}
	    \label{fig:zinc_memory}
	\end{subfigure}\quad\;
	\begin{subfigure}{0.18\textwidth}
	    \centering
	    \includegraphics[width=3cm]{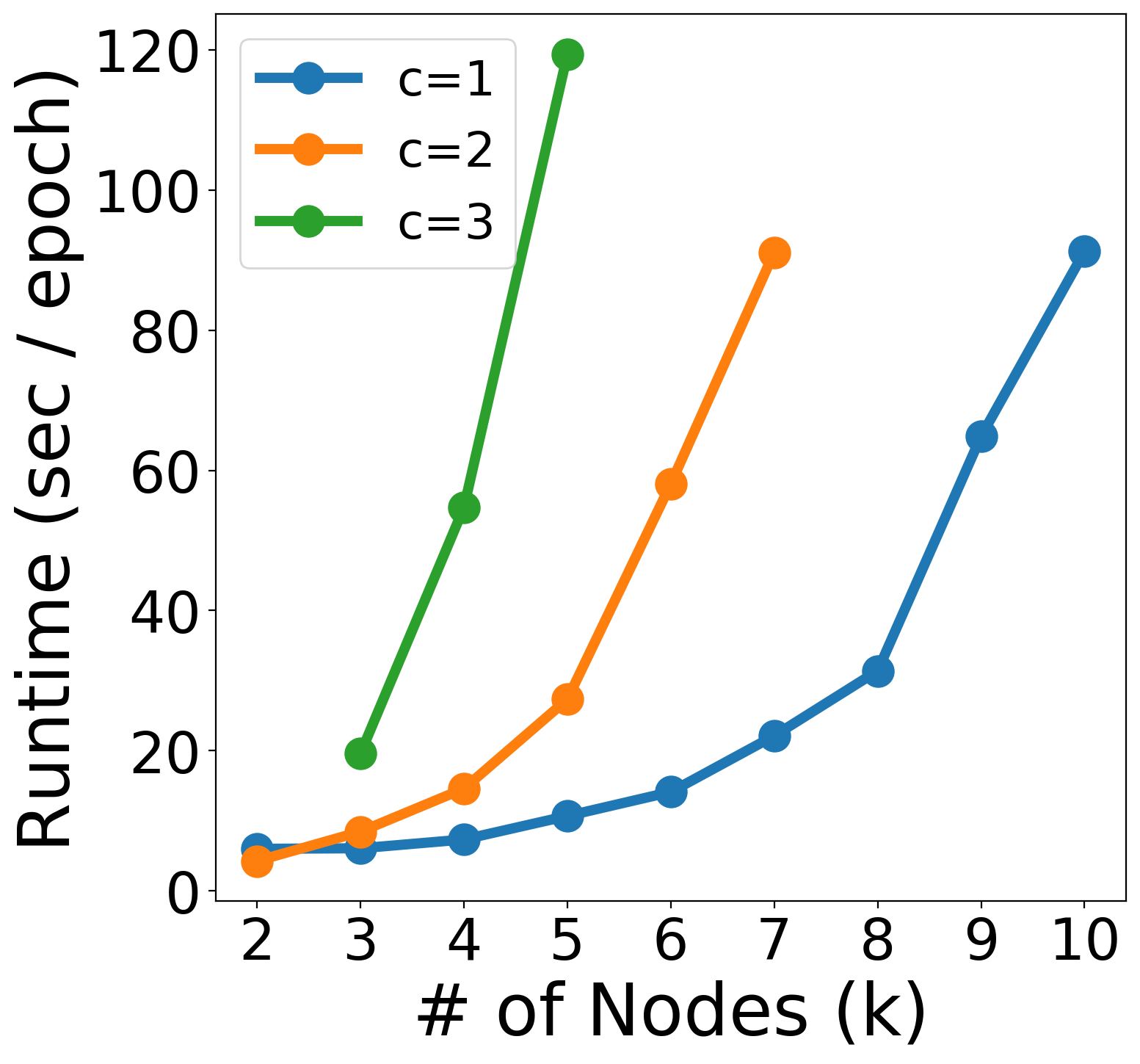}
	    \caption{Training time}
	    \label{fig:zinc_runtime}
	\end{subfigure}
	\caption{\methods's footprint scales practically with both $k$ and $c$ in memory (a) and running time (b) -- results on ZINC-12K.  %Solid blue, orange and green lines track scaling as $k$ increases, when running \method on the ZINC-12K dataset with $c=1$, 2 and 3 respectively.
	}
	\label{fig:zinc_scaling}
	\vspace{-0.15in}
\end{wrapfigure}
We next investigate how increasing $k$ and $c$ change the computational footprint of %running 
\methods in practice.  
Fig. \ref{fig:zinc_scaling} shows that increasing these parameters expectedly increases both memory consumption (in MB in (a)) as well as runtime (in seconds per epoch in (b)).  Notably, since larger $k$ and $c$ increases our model's expressivity, we observe that suitable choices for $k$ and $c$ allow us to practically realize these increasingly expressive models on commodity hardware. With conservative values of $c$ (e.g. $c$$=$$1$), we are able to consider passing messages between sets of $k$ (e.g. 10) nodes far larger than $k$-WL-style higher order models can achieve ($\leq$3).

\section{Conclusion}
%Our work is motivated by the expressiveness limitations of MPNNs, and the growing area of work in improving GNN expressiveness by following the ``coarse'' but established $k$-WL GI testing hierarchy.  Namely, increasing $k$ beyond modest values is computationally extremely challenging, hence modern $k$-WL equivalent GNN models largely achieve expressiveness only in theory but are unable to do so in practice.  At the same time, it is questionable whether such massive increases in expressiveness are practically required for most graph machine learning tasks.  To well-characterize these tradeoffs and develop practical models with more suited expressivity-complexity tradeoffs, a more ``fine-grained'' hierarchy and associated models are desirable.

Our work is motivated by the impracticality of higher-order GNN models based on the $k$-WL hierarchy, which make it challenging to study how much expressiveness real-world tasks truly necessitate. %benefit from. 
To this end, we proposed $(k,c)(\leq)$-SetWL,  a more practical and progressively-expressive hierarchy with theoretical connections to $k$-WL and drastically lowered complexity.  We also designed and implemented a practical model $(k,c)(\leq)$-SetGNN$(^*)$,
expressiveness of which is gradually increased by larger $k$ and $c$.
Our model
achieves strong performance, including several
new best results on graph-level tasks like ZINC-12K and expressiveness-relevant tasks like substructure counting, while being practically trainable on commodity hardware. % with $k$ as large as 10. 

%which is more practical, and yet capable of achieving strong performance in  with much lowered complexity compared to SOTA highly-expressive GNNs.  Our model and hierarchy are progressively expressive in $k$ and $c$, and achieves new best results on several graph-level tasks including ZINC-12K and substructure counting, while being practically trainable on commodity hardware. 
%$k$-WL model instantiations. % \ns{can be made more precise}.

\clearpage
\bibliographystyle{abbrvnat}
\bibliography{reference}

%%%%%%%%%%%%%%%%%%%%%%%%%%%%%%%%%%%%%%%%%%%%%%%%%%%%%%%%%%%%
% \clearpage
% \section*{Checklist}
% \input{99-Checklist}
%%%%%%%%%%%%%%%%%%%%%%%%%%%%%%%%%%%%%%%%%%%%%%%%%%%%%%%%%%%%

\appendix
\clearpage
\section{Appendix}

\subsection{Visualization Supergraph Connection of 2-SetWL}
\begin{figure}[h]
    \centering
    \includegraphics[width=\textwidth]{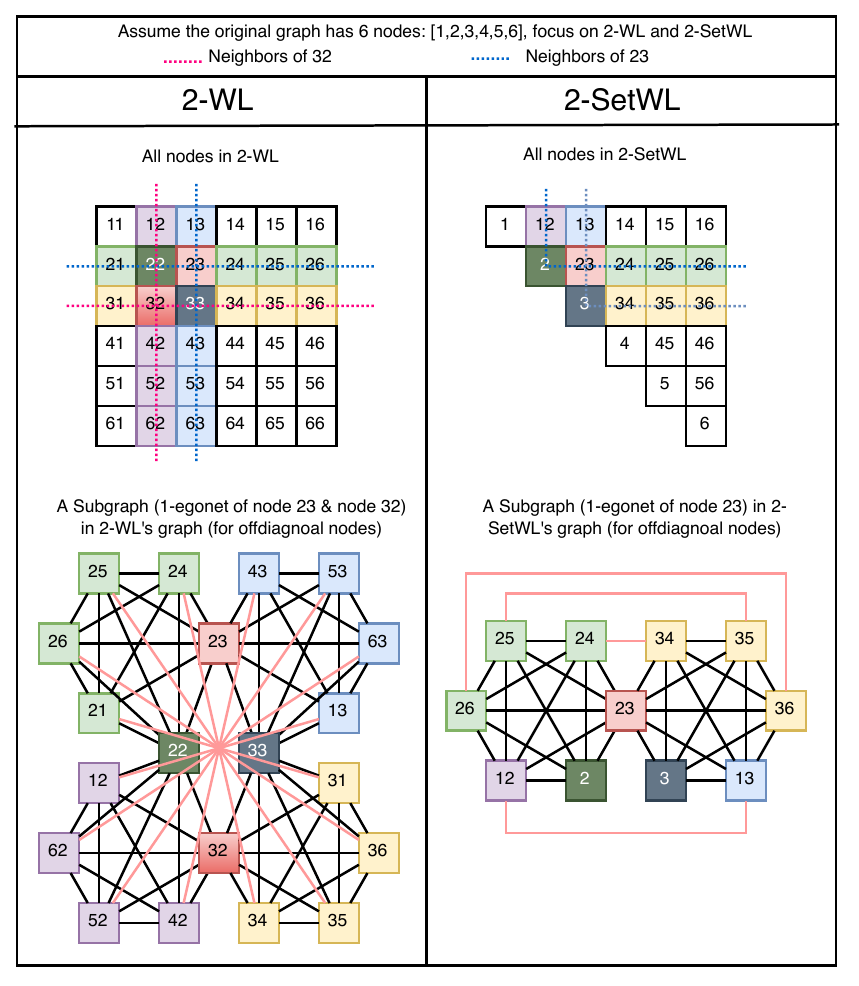}
    \caption{Comparison between 2-SetWL and 2-WL in their supergraph connection}
    \label{fig:subgraph_setgnn}
\end{figure}

\subsection{Bijective $k$-Pebble Game for $k$-WL}\label{apdx:wl-pebble}
The pebble game characterization of k-FWL appeared in \cite{cai1992optimal}. We use the pebble game defined in \cite{grohe2015pebble} for k-WL. Let $G,G'$ be graphs and $\tps{v}_0,\tps{v}_0'$ be $k$-tuple. Then the bijective $k$-pebble game $\text{BP}_k$($G,\tps{v}_0, G', \tps{v}'_0$) is defined as the follows. The game is played by two players Player I and Player II. The game proceeds in rounds starting from initial position $(\tps{v}_0,\tps{v}_0')$ and continues to new position $(\tps{v}_t,\tps{v}_t')$ as long as after $t$ rounds $\tps{v}_t[i] \mapsto \tps{v}'_t[i]$ defines an isomorphism between $G[\tps{v}_t]$ and $G'[\tps{v}'_t]$. 
 If the game has not ended after $t$ rounds, Player II wins the $t$-step $\text{BP}_k$($G,\tps{v}_0, G', \tps{v}'_0$), otherwise Player I wins.
 
 The $t$-th round is played as follows. 
 \begin{enumerate}
	\item Player I picks up the $i$-th pair of pebbles with $i\in [k]$.
	\item Player II chooses bijection $f : V(G) \to V(G') $.
	\item Player I chooses $x \in V(G)$.
	\item The new position is ($\tps{v}_t[x / i]$, $\tps{v}'_t[f(x) / i]$). If $G[\tps{v}_t[x / i]]$ and $G[\tps{v}'_t[f(x) / i]]$ are still isomorphic, the game continues. 
	Otherwise, the game ends and Player II loses. 
\end{enumerate}

The game $\text{BP}_k$($G,\tps{v}_0, G', \tps{v}'_0$) has the same expressivity as $k$-WL in distinguishing ($G,\tps{v}_0$) and ($G', \tps{v}'_0$). 

\begin{theorem}\label{thm:bp_v}
$k$-WL cannot distinguish ($G,\tps{v}_0$) and ($G', \tps{v}'_0$) at step $t$, i.e. $\bm{wl}_k^{(t)}(G, \tps{v}_0) = \bm{wl}_k^{(t)}(G', \tps{v}'_0)$  , if and only if Player II has a winning strategy for $t$-step $\text{BP}_k$($G,\tps{v}_0, G', \tps{v}'_0$).  
\end{theorem}

There is an extension of the pebble game, $\text{BP}_k$($G, G'$), without specifying the starting position. Specifically, at the beginning of the game, Player II is first asked to provide a bijection $g: V(G)^k \to V(G')^k.$ Then Player I chooses $\tps{v}_0 \in V(G)^k$. Then Player I and Player II start to play $\text{BP}_k$($G,\tps{v}_0, G', g(\tps{v}_0)$).

\begin{theorem}\label{thm:bp_g}
$k$-WL cannot distinguish G and $G'$ at step $t$, i.e. $\bm{gwl}_k^{(t)}(G) = \bm{gwl}_k^{(t)}(G')$  , if and only if Player II has a winning strategy for $t$-step $\text{BP}_k$($G, G')$.  
\end{theorem}

The proof of Theorem \ref{thm:bp_v} and Theorem \ref{thm:bp_g} can be found in Hella's work \cite{hella}. 

\subsection{Doubly Bijective $k$-Pebble Game for $k$-MultisetWL}\label{apdx:mwl-pebble}
% \revise{\kmwl pebble game}

Let $G,G'$ be graphs and $\ms{v}_0,\ms{v}'_0$ be $k$-multisets. We adapt the $\text{BP}_k$ game for \kmwl and call it doubly bijective $k$-pebble game, i.e. $\text{DBP}_k(G, \ms{v}_0, G', \ms{v}'_0 )$. A similar version of the pebble game for \kmwl was suggested by Grohe in \cite{grohep}.
% We acknowledge that we have discuss with Martin Grohe for the pebble game characterization.  

The $\text{DBP}_k(G, \ms{v}_0, G', \ms{v}'_0 )$ is defined as the follows. The game starts at the position ($\ms{v}_0,\ms{v}'_0$). 

Let the current position be $(\ms{v}_t,\ms{v}'_t)$. The $t$-th round is played as follows.
\begin{enumerate}
	\item Player II chooses a bijection $h: \ms{v}_t \to \ms{v}'_t $
	\item Player I chooses the $y \in \ms{v}_t$ 
% 	\begin{itemize}
% 		\item Compare isomorphism types of $G[(v_t,x)]$ and $G'[(v_t',f(x))]$. If they are not equal, the game ends and Spoiler wins. Note that different multiplicities of elements in $v_t$ and $v_t'$ can also be distinguished in their induced subgraphs on $G$ and $G'$.  
% 	\end{itemize}
	\item Player II chooses bijection $f: V(G) \to V(G')$
	\item Player I chooses $x\in V(G)$
	\item The new position is $(\ms{v}_t[x/\text{idx}_{\ms{v}_t}(y)], \ms{v}'_t[f(x)/\text{idx}_{\ms{v}'_t}(h(y))])$ and the game continues, if $G[\ms{v}_t[x/\text{idx}_{\ms{v}_t}(y)]]$ and $G'[\ms{v}'_t[f(x)/\text{idx}_{\ms{v}'_t}(h(y))]]$ are isomorphic. Otherwise the game ends and Player II loses. 
\end{enumerate}

To extend the game to the setting of no starting position, we define the corresponding game $\text{DBP}_k(G, G')$. Simliar to $\text{BP}_k(G, G')$, at the beginning Player II is asked to pick up a bijection $g: \text{Multiset}(V(G)^k) \to \text{Multiset}(V(G')^k)$. Then Player I picks up a $\ms{v}_0$ and the game $\text{DBP}_k(G, \ms{v}_0, G', g(\ms{v}_0))$ starts.

\begin{customthm}{3}
 \kmwl has the same expressivity as the doubly bijective $k$-pebble game.
\end{customthm}

\begin{proof}
Specifically, given graph $G$ with $k$-multiset $\ms{v}$ and graph $G'$ with $k$-multiset $\ms{v}'$, we are going to prove that $\bm{mwl}_k^{(t)}(G, \ms{v}) = \bm{mwl}_k^{(t)}(G', \ms{v}') \Longleftrightarrow $ Player II has a winning strategy for $t$-step game $\text{DBP}_k(G, \ms{v}, G', \ms{v'})$.

We now prove it by induction on $t$. When $t=0$, it's obvious that the statement is correct, as $\bm{mwl}_k^{(0)}(G, \ms{v}) = \bm{mwl}_k^{(0)}(G', \ms{v}')$ is equivalent to $G[\ms{v}]$ and $G'[\ms{v}']$ being isomorphic, which implies that Player II can start the game without losing. Now assume that for $t\le l$ the statement is correct. 
For step $t=l+1$, let's first prove left $\Longrightarrow$ right. 

 $\bm{mwl}_k^{(l+1)}(G, \ms{v}) = \bm{mwl}_k^{(l+1)}(G', \ms{v}')$. By Eq.\eqref{eq:kmwl} we know this is equivalent to \\
(1) $\bm{mwl}_k^{(l)}(G, \ms{v}) = \bm{mwl}_k^{(l)}(G', \ms{v}')$ \\
(2)  $\exists \text{ bijective } h: \ms{v}   \to \ms{v}'$, 
$\forall y \in \ms{v}$, $\mms{\bm{mwl}_k^{(l)}(G, \ms{v}[x/\text{idx}_{\ms{v}}(y)])\  \big|\ x \in V(G)} = \mms{\bm{mwl}_k^{(l)}(G', \ms{v}'[x/\text{idx}_{\ms{v}'}(h(y))])\ \big|\ x \in V(G')} $. 

Now let's start the $\text{DBP}_k$ game at position ($\ms{v}, \ms{v}'$). By (2) we know that there exist a $h$ satisfying (2). Then at the first round, we as Player II pick the $h$ as the bijection. Next Player I will choose a $y \in \ms{v}$. According to (2), for the $h$ and $y$ we have $\mms{\bm{mwl}_k^{(l)}(G, \ms{v}[x/\text{idx}_{\ms{v}}(y)])\  \big|\ x \in V(G)} = \mms{\bm{mwl}_k^{(l)}(G', \ms{v}'[x/\text{idx}_{\ms{v}'}(h(y))])\ \big|\ x \in V(G')}$. This implies that there exists a bijection $f: V(G) \to V(G')$ such that $\forall  x \in  V(G), \bm{mwl}_k^{(l)}(G, \ms{v}[x/\text{idx}_{\ms{v}}(y)]) = \bm{mwl}_k^{(l)}(G, \ms{v}'[f(x)/\text{idx}_{\ms{v}'}(h(y))])$. Hence let Player II pick the $f$. Now player I will choose a $x\in V(G)$. Then $\bm{mwl}_k^{(l)}(G, \ms{v}[x/\text{idx}_{\ms{v}}(y)]) = \bm{mwl}_k^{(l)}(G, \ms{v}'[f(x)/\text{idx}_{\ms{v}'}(h(y))])$ implies $\bm{mwl}_k^{(0)}(G, \ms{v}[x/\text{idx}_{\ms{v}}(y)]) = \bm{mwl}_k^{(0)}(G, \ms{v}'[f(x)/\text{idx}_{\ms{v}'}(h(y))])$, hence $G[\ms{v}_t[x/\text{idx}_{\ms{v}_t}(y)]]$ and $G'[\ms{v}'_t[f(x)/\text{idx}_{\ms{v}'_t}(h(y))]]$ are isomorphic. So the game doesn't end. At the next round, the game starts at the position  $(\ms{v}[x/\text{idx}_{\ms{v}}(y)], \ms{v}'[f(x)/\text{idx}_{\ms{v}'}(h(y))])$,
 and we know $\bm{mwl}_k^{(l)}(G, \ms{v}[x/\text{idx}_{\ms{v}}(y)]) = \bm{mwl}_k^{(l)}(G, \ms{v}'[f(x)/\text{idx}_{\ms{v}'}(h(y))])$. By the inductive hypothesis, Player II has a strategy to play DBP$_k$ at the new position for $l$ rounds. Hence Player II has a winning strategy to play  DBP$_k$ at original position ($\ms{v} , \ms{v}'$) for $l+1$ rounds. 
 
 Next we prove right $\Longrightarrow$ left by showing that  $\bm{mwl}_k^{(l+1)}(G, \ms{v}) \neq \bm{mwl}_k^{(l+1)}(G', \ms{v}') \Longrightarrow $ Player I has a winning strategy for $l+1$-step game $\text{DBP}_k(G, \ms{v}, G', \ms{v'})$.
 
 $\bm{mwl}_k^{(l+1)}(G, \ms{v}) \neq \bm{mwl}_k^{(l+1)}(G', \ms{v}')$ implies:\\
 (1) $\bm{mwl}_k^{(l)}(G, \ms{v}) \neq \bm{mwl}_k^{(l)}(G', \ms{v}')$ \\
 or 
 (2) for any bijection $h: \ms{v} \to \ms{v}'$, $\exists y \in \ms{v}$, such that 
 $\mms{\bm{mwl}_k^{(l)}(G, \ms{v}[x/\text{idx}_{\ms{v}}(y)])\  \big|\ x \in V(G)} \neq \mms{\bm{mwl}_k^{(l)}(G', \ms{v}'[x/\text{idx}_{\ms{v}'}(h(y))])\ \big|\ x \in V(G')}$\\
 
 If (1) holds, then by induction we know that Player I has a winning strategy within $l$-steps hence Player I also has a winning strategy within $l+1$ steps. If (2) holds, then at the first round after Player II picks up a bijection $h$, Player I can choose the specific $y \in \ms{v}$ with $\mms{\bm{mwl}_k^{(l)}(G, \ms{v}[x/\text{idx}_{\ms{v}}(y)])\  \big|\ x \in V(G)} \neq \mms{\bm{mwl}_k^{(l)}(G', \ms{v}'[x/\text{idx}_{\ms{v}'}(h(y))])\ \big|\ x \in V(G')}$. Then no matter which bijection $f:V(G) \to V(G')$ Player II chooses, Player I can always choose a $x \in V(G)$ such that $\bm{mwl}_k^{(l)}(G, \ms{v}[x/\text{idx}_{\ms{v}}(y)]) \neq \bm{mwl}_k^{(l)}(G', \ms{v}'[x/\text{idx}_{\ms{v}'}(h(y))])$. Then by induction, Player I has a winning strategy within $l$-steps for the DBP$_k$ starts at position $(\ms{v}[x/\text{idx}_{\ms{v}}(y)],  \ms{v}'[x/\text{idx}_{\ms{v}'}(h(y))])$. Hence even if Player I doesn't win in the first round, he/she can still win in $l+1$ rounds. 
 
 Combining both sides we know that the equivalence between $t$-step DBP$_{k}$ and $t$-step \kmwl holds for any $t$ and any $k$-multisets.

\end{proof}

\begin{theorem}\label{thm:dbp_g}
\kmwl cannot distinguish G and $G'$ at step $t$, i.e. $\bm{gmwl}_k^{(t)}(G) = \bm{gmwl}_k^{(t)}(G')$  , if and only if Player II has a winning strategy for $t$-step $\text{DBP}_k$($G, G')$.  
\end{theorem}

\begin{proof}
The proof is strict forward with using the proof inside Theorem \ref{dbp_mwl}. We just need to show that the pooling of all $k$-multisets representations is equivalent to Player II finding a bijection $g: \text{Multiset}(V(G)^k) \to \text{Multiset}(V(G')^k)$ at the first step of the game. We omit that given its simplicity. 
\end{proof}

% \revise{\kswl pebble game}

% Let $G,G'$ be graphs and $v_0,v_0'$ be $m_0$-sets, for $m_0 \in [k]$. The $(\leq) k$-SetWL game is played by Spoiler and Duplicator. Positions are $m_t$-sets $(v_t,v_t')$, for $m_t \in [k]$. The game proceeds in rounds starting from initial position $(v_0,v_0')$ and continues as long as after $t$ rounds the subgraphs induced by the current position are isomorphic. If the game ends after a finite number of rounds Spoiler wins, otherwise Duplicator wins.
	
% Let the current position be $(v_t,v_t')$. A round is played as follows.
% \begin{enumerate}
% 	\item Duplicator chooses bijection $f : V \to V'$
% 	\item Spoiler chooses pair $(x,f(x))$
% 	\begin{itemize}
% 		\item Compare isomorphism types of $G[(v_t,x)]$ and $G'[(v_t',f(x))]$. If they are not equal, the game ends and Spoiler wins.
% 	\end{itemize}
% 	\item Duplicator chooses bijective correspondence $g : v_t \to v_t'$
% 	\item Spoiler chooses $y\in v_t$
% 	\item if $m_t < k$: Spoiler decides to continue with
% 	\begin{itemize}
% 		\item position ${(v_t\cup \{x\},v_t'\cup \{f(x)\})}$, or
% 		\item position ${(v_t\setminus \{y\} \cup \{x\},v_t'\setminus \{g(y)\} \cup \{f(x)\})}$
% 	\end{itemize}
% 	\item if $m_t = k$: continue with position ${(v_t\setminus \{y\} \cup \{x\},v_t'\setminus \{g(y)\} \cup \{f(x)\})}$
% \end{enumerate}

\subsection{Proofs of Theorems}\label{apdx:proofs}
\subsubsection{Bound of Summation of Binomial Coefficients}
Derivation of: 
\begin{align}
    \sum_{i=1}^k \binom{n}{i} \le \binom{n}{k}\frac{n-k+1}{n-2k+1}
\end{align}

\begin{proof}
\begin{align}
   \frac{\sum_{i=1}^k \binom{n}{i}}{\binom{n}{k}} & = \frac{\binom{n}{k} + \binom{n}{k-1} + \binom{n}{k-2} +...}{\binom{n}{k}} \\
   & = 1 + \frac{k}{n-k+1} + \frac{k(k-1)}{(n-k+1)(n-k+2) } + ... \\
   & \le 1 + \frac{k}{n-k+1} + (\frac{k}{n-k+1})^2 + ...  \\
   & \le \frac{n-k+1}{n-2k+1}
\end{align}
\end{proof}

\subsubsection{Proof of Theorem 1}

\begin{customthm}{1}
 Let $k\ge 1$ and $\bm{wl}_k^{(t)}(G, \ms{v}) := \mms{\bm{wl}_k^{(t)}(G, p(\ms{v})) | p \in \text{perm[k]}}$.
 For all $t\in \mathbb{N}$ and all graphs $G, G'$:
 %\cbit
%\item[(1)] 
 \kmwl is upper bounded by $k$-WL in distinguishing multisets $G, \ms{v}$ and $G', \ms{v}'$ at $t$-th iteration, i.e.
$\bm{wl}_k^{(t)}(G, \ms{v}) = \bm{wl}_k^{(t)}(G', \ms{v}')$  $\Longrightarrow$ $\bm{mwl}_k^{(t)}(G, \ms{v}) = \bm{mwl}_k^{(t)}(G', \ms{v}')$. 
\end{customthm}

\begin{proof}
By induction on $t$. It's obvious that when $t=0$ the above statement hold, as both side are equivalent to $G[
\ms{v}]$ and $G'[\ms{v}']$ being isomorphic to each other. Assume $\leq t$ the above statement is true. 
For $t+1$ case, by definition the left side is equivalent to $ \mms{\bm{wl}_k^{(t+1)}(G, p(\ms{v})) | p \in \text{perm[k]}} =  \mms{\bm{wl}_k^{(t+1)}(G', p(\ms{v}')) | p \in \text{perm[k]}}$. Let $\tps{v}$ be the ordered version of $\ms{v}$ following canonical ordering over $G$, then there exists a bijective mapping $f$ between $\ms{v}$ and $\ms{v}'$, such that $\bm{wl}_k^{(t+1)}(G, \tps{v}) = \bm{wl}_k^{(t+1)}(G', f(\tps{v}))$, where $f(\tps{v}) := (f(\tps{v}_1),...,f(\tps{v}_k))$. By \cite{cai1992optimal}'s Theorem 5.2, for any $t$, $\bm{wl}_k^{(t)}(G, \tps{v}) = \bm{wl}_k^{(t)}(G', f(\tps{v}))$ is equivalent to that player II has a winner strategy for $t$-step pebble game with initial configuration $G,\tps{v}$ and $G', f(\tps{v})$ (please refer to \cite{cai1992optimal} for the description of pebble game). Notice that applying any permutation to the pebble game's initial configuration won't change having winner strategy for player II, hence we know that $\forall p \in \text{perm[k]}$, $\bm{wl}_k^{(t+1)}(G, p(\tps{v})) = \bm{wl}_k^{(t+1)}(G', p(f(\tps{v})))$. Now applying Eq.\eqref{eq:kwl}, we know that (1) $\bm{wl}_k^{(t)}(G, p(\tps{v})) = \bm{wl}_k^{(t)}(G', p(f(\tps{v})))$, and (2)$\forall i \in [k]$, $\mms{\bm{wl}_k^{(t)}(G, p(\tps{v}[x/i]))\ |\ x \in V(G) }$ = $\mms{\bm{wl}_k^{(t)}(G', p(f(\tps{v})[x/i]))\ |\ x \in V(G') }$. We rewrite (2) as,  $\forall i \in [k]$, $\exists \text{ bijective } h_i:V(G)\rightarrow V(G')$, $\forall x \in V(G)$, $\bm{wl}_k^{(t)}(G, p(\tps{v}[x/i])) = \bm{wl}_k^{(t)}(G', p(f(\tps{v})[h_i(x)/i])) $. As (1) and (2) hold for any $p\in \text{perm[k]}$ , now applying induction hypothesis to both (1) and (2), we can get (a) $\bm{mwl}_k^{(t)}(G, \ms{v}) = \bm{mwl}_k^{(t)}(G', \ms{v}')$, and (b) $\forall i \in [k]$, $\exists \text{ bijective } h_i:V(G)\rightarrow V(G')$, $\forall x \in V(G)$, $\bm{mwl}_k^{(t)}(G, \ms{v}[x/i]) = \bm{mwl}_k^{(t)}(G', \ms{v}'[h_i(x)/g(i)]) $, where $g:[k]\rightarrow [k] $ is the index mapping function corresponding to $f$. Now we rewrite (b) as $\exists g: [k] \rightarrow [k]$, $\forall i \in [k]$, $\mms{\bm{mwl}_k^{(t)}(G, \ms{v}[x/i])\ |\ x\in V(G)} = \mms{\bm{mwl}_k^{(t)}(G', \ms{v}'[x/g(i)])\ |\ x\in V(G')} $. Combining (a) and (b), using Eq.\eqref{eq:kmwl} we can get $\bm{mwl}_k^{(t+1)}(G, \ms{v}) = \bm{mwl}_k^{(t+1)}(G', \ms{v}')$. Thus for any $t$ the above statement is correct.
\end{proof}

\subsubsection{Proof of Theorem 2}
\textbf{CFI Graphs and Their Properties}

Cai, Furer and Immerman \cite{cai1992optimal} designed a construction of a series of pairs of graphs CFI(k) such that for any $k$, $k$-WL cannot distinguish the pair of graphs CFI$(k)$. We will use the CFI graph construction to prove the theorem. Here we use a variant of CFI construction that is proposed in Grohe et al.'s work \cite{grohe2015pebble}. Let $\calK$ he a complete graph with $k$ nodes. We first construct an enlarged graph $\mathcal{X}(\mathcal{K})$ from the base graph $\mathcal{K}$ by mapping each node and edge to a group of vertices and connecting all $(|V(\calK)| + |E(\calK)|)$ groups of vertices following certain rules. Notice that we use node for base graph and use vertex for the enlarged graph for a distinction. Let $vw$ denotes the edge connecting node $v$ and node $w$. For a node $v \in V(\calK)$, let $E(v):= \big\{vw| vw \in E(\calK) \big\}$ denotes the set of all adjacent edges of $v$ in the graph $\calK$. 

For every node $v \in V(\calK)$, we map node $v$ to a group of vertices $S_v:= \{v^X | X \subseteq E(v) \}$ with size  $|S_v| = 2^{deg(v)} = 2^{k-1}$. For every edge $e \in E(\calK)$, the construction maps $e$ to two vertices $S_{e}:= \{e^0, e^1\}$. Hence there are $2*|E(\calK)|+ |V(\calK)|*2^{k-1} = k(k-1)+k(2^{k-1})$ number of vertices in the enlarged graph $\calX(\calK)$ with $V(\calX(\calK)) = (\cup_{v\in{V(\calK)}}S_v) \cup (\cup_{e\in{E(\calK)}}S_e)$. Let $V^*:=\cup_{v\in{V(\calK)}}S_v$ and $E^*:=\cup_{e\in{E(\calK)}}S_e$. 

Now edges inside $\calX(\calK)$ are defined as follows
\begin{align}
    E(\calX(\calK)):= & \{ v^{X}e^1 \ |\ v\in V(\calK), X \subseteq E(v), \text{and } e\in X  \}  \ \cup \\
     & \{ v^{X}e^0 \ |\ v\in V(\calK), X \subseteq E(v), \text{and } e\not\in X  \} \cup \{ e^0e^1\ |\ e\in E(\calK)\}
\end{align}

\begin{figure}
    \centering
    \includegraphics{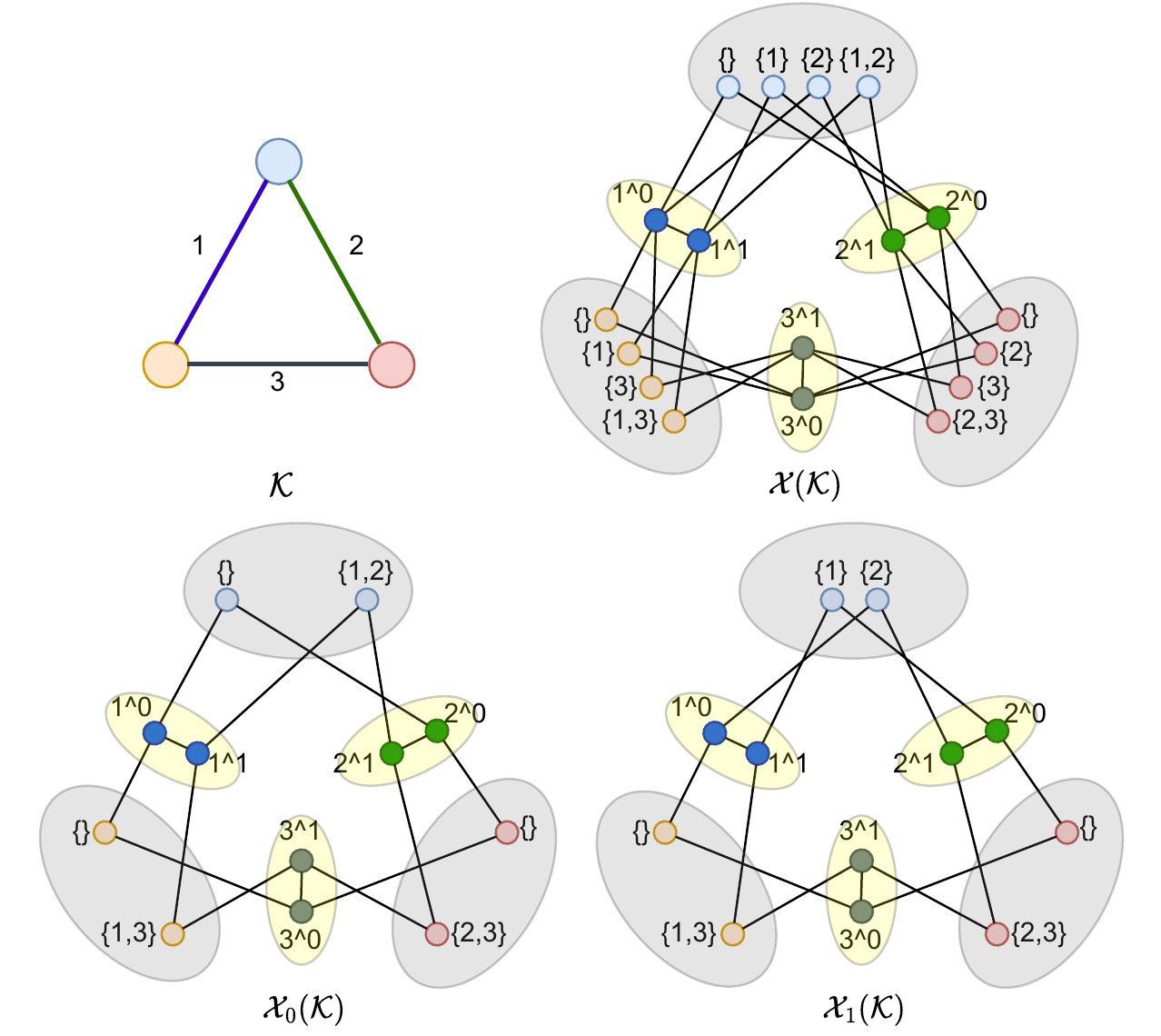}
    \caption{CFI$(k)$ construction visualization for $k=3$}
    \label{fig:cfi}
\end{figure}

What's more, we also color the vertices such that all vertices inside $S_v$ have color $v$ for every $v\in V(\calK)$ and all vertices inside $S_e$ have color $e$ for every $e\in E(\calK)$. See Figure.\ref{fig:cfi} top right for the visualization of transforming from base graph $\calK$ to $\calX(\calK)$ with $k=3$. 

There are several important properties about the automorphisms of $\calX(\calK)$. Let $h \in \text{Aut}(\calX (\calK))$ be an automorphism, then
\begin{enumerate}
    \item $h(S_v) = S_v$ and $h(S_e) = S_e$ for all $v\in V(\calK)$ and $e \in E(\calK)$.  
    \item For every subset $F\subseteq E(\calK)$, there is exactly one automorphism $h_F$ that flips precisely all edges in F, i.e. $h_F(e^0) = e^1$ and $h_F(e^1) = e^0$ if and only if $e\in F$. More specifically,
    \begin{itemize}
        \item $h_F(e^i) = e^{1-i},\  \forall e \in F$
        \item $h_F(e^i) = e^{i},\  \forall e \not \in F$
        \item $h_F(v^X) = v^Y$, $\forall v\in{E(\calK)}, X\subseteq E(v)$ \\
        where $Y:= X \triangle \big(F \cap E(v) \big) = \Big(X \setminus  \big(F \cap E(v) \big) \Big) \cup \Big( \big(F \cap E(v) \big) \setminus X \Big) $
    \end{itemize}
\end{enumerate}

\begin{proof}
These properties are not hard to prove. First for property 1, it is true based on the coloring rules of vertices in $\calX(\calK)$. Now for the second property. As $h_F$ flips precisely all edges in $F$, we have $h_F(e^i) = e^{1-i},\  \forall e \in F$ and $h_F(e^i) = e^{i},\  \forall e \not \in F$. Now let $h_F(v^X) = v^Y$ $\forall v\in{E(\calK)}, X\subseteq E(v)$, we need to figure out $Y$'s formulation. Let's focus on node $v$ without losing generality, and we can partition the set $E(v)$ to two parts: $E(v)\cap F$ and $E(v)\setminus F$. Then for any $e \in E(v) \cap F$, we know that $h_F(e^i) = e^{1-i}$. Let $v \leftrightarrow w$ denote that $v$ and $w$ are connected in $\calX(\calK)$. Then $\forall X \subseteq E(v)$, $ e\in X \Longleftrightarrow e^0 \leftrightarrow v^X \Longleftrightarrow h_F(e^0) \leftrightarrow h_F(v^X) \Longleftrightarrow e^1 \leftrightarrow v^Y \Longleftrightarrow e \not \in Y$. And similarly we have $\forall X \subseteq E(v)$,  $ e \not \in X \Longleftrightarrow e \in Y$. Hence $ \forall e \in  E(v)\cap F, \forall X \subseteq E(v)$, $e \in X \triangle Y$. This implies that $E(v)\cap F \subseteq X \triangle Y$. Following the same logic we can also get $ \forall e \in  E(v)\setminus F, \forall X \subseteq E(v)$, $e \in E(v) \setminus ( X \triangle Y) $, which is equivalent to $E(v)\setminus F \subseteq E(v) \setminus ( X \triangle Y)$, which further implies that $E(v)\cap F  \supseteq X \triangle Y $ as $X \triangle Y \subseteq E(v)$. Then combining both side we know that $\forall v \in E(\calK), \forall X \subseteq E(v), E(v)\cap F  = X \triangle Y $. Hence we get $Y =  X \triangle (X\triangle Y) = X \triangle \big( F\cap E(v) \big)$.
\end{proof}

In the proof we can also know another important property. That is $\forall v, X\subseteq E(v), X\triangle h_F(X) = E(v) \cap F $ is constant for any input $X\in E(v)$. 

Now we are ready to construct variants of graphs that are not isomorphic from the enlarged graph $\calX(\calK)$. Now let T be a subset of $V(\calK)$. Now we define an induced subgraph $\calX_T(\calK)$ of the enlarged graph $\calX(\calK)$. Specifically, we define the new node group as follows 
\begin{align}
    S_v^T := 
    \begin{cases}
        \{v^X \in S_v \ |\ |X|  \equiv 0\mod2 \} \text{ if } v \not \in T\\
        \{v^X \in S_v \ |\ |X|  \equiv 1\mod2 \} \text{ if } v \in T
    \end{cases}
\end{align}

Then the induced subgraph is defined as $\calX_T(\calK) := \calX(\calK)[ (\cup_{v\in{V(\calK)}}S^T_v) \cup E^*]$. In Figure \ref{fig:cfi} we show  $\calX_{\emptyset}(\calK)$ in bottom left (labeled with $\calX_{0}(\calK)$) and $\calX_{\{v_1 \}}(\calK)$ in the bottom right (labeled with $\calX_{1}(\calK)$).  

\begin{lemma}\label{lma: TU_isomorphism}
For all $T, U \subseteq V(\calK)$, $\calX_{T}(\calK) \cong \calX_{U}(\calK)$ if and only if $|T| \equiv |U| \mod 2 $. And if they are isomorphic, one isomorphism between  $\calX_{T}(\calK)$ and $\calX_{U}(\calK)$ is $h_F$ with $F= E( \calK[T\triangle U])$, where $E(\calK[T\triangle U])$ denotes the set of all edges $\{v_i,v_j\} \subseteq T\triangle U$.
\end{lemma}

Notice that $h_F$ is an automorphism for $\calX(\calK)$, but with restricting its domain to $ (\cup_{v\in{V(\calK)}}S^T_v) \cup E^*$ it becomes the isomorphism between $\calX_T(\calK)$ and $\calX_U(\calK)$.

The proof of the lemma can be found in \cite{cai1992optimal} and \cite{grohe2015pebble}. With this lemma we know that $\calX_{\{v_1 \}}(\calK)$ and $\calX_{\emptyset}(\calK)$ are not isomorphic. In next part we show that $\calX_{\emptyset}(\calK)$ and $\calX_{\{v_1 \}}(\calK)$ cannot be distinguished by $(k-1)$-WL but can be distinguished by \kmwl, thus proving Theorem \ref{thm:cfi-mwl}.

\textbf{Main Proof}

\begin{customthm}{2}
    \kmwl is no less powerful than  ($k$-$1$)-WL in distinguishing graphs: for $k\ge 3$ there is a pair of graphs that can
    be distinguished by \kmwl but not by ($k$-$1$)-WL.
\end{customthm}

\begin{proof}
To prove the theorem, we show that for any $k\ge 3$, $\calX_{\{v_1 \}}(\calK)$ and $\calX_{\emptyset}(\calK)$, defined previously, are two nonisomorphic graphs that can be distinguished by \kmwl but not by $(k-1)$-WL. It's well known that these two graphs cannot be distinguished by $(k-1)$-WL, and one can refer to Theorem 5.17 in \cite{grohe2015pebble} for its proof. Now to prove these two graphs can be distinguished by \kmwl, using Theorem \ref{thm:dbp_g} we know it's equivalent to show that Player I has a winning strategy for the doubly bijective $k$-pebble game $\text{DBP}_k(\calX_{\emptyset}(\calK), \calX_{\{v_1 \}}(\calK))$. 

At the start of the pebble game $\text{DBP}_k(\calX_{\emptyset}(\calK), \calX_{\{v_1 \}}(\calK))$, Player II is asked to provide a bijection between all $k$-multisets of $\calX_{\emptyset}(\calK)$ to all $k$-multisets of $ \calX_{\{v_1 \}}(\calK)$. For any bijection the Player II chosen, the Player I can pickup $\ms{v}_0= \mms{v_1^{\emptyset},  v_2^{\emptyset},...,  v_k^{\emptyset}}$, with corresponding position $\ms{v}'_0 = \mms{v_1^{X_1},  v_2^{X_2},...,  v_k^{X_k}}$. Notice that for a position $(\ms{x}, \ms{y}):= (\mms{x_1,...,x_k}, \mms{y_1,...,y_k})$, the position is called \textit{consistent} if there exists a $F\subseteq E(\calK) $, such that $h_F(\ms{x}) = \mms{h_F(x_1),...,h_F(x_k)} = \ms{y}$. One can show that Player I can easily win the game $\text{DBP}_k(\calX_{\emptyset}(\calK), \calX_{\{v_1 \}}(\calK))$ after one additional step if the current position $(\ms{x}, \ms{y})$ is not consistent, even if $\calX_{\emptyset}(\calK)[\ms{x}]$ and $\calX_{\{v_1\}}(\calK)[\ms{y}]$ are isomorphic \cite{grohe2015pebble}.

We claim that for any $k$-multiset $\ms{v}'_0 = \mms{v_1^{X_1},  v_2^{X_2},...,  v_k^{X_k}}$ the position $(\ms{v}_0,\ms{v}'_0)$ cannot be consistent, and thus Player II loses $\text{DBP}_k(\calX_{\emptyset}(\calK), \calX_{\{v_1 \}}(\calK))$ after the first round. Assume that there exists a subset $F\subseteq E(\calK)$, such that $h_F(\ms{v}_0) = \ms{v}'_0$. By property 1 of automorphisms of $\calX(\calK)$ we have that $h_F(S_{v_1}) = S_{v_1}$, and since $\ms{v}_0$ contains exactly one vertex of each color, it also holds that $h_F(v_1^{\emptyset}) = v_1^{X_1}$. It follows from property 2 that $X_1 = \emptyset \triangle \big(F \cap E(v_1) \big) = F \cap E(v_1)$. Based on the definition of $\calX_{\{v_1\}}(\calK)$, we know that $|X_1|\equiv 1 \mod 2$, and thus $F$ flips an odd number of neighbors of $v_1$, i.e. $|F \cap E(v_1)|\equiv 1 \mod 2$. Similarly, we have that $h_F(S_{v_i}) = S_{v_i}$ and $|X_i| \equiv 0 \mod 2, \forall i \ge 2$, hence $|F \cap E(v_i)|\equiv 0 \mod 2$. It follows from a simple handshake argument that there exists an $m\geq 0$ such that

$$2m = \sum_{v\in V}|(F \cap E)(v)| = \sum_{i\in [k]\setminus \{1\}}|F \cap E(v_i)| + |F \cap E(v_1)| \equiv 1 \mod 2,$$

which is a contradiction. Hence Player I has a winning
strategy for $\text{DBP}_k(\calX_{\emptyset}(\calK), \calX_{\{v_1 \}}(\calK))$.

%Hence we can assume that Player II picks a good bijection such that there exists a $F$ such that $h_F(v_i^{\emptyset}) = v_i^{X_i}, \forall i \in [k]$. And equivalently, $X_i = \emptyset \triangle (F\cap E(v)) = F \cap E(v), \forall i \in [k]$. Based on the definition of $\calX_{\emptyset}(\calK)$, we know that for any edge $e_{ij} \in E(\calK)$, $e_{ij}^0\leftrightarrow v_i^{\emptyset}$ and $e_{ij}^0\leftrightarrow v_j^{\emptyset}$. Also we know that $e_{ij} \in E(i)$ and $e_{ij} \in E(j)$, hence $e_{ij} \in E() $

% We have discussed properties of global automorphisms
% of $\calX(\calK)$. We need shift the focus to isomorphism between $\calX_T(\calK)$ and $\calX_U(\calK)$ as to prove the targeting theorem we will wor

% and isomorphisms. In the game setting, Players are asked to maintain a local isomorphism along the game until it ends. Let $g$ be a local isomorphism with 

\end{proof}
\subsubsection{Proof of Theorem 3}
See Appendix.\ref{apdx:mwl-pebble}.

\subsubsection{Proof of Theorem 4}

\begin{customthm}{4}
Let $k\ge 1$, then $\forall t\in \mathbb{N}$ and all graphs $G, G'$:
$\bm{mwl}_k^{(t)}(G, \s{v}) = \bm{mwl}_k^{(t)}(G', \s{v}') \Longrightarrow \bm{swl}_k^{(t)}(G, \s{v}) = \bm{swl}_k^{(t)}(G', \s{v}')$. 
\end{customthm}

\begin{proof}
% We will first prove the following Lemma.
% \begin{lemma}
    
% \end{lemma}
\textbf{Notation:} Let $s(\cdot)$ transform a multiset to set by removing repeats, and let $r(\cdot)$ return a tuple with the number of repeats for each distinct element in a multiset.  Let $f$ be the inverse mapping such that $\ms{v} = f(s(\ms{v}), r(\ms{v}))$.

Define $F^{(t+1)}(G,G',\ms{v}, \ms{v}'):= \{\text{ injective } f:\ms{v} \rightarrow \ms{v}' \ | \ f\in F^{(t)}(G,G',\ms{v}, \ms{v}'), \text{ AND }, \forall y \in \ms{v}, \exists h_y:V(G) \rightarrow V(G'), \forall x, f \in F^{(t)}(G, G', \ms{v}[x / \text{idx}_{\ms{v}}(y)], \ms{v}'[h_y(x) / \text{idx}_{\ms{v}'} (f(y)) ] )   \}$. 
% \lz{ Let $H^{(t)}(G,G', \ms{v}, \ms{v}' ):= \{p: \ms{v}\rightarrow \ms{v}' \ |\  \bm{wl}_k^{(t)}(G, \tps{v}) = \bm{wl}_k^{(t)}(G', p(\tps{v}) )  \}$, 
% then can we prove that $H^{(t)}(G,G', \ms{v}, \ms{v}' ) = F^{(t)}(G,G', \ms{v}, \ms{v}' )$ ?}
Let $F^{(0)}(G,G',\ms{v}, \ms{v}'):= \{f \ | \ f \text{ is an isomorphism from } G[\ms{v}] \text{ to }  G'[\ms{v}']   \} $.

% \begin{lemma}
% $\forall t$, $\bm{mwl}_k^{(t)}(G, \s{v}) = \bm{mwl}_k^{(t)}(G', \s{v}') \Longleftrightarrow $ 
% $\exists h: \s{v} \rightarrow \s{v}'$, $\forall \s{n} \text{ with } (\sum_{i=1}^m \s{n}_i = k, \forall i \ \s{n}_i \geq 1) $, 
% $\bm{mwl}_k^{(t)}(G, f(\s{v}, \s{n}) ) = \bm{mwl}_k^{(t)}(G', f(h(\s{v}), \s{n}) )$ .
% \end{lemma}

\begin{lemma}\label{lm:3}
$\forall t$, $\bm{mwl}_k^{(t)}(G, \ms{v}) = \bm{mwl}_k^{(t)}(G', \ms{v}')$ $ \Longleftrightarrow $ 
% $\exists h:\ms{v}\rightarrow \ms{v}' $,
$\forall h \in F^{(t)}(G,G',\ms{v}, \ms{v}') $,
$\forall \s{n} \text{ with } (\sum_{i=1}^m \s{n}_i = k, \forall i \ \s{n}_i \geq 1) $,
$\bm{mwl}_k^{(t)}(G, f(s(\ms{v}), \s{n}) ) = \bm{mwl}_k^{(t)}(G', f(h(s(\ms{v})), \s{n}) )$.
\end{lemma}
% \lz{This relies on the correctness of Lemma 1.}

When $t=0$, by the definition of $\bm{swl}_k^{(0)}$, the statement is true. Now hypothesize that the statement is true for $\leq t$ case. 
For $t+1$ case, the left side implies that existing a $\ms{v}$ with $s(\ms{v}) =\s{v}$ and a $\ms{v}'$ with $s(\ms{v}') =\s{v}'$, such that $\bm{mwl}_k^{(t+1)}(G, \ms{v}) = \bm{mwl}_k^{(t+1)}(G', \ms{v}')$. And for a mapping $h\in F^{(t+1)}(G,G', \ms{v}, \ms{v}')$, we have $\forall \s{n} \text{ with } (\sum_{i=1}^m \s{n}_i = k, \forall i \ \s{n}_i \geq 1) $,
$\bm{mwl}_k^{(t+1)}(G, f(s(\ms{v}), \s{n}) ) = \bm{mwl}_k^{(t+1)}(G', f(h(s(\ms{v})), \s{n}) )$. 

For a specific $\s{n}$, define $\ms{u} = f(s(\ms{v}), \s{n}) $ and $\ms{u}' = f(h(s(\ms{v})), \s{n}) = h(\ms{u}) $.
By Eq.\eqref{eq:kmwl}:\\
 (1) $\bm{mwl}_k^{(t)}(G, \ms{u} ) = \bm{mwl}_k^{(t)}(G', \ms{u}' )$; \\
 (2) $\forall  f \in {F^{(t+1)}(G,G',\ms{u}, \ms{u}') }$, 
 $\forall y \in \ms{u}$, $\mms{\bm{mwl}_k^{(t)}(G, \ms{u}[x/\text{idx}_{\ms{v}}(y)]) \big| x \in V(G)} = \mms{\bm{mwl}_k^{(t)}(G', \ms{u}'[x/\text{idx}_{\ms{v}'}(f(y))]) \big| x \in V(G')} $. 

As $h \in  {F^{(t+1)}(G,G',\ms{v}, \ms{v}') }= {F^{(t+1)}(G,G',\ms{u}, \ms{u}') } $, we can change (2) by choosing $f$ as $h$. Hence we update it as:\\
(2) $\forall y \in \ms{u}$, $\mms{\bm{mwl}_k^{(t)}(G, \ms{u}[x/\text{idx}_{\ms{v}}(y)]) \big| x \in V(G)} = \mms{\bm{mwl}_k^{(t)}(G', h(\ms{u})[x/\text{idx}_{\ms{v}'}(h(y))]) \big| x \in V(G')} $ \\
And (2)  can be split into two parts:\\
(2) $\forall y \in \ms{u}$, $\mms{\bm{mwl}_k^{(t)}(G, \ms{u}[x/\text{idx}_{\ms{v}}(y)]) \big| x \in s(\ms{u})} = \mms{\bm{mwl}_k^{(t)}(G', h(\ms{u})[x/\text{idx}_{\ms{v}'}(h(y))]) \big| x \in s(h(\ms{u}))} $  and\\ $\mms{\bm{mwl}_k^{(t)}(G, \ms{u}[x/\text{idx}_{\ms{v}}(y)]) \big| x \in V(G)\setminus s(\ms{u})} = \mms{\bm{mwl}_k^{(t)}(G', h(\ms{u})[x/\text{idx}_{\ms{v}'}(h(y))]) \big| x \in V(G') \setminus s(h(\ms{u}))} $ \\

Combining the Lemma.\ref{lm:3} and induction hypothesis, also knowing that $s(\ms{u}) = \s{v}$ and $s(\ms{u}') = \s{v}' $, we get:\\
(a) $\bm{swl}_k^{(t)}(G, \s{v} ) = \bm{swl}_k^{(t)}(G', \s{v}' )$; \\
(b) $\forall y \in \s{v}$,  $\mms{\bm{swl}_k^{(t)}(G, \s{v}\setminus y ) \big| x \in \s{v} } = \mms{\bm{swl}_k^{(t)}(G',  \s{v}' \setminus h(y) ) \big| x \in  \s{v}'} $ , this is derived with picking $\s{n}$ with $\s{n}[y]=1$, such that $\ms{u}$ only has 1 $y$ and $\ms{u}'$ only has 1 $h(y)$.  \\
(c) $\forall y \in \s{v}$,  $\mms{\bm{swl}_k^{(t)}(G, \s{v}\cup \{x\} ) \big| x \in V(G)\setminus \s{v} } = \mms{\bm{swl}_k^{(t)}(G',  \s{v}' \cup \{x\} ) \big| x \in  V(G')\setminus \s{v}'} $, this is derived with picking $\s{n}$ with $\s{n}[y]>1$.\\
(d) $\forall y \in \s{v}$,  $\mms{\bm{swl}_k^{(t)}(G, \s{v}\setminus y \cup \{x\} ) \big|  x \in V(G)\setminus \s{v}  } = \mms{\bm{swl}_k^{(t)}(G',  \s{v}' \setminus h(y)  \cup \{x\} ) \big|x \in  V(G')\setminus \s{v}'} $ , this is derived with picking $\s{n}$ with $\s{n}[y]=1$, such that $\ms{u}$ only has 1 $y$ and $\ms{u}'$ only has 1 $h(y)$.  \\

Now combining (a) (b) (c) (d) and using Eq.\eqref{eq:kswl}, we can get that $\bm{swl}_k^{(t+1)}(G, \s{v}) = \bm{swl}_k^{(t+1)}(G', \s{v}')$, and we proved the statement.

% For $t+1$ case, left side implies that $\exists h: \s{v} \rightarrow \s{v}'$, $\forall \s{n}, \bm{mwl}_k^{(t+1)}(G, f(\s{v}, \s{n}) ) = \bm{mwl}_k^{(t+1)}(G', f(h(\s{v}), \s{n}) )$. Then we can derive: (1) $\forall \s{n}$, $\bm{mwl}_k^{(t)}(G, f(\s{v}, \s{n}) ) = \bm{mwl}_k^{(t)}(G', f(h(\s{v}), \s{n}) )$; 
% (2) $\forall \s{n}$, $\forall i\in [m]$, $\exists p$, $\forall x \in V(G)$, $\bm{mwl}_k^{(t)}(G, \{x\} \cup f(\s{v}, \s{n}-\bm{e}_{i} ) ) = \bm{mwl}_k^{(t)}(G', \{p(x)\} \cup f(h(\s{v}), \s{n} - \bm{e}_{q(i)} )  )$

% \lz{Some hardness of dealing with ordering matching left}

% % $\mms{\bm{mwl}_k^{(t)}(G, f(\s{v}, \s{n}) ) \ \big|\  \sum_{i=1}^m \s{n}_i = k, \forall i \ \s{n}_i \geq 1}$

\end{proof}

\subsubsection{Proof of Theorem 5}

\begin{customthm}{5}
Let $k\ge 1$, then $\forall t\in \mathbb{N}$ and all graphs $G, G'$:

\vspace{-0.05in}
\cbit
\item[(1)] when $1\leq c_1 <c_2 \leq k$,
if $G,G'$ cannot be distinguished by  {\sc $(k,c_2)(\leq)$-SetWL}, they cannot be distinguished by {\sc $(k,c_1)(\leq)$-SetWL} 
\item[(2)] when $k_1 < k_2$, $\forall c\leq k_1$, if $G,G'$ cannot be distinguished by  {\sc $(k_2,c)(\leq)$-SetWL}, they cannot be distinguished by {\sc $(k_1,c)(\leq)$-SetWL}
\ceit
\end{customthm}

\begin{proof}
We will prove (2) first, and the proof for (1) follows the same argument. To help understand, we present the formulation for \kcswl first.

\begin{align}
    &\bm{swl}_{k,c}^{(t+\frac{1}{2})}(G, \s{v}) = \mms{\bm{swl}_{k,c}^{(t)}(G, \s{u}) \ |\ \s{u} \in \mathcal{N}^G_{k,c,\text{right}}(\s{v})} \\
    &\bm{swl}_{k,c}^{(t+1)}(G, \s{v}) = \bigg( 
    \bm{swl}_{k,c}^{(t)}(G, \s{v}), 
    \bm{swl}_{k,c}^{(t+\frac{1}{2})}(G, \s{v}), 
    \mms{\bm{swl}_{k,c}^{(t)}(G, \s{u}) \ |\ \s{u} \in \mathcal{N}^G_{k,c,\text{left}}(\s{v})},\\
    &\mms{\bm{swl}_{k,c}^{(t+\frac{1}{2})}(G, \s{u}) \ |\ \s{u} \in \mathcal{N}^G_{k,c,\text{left}}(\s{v}) } 
    \bigg)   
% \vspace{-0.3in}
\end{align}

\begin{lemma}\label{lm:k1k2}
For $k_1 < k_2$, for any $t$, for any $\s{v}$ and $\s{v}'$ with $\leq k_1$ nodes inside, $\bm{swl}_{k_2,c}^{(t)}(G, \s{v}) = \bm{swl}_{k_2,c}^{(t)}(G', \s{v}')$ $\Longrightarrow$ $\bm{swl}_{k_1,c}^{(t)}(G, \s{v}) = \bm{swl}_{k_1,c}^{(t)}(G', \s{v}')$.
\end{lemma}
\textit{Proof of Lemma.\ref{lm:k1k2}:}
\begin{proof}
We prove it by induction. As the color initialization stage of   {\sc $(k_1,c)(\leq)$-SetWL} and  {\sc $(k_2,c)(\leq)$-SetWL} are the same, when $t=0$ the statement of the Lemma is correct. 
Assume it holds correct for $\leq t$. When $t+1$,  $\bm{swl}_{k_2,c}^{(t+1)}(G, \s{v}) = \bm{swl}_{k_2,c}^{(t+1)}(G', \s{v}')$ implies:

(1) $\bm{swl}_{k_2,c}^{(t)}(G, \s{v})$ = $\bm{swl}_{k_2,c}^{(t)}(G', \s{v}')$\\
(2) $\bm{swl}_{k_2,c}^{(t+1/2)}(G, \s{v})$ = $\bm{swl}_{k_2,c}^{(t+1/2)}(G', \s{v}')$\\
(3) $\mms{\bm{swl}_{k_2,c}^{(t)}(G, \s{u}) \ |\ \s{u} \in \mathcal{N}^G_{k_2,c,\text{left}}(\s{v})}$
= $\mms{\bm{swl}_{k_2,c}^{(t)}(G', \s{u}') \ |\ \s{u}' \in \mathcal{N}^{G'}_{k_2,c,\text{left}}(\s{v}')}$ \\
(4) $\mms{\bm{swl}_{k_2,c}^{(t+1/2)}(G, \s{u}) \ |\ \s{u} \in \mathcal{N}^G_{k_2,c,\text{left}}(\s{v})}$
= $\mms{\bm{swl}_{k_2,c}^{(t+1/2)}(G', \s{u}') \ |\ \s{u}' \in \mathcal{N}^{G'}_{k_2,c,\text{left}}(\s{v}')}$

(1) + induction hypothesis $\Longrightarrow$ (a)  $\bm{swl}_{k_1,c}^{(t)}(G, \s{v})$ = $\bm{swl}_{k_1,c}^{(t)}(G', \s{v}')$ \\
(2) has two situations: $|\s{v}| = |\s{v}'| < k_1$ and  $|\s{v}| = |\s{v}'| = k_1$. For the first situation,  $\mathcal{N}^G_{k_2,c,\text{right}}(\s{v}) = \mathcal{N}^G_{k_1,c,\text{right}}(\s{v})$ and $\mathcal{N}^{G'}_{k_2,c,\text{right}}(\s{v}') = \mathcal{N}^{G'}_{k_1,c,\text{right}}(\s{v}') $, $\exists $ a bijective mapping $b$ between $\mathcal{N}^{G'}_{k_2,c,\text{right}}(\s{v}')$ and $\mathcal{N}^{G}_{k_2,c,\text{right}}(\s{v})$ such that $\forall \s{u}\in \mathcal{N}^{G}_{k_2,c,\text{right}}(\s{v}) $, $\bm{swl}_{k_2,c}^{(t)}(G, \s{u})$ = $\bm{swl}_{k_2,c}^{(t)}(G', b(\s{u}))$ and by induction $\bm{swl}_{k_1,c}^{(t)}(G, \s{u})$ = $\bm{swl}_{k_1,c}^{(t)}(G', b(\s{u}))$, hence $\bm{swl}_{k_1,c}^{(t+1/2)}(G, \s{v})$ = $\bm{swl}_{k_1,c}^{(t+1/2)}(G', \s{v}')$. For the second situation, $\mathcal{N}^{G'}_{k_2,c,\text{right}}(\s{v}') $ elements while  $\mathcal{N}^{G'}_{k_1,c,\text{right}}(\s{v}') = \emptyset $, then clearly (b) $\bm{swl}_{k_1,c}^{(t+1/2)}(G, \s{v})$ = $\bm{swl}_{k_1,c}^{(t+1/2)}(G', \s{v}')$ (both are empty multisets). 
\\
(3) follows the same argument step in (2)'s first situation, with induction hypothesis we can get (c) $\mms{\bm{swl}_{k_1,c}^{(t)}(G, \s{u}) \ |\ \s{u} \in \mathcal{N}^G_{k_1,c,\text{left}}(\s{v})}$
= $\mms{\bm{swl}_{k_1,c}^{(t)}(G', \s{u}') \ |\ \s{u}' \in \mathcal{N}^{G'}_{k_1,c,\text{left}}(\s{v}')}$ \\
(4) implies that $\exists $ a mapping $b$ between $\mathcal{N}^{G'}_{k_2,c,\text{left}}(\s{v}')$ and $\mathcal{N}^{G}_{k_2,c,\text{left}}(\s{v})$ such that for any $\s{u}\in \mathcal{N}^{G'}_{k_2,c,\text{left}}(\s{v}') $, $\bm{swl}_{k_2,c}^{(t+1/2)}(G, \s{u})$ = $\bm{swl}_{k_2,c}^{(t+1/2)}(G', b(\s{u}'))$. Using the same argument in (2) and induction hypothesis, this implies that for any $\s{u}\in \mathcal{N}^{G'}_{k_1,c,\text{left}}(\s{v}') $, $\bm{swl}_{k_1,c}^{(t+1/2)}(G, \s{u})$ = $\bm{swl}_{k_1,c}^{(t+1/2)}(G', b(\s{u}'))$. Hence we get (d) $\mms{\bm{swl}_{k_1,c}^{(t+1/2)}(G, \s{u}) \ |\ \s{u} \in \mathcal{N}^G_{k_1,c,\text{left}}(\s{v})}$
= $\mms{\bm{swl}_{k_1,c}^{(t+1/2)}(G', \s{u}') \ |\ \s{u}' \in \mathcal{N}^{G'}_{k_1,c,\text{left}}(\s{v}')}$.

Combining (a) (b) (c) (d) we get $\bm{swl}_{k_1,c}^{(t+1)}(G, \s{v}) = \bm{swl}_{k_1,c}^{(t+1)}(G', \s{v}')$.
\end{proof}

With Lemma.\ref{lm:k1k2} we are ready to prove (2) in Theorem 3. When two graphs $G$ and $G'$ cannot be distinguished by  {\sc $(k_2,c)(\leq)$-SetWL}, we have $\mms{ \bm{swl}_{k_2,c}^{(t)}(G, \s{v}) \ |\  \s{v} \in  V(S_{k_2,c\text{-swl}}(G) ) } $ = $\mms{ \bm{swl}_{k_2,c}^{(t)}(G', \s{v}') \ |\  \s{v}' \in  V(S_{k_2,c\text{-swl}} (G')) } $. When $\s{v}$ and $\s{u}$ have different $k$ (number of nodes) and $c$ (number of components of its induced subgraph), their color cannot be the same (as $t=0$ already be different). Hence 
$\mms{ \bm{swl}_{k_2,c}^{(t)}(G, \s{v}) \ |\  \s{v} \in  V(S_{k_2,c\text{-swl}}(G) ) } $ = $\mms{ \bm{swl}_{k_2,c}^{(t)}(G', \s{v}') \ |\  \s{v}' \in  V(S_{k_2,c\text{-swl}} (G')) } $ is equivalent to $\forall k\leq k_2, cc \leq c$, $\mms{ \bm{swl}_{k_2,c}^{(t)}(G, \s{v}) \ |\  \s{v} \in  V_{k,cc}(S_{k_2,c\text{-swl}}(G) ) } $ = $\mms{ \bm{swl}_{k_2,c}^{(t)}(G', \s{v}') \ |\  \s{v}' \in  V_{k,cc}(S_{k_2,c\text{-swl}} (G')) } $. With Lemma.\ref{lm:k1k2} we can get that $\forall k\leq k_1, cc \leq c$, $\mms{ \bm{swl}_{k_1,c}^{(t)}(G, \s{v}) \ |\  \s{v} \in  V_{k,cc}(S_{k_1,c\text{-swl}}(G) ) } $ = $\mms{ \bm{swl}_{k_1,c}^{(t)}(G', \s{v}') \ |\  \s{v}' \in  V_{k,cc}(S_{k_1,c\text{-swl}} (G')) } $. Hence  two graphs $G$ and $G'$ cannot be distinguished by  {\sc $(k_1,c)(\leq)$-SetWL}.

% \lz{Mark here, further give the proof of (1), should be easy, just need time to write. Follows the same argument.}

\end{proof}
\subsubsection{Proof of Theorem 6}
\begin{customthm}{6}
When ($i$) BaseGNN can distinguish any non-isomorhpic graphs with at most $k$ nodes, ($ii$) all MLPs have sufficient depth and width, and ($iii$) POOL is an injective function, then for any $t\in \mathbb{N}$, $t$-layer \method is as expressive as \kcswl at the $t$-th iteration. 
\end{customthm}

\begin{proof}
For any $G, \s{v}$, let $\bm{swl}_{k,c}^{(t)}(G, \s{v})$ denotes the color of $\s{v}$ at $t$-iteration \kcswl and $h_{k,c}^{(t)}(G, \s{v})$ denotes the embedding of $\s{v}$ at $t$-th layer \method. We prove the above theorem by showing that 
$\bm{swl}_{k,c}^{(t)}(G, \s{v})$ = $\bm{swl}_{k,c}^{(t)}(G', \s{v}’)$ $\Longleftrightarrow$   $h_{k,c}^{(t)}(G, \s{v})$ = $h_{k,c}^{(t)}(G', \s{v}’)$. We remove the subscript $k,c$ when possible without introducing confusion.  

For easier reference, recall the updating formulation for $t$-iteration \kcswl is 
\begin{align}
\bm{swl}^{(t+\frac{1}{2})}(G, \s{v})=& \mms{\bm{swl}_{k,c}^{(t)}(G, \s{u}) \ |\ \s{u} \in \mathcal{N}^G_{\text{right}}(\s{v})}\\
    \bm{swl}^{(t+1)}(G, \s{v}) = & \bigg( 
    \bm{swl}^{(t)}(G, \s{v}), 
    \bm{swl}^{(t+\frac{1}{2})}(G, \s{v}), \\
    &\mms{\bm{swl}^{(t)}(G, \s{u}) \ |\ \s{u} \in \mathcal{N}^G_{\text{left}}(\s{v})},
    \mms{\bm{swl}^{(t+\frac{1}{2})}(G, \s{u}) \ |\ \s{u} \in \mathcal{N}^G_{\text{left}}(\s{v}) } 
    \bigg)   \label{eq:kcswl-A}
% \vspace{-0.3in}
\end{align}

And the updating formulation for \method is 
\begin{align}
    h^{(t+\frac{1}{2})}(\s{v}) =& \sum_{\s{u}\in \mathcal{N}^G_{\text{right}}(\s{v})} \text{MLP}^{(t+\frac{1}{2})}(h^{(t)}(\s{u})) \label{eq:SetGNN1-A}\\
    h^{(t+1)}(\s{v}) = &  \text{MLP}^{(t)}\Big(h^{(t)}(\s{v}),h^{(t+\frac{1}{2})}(\s{v}), \\
    & \sum_{\s{u}\in \mathcal{N}^G_{\text{left}}(\s{v})} \text{MLP}^{(t)}_A(h^{(t)}(\s{u})),
    \sum_{\s{u}\in \mathcal{N}^G_{\text{left}}(\s{v})} \text{MLP}^{(t)}_B(h^{(t+\frac{1}{2})}(\s{u}))
    \Big)\label{eq:SetGNN2-A}
\end{align}

At $t=0$, given powerful enough BaseGNN with condition (i) in the theorem,  $h^{(0)}(G, \s{v})$ = $h^{(0)}(G', \s{v}’)$ $\Longleftrightarrow$ $G[\s{v}]$ and $G'[\s{v}']$ are isomorphic $\Longleftrightarrow$ $\bm{swl}^{(0)}(G, \s{v})$ = $\bm{swl}^{(0)}(G', \s{v}’)$.

Now assume for $\leq t$ iterations the claim $\bm{swl}^{(t)}(G, \s{v})$ = $\bm{swl}^{(t)}(G', \s{v}’)$ $\Longleftrightarrow$   $h^{(t)}(G, \s{v})$ = $h^{(t)}(G', \s{v}’)$ holds (for any $\s{v}$ and $\s{v}'$). We prove it holds for $t+1$ iteration. We first prove the forward direction. $\bm{swl}^{(t+1)}(G, \s{v})$ = $\bm{swl}^{(t+1)}(G', \s{v}’)$ imples that 

(1) $\bm{swl}^{(t)}(G, \s{v})$ = $\bm{swl}^{(t)}(G', \s{v}')$\\
(2) $\bm{swl}^{(t+1/2)}(G, \s{v})$ = $\bm{swl}^{(t+1/2)}(G', \s{v}')$\\
(3) $\mms{\bm{swl}^{(t)}(G, \s{u}) \ |\ \s{u} \in \mathcal{N}^G_{\text{left}}(\s{v})}$
= $\mms{\bm{swl}^{(t)}(G', \s{u}') \ |\ \s{u}' \in \mathcal{N}^{G'}_{\text{left}}(\s{v}')}$ \\
(4) $\mms{\bm{swl}^{(t+1/2)}(G, \s{u}) \ |\ \s{u} \in \mathcal{N}^G_{\text{left}}(\s{v})}$
= $\mms{\bm{swl}^{(t+1/2)}(G', \s{u}') \ |\ \s{u}' \in \mathcal{N}^{G'}_{\text{left}}(\s{v}')}$

(1) and (3) can be directly transformed to relationship between $h$ using the inductive hypothesis. Formally, we have \\
(a) $h^{(t)}(G, \s{v})$ = $h^{(t)}(G', \s{v}')$  and \\
(c) $\mms{h^{(t)}(G, \s{u}) \ |\ \s{u} \in \mathcal{N}^G_{\text{left}}(\s{v})}$
= $\mms{h^{(t)}(G', \s{u}') \ |\ \s{u}' \in \mathcal{N}^{G'}_{\text{left}}(\s{v}')}$ \\

(2) and (4) need one additional process. Notice that (2) is equivalent to $\mms{\bm{swl}^{(t)}(G, \s{u}) \ |\ \s{u} \in \mathcal{N}^G_{\text{right}}(\s{v})} = \mms{\bm{swl}^{(t)}(G', \s{u}') \ |\ \s{u} \in \mathcal{N}^{G'}_{\text{right}}(\s{v}')}$ and by inductive hypothesis we know $\mms{h^{(t)}(G, \s{u}) \ |\ \s{u} \in \mathcal{N}^G_{\text{right}}(\s{v})} = \mms{h^{(t)}(G', \s{u}') \ |\ \s{u} \in \mathcal{N}^{G'}_{\text{right}}(\s{v}')}$. As the formulation Eq.\eqref{eq:SetGNN1-A} applies a MLP with summation, which is  permutation invariant to ordering, to the multiset $\mms{h^{(t)}(G, \s{u}) \ |\ \s{u} \in \mathcal{N}^G_{\text{right}}(\s{v})}$, the same multiset leads to the same output. Hence we know \\
(b) $h^{(t+1/2)}(G, \s{v})$ = $h^{(t+1/2)}(G', \s{v}')$. 

For (4) we know that there is a bijective mapping $g$ between $\mathcal{N}^G_{\text{left}}(\s{v})$ and $\mathcal{N}^G_{\text{left}}(\s{v}')$ such that $\bm{swl}^{(t+1/2)}(G, \s{u}) = \bm{swl}^{(t+1/2)}(G', g(\s{u}))$. Then using the same argument as (2) =>(b) we can get $h^{(t+1/2)}(G, \s{u}) = h^{(t+1/2)}(G', g(\s{u}))$ for any $\s{u} \in \mathcal{N}^G_{\text{left}}(\s{v})$, which is equivalent to \\
(d) $\mms{h^{(t+1/2)}(G, \s{u}) \ |\ \s{u} \in \mathcal{N}^G_{\text{left}}(\s{v})}$
= $\mms{h^{(t+1/2)}(G', \s{u}') \ |\ \s{u}' \in \mathcal{N}^{G'}_{\text{left}}(\s{v}')}$.

Combining (a)(b)(c)(d) and using the permutation invariant property of MLP with summation, we can derive that $h^{(t+1)}(G, \s{v})$ = $h^{(t+1)}(G', \s{v}’)$. 

Now for the backward direction. We first characterize the property of MLP with summation from DeepSet \cite{zaheer2017deep} and GIN \cite{Xu:2019ty}'s Lemma 5.

\begin{lemma}[Lemma 5 of \cite{Xu:2019ty}]
Assume $\mathcal{X}$ is countable. There exists a function $f:\mathcal{X} \rightarrow \R^n$ so that $h(X) = \sum_{x\in X} f(x) $ is unique for each multiset $X\in \mathcal{X}$ of bounded size.  
\end{lemma}

Using this Lemma, we know that given enough depth and width of a MLP, there exist a MLP that $\sum_{x\in X} \text{MLP}(x) = \sum_{y\in Y} \text{MLP}(y) \Longleftrightarrow X = Y$ or in other words two multisets $X$ and $Y$ are equivalent. Now by $h^{(t+1)}(G, \s{v})$ = $h^{(t+1)}(G', \s{v}’)$ and using the Eq.\eqref{eq:SetGNN2-A}, we know there exists MLPs inside Eq.\eqref{eq:SetGNN1-A} and Eq.\eqref{eq:SetGNN2-A}, such that

(1) $h^{(t)}(G, \s{v})$ = $h^{(t)}(G', \s{v}')$\\
(2) $h^{(t+1/2)}(G, \s{v})$ = $h^{(t+1/2)}(G', \s{v}')$\\
(3) $\mms{h^{(t)}(G, \s{u}) \ |\ \s{u} \in \mathcal{N}^G_{\text{left}}(\s{v})}$
= $\mms{h^{(t)}(G', \s{u}') \ |\ \s{u}' \in \mathcal{N}^{G'}_{\text{left}}(\s{v}')}$ \\
(4) $\mms{h^{(t+1/2)}(G, \s{u}) \ |\ \s{u} \in \mathcal{N}^G_{\text{left}}(\s{v})}$
= $\mms{h^{(t+1/2)}(G', \s{u}') \ |\ \s{u}' \in \mathcal{N}^{G'}_{\text{left}}(\s{v}')}$ 

Where (3) and (4) are derived using the provided lemma. Hence following the same argument as the forward process, we can prove that $\bm{swl}^{(t+1)}(G, \s{v})$ = $\bm{swl}^{(t+1)}(G', \s{v}’)$. 

Combining the forward and backward direction, we have $\bm{swl}^{(t+1)}(G, \s{v})$ = $\bm{swl}^{(t+1)}(G', \s{v}’)$ $\Longleftrightarrow$   $h^{(t+1)}(G, \s{v})$ = $h^{(t+1)}(G', \s{v}’)$. Hence by induction we proved for any step $t$ and any $\s{v}, \s{v}'$, the above statement is true. This shows that \method and \kcswl have same expressivity.

\end{proof}

\subsubsection{Proof of Theorem 7}
\begin{customthm}{7}
For any $t\in\mathbb{N}$, the $t$-layer \methods is more expressive than the $t$-layer \method. As $\lim_{t\to\infty}$, \method is as expressive as \methods.
\end{customthm}

\begin{proof}
We first prove that $t$-layer \methods is more expressive than $t$-layer \method, by showing that if a $m$-set $\s{u}$ has the same representation to another $m$-set $\s{v}$ in $t$-layer \methods, then they also have the same representation in $t$-layer \method. 
Let $h$ be the representation inside \methods, and $g$ be the representation inside \method.

To simplify the proof, we first simplify the formulations of \methods Eq.\ref{eq:bmp1} and Eq.\ref{eq:bmp2} by removing unnecessary super and under script of MLP without introducing ambiguity. We then add another superscript to embeddings to indicate which step inside the one-layer bidirectional propagation. Notice that for one-layer bidirectional propagation, there are $2k-2$ intermediate sequential steps. The proof assumes that all MLPs are injective.
We rewrite Eq.\ref{eq:bmp1} and Eq.\ref{eq:bmp2} as follows

\begin{footnotesize}
\begin{align}
    &\forall m\text{-set } \s{v},    h^{(t+\frac{1}{2})}(\s{v}) := h^{(t+\frac{1}{2}, k-m)}(\s{v}) =  \text{MLP}\Big(
    h^{(t)}(\s{v}),
    \sum_{\s{w}\in \mathcal{N}^G_{\text{right}}(\s{v})} \text{MLP}(h^{{(t+\frac{1}{2}, k-m-1)}}(\s{w})) 
    \Big) \label{eq:seq-1} \\ 
    &\forall m\text{-set } \s{v},   
      h^{(t+1)}(\s{v}) :=h^{(t+1, m-1)}(\s{v}) =  \text{MLP}\Big(
    h^{(t+\frac{1}{2}, k-m)}(\s{v}),
    \sum_{\s{w}\in \mathcal{N}^G_{\text{left}}(\s{v})} \text{MLP}(h^{ {(t+1, m-2)}}(\s{w})) \label{eq:seq-2}
    \Big) 
\end{align}
\end{footnotesize}
Where we have boundary case with $h^{{(t+\frac{1}{2}, 0)}}(\s{u}) = h^{{(t)}}(\s{u})$ and $h^{{(t+1, 0)}}(\s{u}) = h^{{(t+\frac{1}{2})}}(\s{u})$. $  h^{(t+\frac{1}{2})}(\s{v}):= h^{(t+\frac{1}{2}, k-m)}(\s{v})$ represents that the representation is calculated at $k-m$ step for $t+\frac{1}{2}$ layer. 

We prove the theorem by induction on $t$. Specifically, we want to prove that for any $t$, $\s{v}$, $\s{u}$, $h^{(t)}(\s{u}) = h^{(t)}(\s{v}) \Longrightarrow g^{(t)}(\s{u}) = g^{(t)}(\s{v}) $ and  $h^{(t-\frac{1}{2})}(\s{u}) = h^{(t-\frac{1}{2})}(\s{v}) \Longrightarrow g^{(t-\frac{1}{2})}(\s{u}) = g^{(t-\frac{1}{2})}(\s{v}) $. The base case is easy to verify as the definition of initialization step is the same.

Now assume that for $t \le l$ we have for any $t$, $\s{v}$, $\s{u}$, $h^{(t)}(\s{u}) = h^{(t)}(\s{v}) \Longrightarrow g^{(t)}(\s{u}) = g^{(t)}(\s{v}) $ and  $h^{(t-\frac{1}{2})}(\s{u}) = h^{(t-\frac{1}{2})}(\s{v}) \Longrightarrow g^{(t-\frac{1}{2})}(\s{u}) = g^{(t-\frac{1}{2})}(\s{v}) $. We first prove that for $t=l+1$,  $h^{(l+\frac{1}{2})}(\s{u}) = h^{(l+\frac{1}{2})}(\s{v}) \Longrightarrow g^{(l+\frac{1}{2})}(\s{u}) = g^{(l+\frac{1}{2})}(\s{v}) $. 

First, Eq.\ref{eq:seq-1} can be rewrite as 
\begin{align}
   & h^{(t+\frac{1}{2})}(\s{v}) : h^{(t+\frac{1}{2}, k-m)}(\s{v})  =  \text{MLP}\Big(
    h^{(t)}(\s{v}),
    \sum_{\s{w}\in \mathcal{N}^G_{\text{right}}(\s{v})} \text{MLP}(h^{{(t+\frac{1}{2}, k-m-1)}}(\s{w}))
     \Big)\\
    & =  \text{MLP}\Big(
        h^{(t)}(\s{v}),
        \sum_{\s{w}\in \mathcal{N}^G_{\text{right}}(\s{v})}
        \text{MLP}( \text{MLP}(h^{(t)}(\s{w})),  
         \sum_{\s{p}\in \mathcal{N}^G_{\text{right}}(\s{w})}
        \text{MLP}(h^{{(t+\frac{1}{2}, k-m-2)}}(\s{p}))
        )
    \Big)
\end{align}
Then $ h^{(l+\frac{1}{2})}(\s{v}) =  h^{(l+\frac{1}{2})}(\s{u}) \Longrightarrow \sum_{\s{w}\in \mathcal{N}^G_{\text{right}}(\s{v})} \text{MLP}(h^{(l)}(\s{w})) =\sum_{\s{w}\in \mathcal{N}^G_{\text{right}}(\s{u})} \text{MLP}(h^{(l)}(\s{w}))$. Hence we can find a bijective mapping $f$ between $\mathcal{N}^G_{\text{right}}(\s{v})$ and $\mathcal{N}^G_{\text{right}}(\s{u})$ such that $\text{MLP}(h^{(l)}(\s{w})) = \text{MLP}(h^{(l)}( f(\s{w}) ))$. Then by inductive hyphothesis, $\text{MLP}(g^{(l)}(\s{w})) = \text{MLP}(g^{(l)}( f(\s{w}) ))$ for all $\s{w} \in \mathcal{N}^G_{\text{right}}(\s{v})$. This implies that $\sum_{\s{w}\in \mathcal{N}^G_{\text{right}}(\s{v})} \text{MLP}(g^{(l)}(\s{w})) =\sum_{\s{w}\in \mathcal{N}^G_{\text{right}}(\s{u})} \text{MLP}(g^{(l)}(\s{w}))$ or equivalently $g^{(l+\frac{1}{2})}(\s{v}) =  g^{(l+\frac{1}{2})}(\s{u})$. 

Now we can assume that for $t \le m$ we have for any $t$, $\s{v}$, $\s{u}$, $h^{(t)}(\s{u}) = h^{(t)}(\s{v}) \Longrightarrow g^{(t)}(\s{u}) = g^{(t)}(\s{v}) $ and for $t \le l+1$  $h^{(t-\frac{1}{2})}(\s{u}) = h^{(t-\frac{1}{2})}(\s{v}) \Longrightarrow g^{(t-\frac{1}{2})}(\s{u}) = g^{(t-\frac{1}{2})}(\s{v}) $. We prove that for $t=l+1$,  $h^{(l+1)}(\s{u}) = h^{(l+1)}(\s{v}) \Longrightarrow g^{(l+1)}(\s{u}) = g^{(l+1)}(\s{v}) $. 

We first rewrite Eq.\ref{eq:seq-2} as 
\begin{align}
    &h^{(t+1)}(\s{v}) :=h^{(t+1, m-1)}(\s{v}) =  \text{MLP}\Big(
        h^{(t+\frac{1}{2}, k-m)}(\s{v}),
        \sum_{\s{w}\in \mathcal{N}^G_{\text{left}}(\s{v})} \text{MLP}(h^{ {(t+1, m-2)}}(\s{w}))
    \Big)\\
    &=\text{MLP}\Bigg(
        h^{(t+\frac{1}{2}, k-m)}(\s{v}),
        \sum_{\s{w}\in \mathcal{N}^G_{\text{left}}(\s{v})} \text{MLP}  
        \Big( 
        \text{MLP}\big(h^{(t+\frac{1}{2}, k-m+1)}(\s{w}),  
        \sum_{\s{p}\in \mathcal{N}^G_{\text{left}}(\s{w})} \text{MLP}(h^{ {(t+1, m-3)}}(\s{p}))
        \big)
        \Big)
    \Bigg)
\end{align}
Then  $h^{(l+1)}(\s{u}) = h^{(l+1)}(\s{v})$ implies 

1) $h^{(t+\frac{1}{2}, k-m)}(\s{v}) = h^{(t+\frac{1}{2}, k-m)}(\s{u})$ or equivalently
$h^{(t+\frac{1}{2})}(\s{v}) = h^{(t+\frac{1}{2})}(\s{u})$ \\
2) $\mms{h^{(t+\frac{1}{2})}(\s{w}) \ | \ \s{w} \in \mathcal{N}^G_{\text{left}}(\s{v})}  $ = $\mms{h^{(t+\frac{1}{2})}(\s{w}) \ | \ \s{w} \in \mathcal{N}^G_{\text{left}}(\s{u})}  $

Also by Eq.\ref{eq:seq-1} we know that $h^{(t+\frac{1}{2})}(\s{v}) = h^{(t+\frac{1}{2})}(\s{u})$ implies  

3) $h^{(t)}(\s{v}) = h^{(t)}(\s{u})$. 

Combining with 2) and 3) we know

4) $\mms{h^{(t)}(\s{w}) \ | \ \s{w} \in \mathcal{N}^G_{\text{left}}(\s{v})}  $ = $\mms{h^{(t)}(\s{w}) \ | \ \s{w} \in \mathcal{N}^G_{\text{left}}(\s{u})}$

Now using our inductive hypothesis, we can transform 1) 2) 3) 4) by replace $h$ with $g$. Then based on the equation in \ref{eq:SetGNN2}, we know $g^{(l+1)}(\s{u}) = g^{(l+1)}(\s{v}) $. 

Combining the two proved inductive hypothesis and applying them alternately, we know that for any $t$, $\s{v}$, $\s{u}$, $h^{(t)}(\s{u}) = h^{(t)}(\s{v}) \Longrightarrow g^{(t)}(\s{u}) = g^{(t)}(\s{v}) $ and  $h^{(t-\frac{1}{2})}(\s{u}) = h^{(t-\frac{1}{2})}(\s{v}) \Longrightarrow g^{(t-\frac{1}{2})}(\s{u}) = g^{(t-\frac{1}{2})}(\s{v}) $. Hence $t$-layer \methods is more expressive than $t$-layer \method.

Now let's considering $t$ to infinity. Then all representations will become stable with $h^{(t)}(\s{v}) = h^{(t)}(\s{u})  \Longleftrightarrow h^{(t+1)}(\s{v}) = h^{(t+1)}(\s{u})$ and $h^{(t-1/2)}(\s{v}) = h^{(t-1/2)}(\s{u})  \Longleftrightarrow h^{(t+1/2)}(\s{v}) = h^{(t+1/2)}(\s{u})$. Notice that for a single set $\s{v}$, the information used from its neighbors are the same in both \method and \methods. Hence the equilibrium equations for set $\s{v}$ should be the same in both \method and \methods. Then they will have the same stable representations at the end.

\end{proof}

\subsubsection{Proof of Theorem 8}
\begin{customthm}{8}
 Let $G$ be a graph with $c$ connected components $C_1,...,C_c$, and $G'$ be a graph also with $c$ connected components $C'_1,...,C'_c$, then $G$ and $G'$ are isomorphic if and only if $\exists p: [c]\rightarrow [c]$, s.t.  $\forall i\in [c]$, $C_i$ and $C'_{p(i)}$ are isomorphic. 
\end{customthm}

\begin{proof}
\textbf{Right $\Longrightarrow$ Left}. Let $h_i:V(C_i)\rightarrow V(C'_{p(i)}) $ be one isomorphism from $C_i$ to $C'_{p(i)}$ for $i\in [c]$.
Then we can create a new mapping $h:V(G)\rightarrow V(G') $, such that for any $x \in V(C)$, it first locates which component $x$ inside, for example $i$, then apply $h_i$ to $x$. Clearly $h$ is a isomorphism between $G$ and $G'$, hence $G$ and $G'$ are isomorphic. 

\textbf{Left $\Longrightarrow$ Right}. 
$G$ and $G'$ are isomorphic, then there exists an isomorphism $h$ from $G$ to $G'$. Now for a specific component $C_i$ in $G$, $h$ maps $V(C_i)$ to a nodes set inside $V(G')$, and name it as $B_i$. For $i\not = j$, $B_i \cap B_j = \emptyset$ and there is not edges $e\in E(G')$ between $B_i$ and $B_j$ otherwise there must be an corresponding edge (apply $h^{-1}$) between $V(C_i)$ and $V(C_j)$ which is impossible. Hence  $\{G'[B_i]\}_{i=1}^{c}$ are disconnected subgraphs. As $h$ also preserves connections, for any $i$, $C_i$ and $G'[B_i]$ are isomorphic. Hence we proved the right side. 

\end{proof}

\subsubsection{Conjecture that k-MWL is equvialent to k-WL}\label{apdx:conjecture}

\textbf{Proof of $\forall t, \bm{wl}_k^{(t)}(G, \ms{v}) = \bm{wl}_k^{(t)}(G', \ms{v}') \Longleftarrow \bm{mwl}_k^{(t)}(G, \ms{v}) = \bm{mwl}_k^{(t)}(G', \ms{v}')$}

\begin{proof}
% \lz{I failed on proving this. There are still many problems inside. }

We first take a closer 
look at the condition of $\bm{mwl}_k^{(t+1)}(G, \ms{v}) = \bm{mwl}_k^{(t+1)}(G', \ms{v}')$. By Eq.\eqref{eq:kmwl} we know this is equivalent to \\
(1) $\bm{mwl}_k^{(t)}(G, \ms{v}) = \bm{mwl}_k^{(t)}(G', \ms{v}')$ and (2) $\Bmms{\mms{\bm{mwl}_k^{(t)}(G, \ms{v}[x/o^{-1}_G(\ms{v},i)]) \big| x \in V(G)} \ | \ i=1,...,k}$ = $\Bmms{\mms{\bm{mwl}_k^{(t)}(G', \ms{v}'[x/o^{-1}_G(\ms{v}',i)]) \big| x \in V(G')} \ | \ i=1,...,k}$. (2) can be rewrite as $\Bmms{\mms{\bm{mwl}_k^{(t)}(G, \ms{v}[x/\text{idx}_{\ms{v}}(y)]) \big| x \in V(G)} \ | \ y\in \ms{v} }$ = $\Bmms{\mms{\bm{mwl}_k^{(t)}(G', \ms{v}'[x/\text{idx}_{\ms{v}'}(y)]) \big| x \in V(G')} \ | \ y\in \ms{v}'}$, which is equivalent to (3) $\exists \text{ bijective } f:V(G)   \rightarrow V(G')$, $\forall y \in \ms{v}$, $\mms{\bm{mwl}_k^{(t)}(G, \ms{v}[x/\text{idx}_{\ms{v}}(y)]) \big| x \in V(G)} = \mms{\bm{mwl}_k^{(t)}(G', \ms{v}'[x/\text{idx}_{\ms{v}'}(f(y))]) \big| x \in V(G')} $. 

Define $F^{(t+1)}(G,G',\ms{v}, \ms{v}'):= \{\text{ injective } f:\ms{v} \rightarrow \ms{v}' \ | \ f\in F^{(t)}(G,G',\ms{v}, \ms{v}'), \text{ AND }, \forall y \in \ms{v}, \exists h_y:V(G) \rightarrow V(G'), \forall x, f \in F^{(t)}(G, G', \ms{v}[x / \text{idx}_{\ms{v}}(y)], \ms{v}'[h_y(x) / \text{idx}_{\ms{v}'} (f(y)) ] )   \}$. 
% \lz{ Let $H^{(t)}(G,G', \ms{v}, \ms{v}' ):= \{p: \ms{v}\rightarrow \ms{v}' \ |\  \bm{wl}_k^{(t)}(G, \tps{v}) = \bm{wl}_k^{(t)}(G', p(\tps{v}) )  \}$, 
% then can we prove that $H^{(t)}(G,G', \ms{v}, \ms{v}' ) = F^{(t)}(G,G', \ms{v}, \ms{v}' )$ ?}
Let $F^{(0)}(G,G',\ms{v}, \ms{v}'):= \{f \ | \ f \text{ is an isomorphism from } G[\ms{v}] \text{ to }  G'[\ms{v}']   \} $.

\begin{lemma}\label{lm:1}
 $\forall t$, $\bm{mwl}_k^{(t)}(G, \ms{v}) = \bm{mwl}_k^{(t)}(G', \ms{v}')$ if and only if $F^{(t)}(G,G',\ms{v}, \ms{v}') \not = \emptyset $.\\
 
 Notice that this Lemma is conjectured to be true. This needs additional proof. 
 
% Some properties:\\
%  (1) $F^{(t+1)}(G,G', \ms{v}, \ms{v}' ) \subseteq F^{(t)}(G,G', \ms{v}, \ms{v}' ) $ \\
%  (2) $\forall f \in F^{(t+1)}(G,G', \ms{v}, \ms{v}') $, $\forall y\in \ms{v}$, $f\in F^{(t)}(G,G', \ms{v}[/y], \ms{v}'[/f(y)])$
%  \lz{Mark here, this lemma need to be proved. This is very important lemma and not easy to prove.}
\end{lemma}

% \textit{Proof of Lemma.\ref{lm:1}:} From right to left is obvious, given that right equation implies (1) and (3). To prove left to right, given $\bm{mwl}_k^{(t+1)}(G, \ms{v}) = \bm{mwl}_k^{(t+1)}(G', \ms{v}')$, now 
% we assume $F^{(t+1)}(G,G',\ms{v}, \ms{v})  = \emptyset $. Then for any $f$ satisfies (3), $\bm{mwl}_k^{(t)}(G, \ms{v}) \not = \bm{mwl}_k^{(t)}(G', f(\ms{v}))$ \lz{not complete}.

% \begin{align}
%     \bm{mwl}_k^{(t+1)} (G, \ms{v}) = \text{HASH} \bigg( 
%     \bm{mwl}_k^{(t)}(G, \ms{v}),   \bigg\{\mskip-10mu \bigg\{  \label{eq:kmwl}
%     & \mms{\bm{mwl}_k^{(t)}(G, \ms{v}[x/o^{-1}_G(\ms{v},1)]) \big| x \in V(G)}, \\..., \nonumber
%     & \mms{\bm{mwl}_k^{(t)}(G, \ms{v}[x/o^{-1}_G(\ms{v},k)]) \big| x \in V(G)}
%     \bigg\}\mskip-10mu \bigg\}
%     \bigg)   
% \end{align}

% Notice that $\bm{mwl}_k^{(t)}(G, \ms{v}) = \bm{mwl}_k^{(t)}(G', \ms{v}')$ corresponds to a subset of isomorphisms $f:V(G[\ms{v}]) \rightarrow V(G'[\ms{v}]')  $ between $G[\ms{v}]$ and $G'[\ms{v}']$ such that $\bm{mwl}_k^{(t)}(G, \ms{v}) = \bm{mwl}_k^{(t)}(G', f(\ms{v}))$ (where $f(\ms{v}) = \mms{f(\ms{v}_1),...,f(\ms{v}_k)}$)

\begin{lemma}\label{lm:2}
 $\forall t,\bm{mwl}_k^{(t)}(G, \ms{v}) = \bm{mwl}_k^{(t)}(G', \ms{v}') \Longrightarrow  \forall f \in F^{(t)}(G,G',\ms{v}, \ms{v}'), \bm{wl}_k^{(t)}(G, \tps{v}) = \bm{wl}_k^{(t)}(G, f(\tps{v})) $
\end{lemma}
\textit{Proof of Lemma.\ref{lm:2}:} We prove it by induction on $t$. When $t=0$, $F^{(0)}(G,G',\ms{v}, \ms{v}')$ contains all isomorphisms between $G[\ms{v}]$ and $G'[\ms{v}']$, hence the right side is correct. Assume the statement is correct for $\leq t$. For  $t+1$ case, the left side implies (1) $\bm{mwl}_k^{(t)}(G, \ms{v}) = \bm{mwl}_k^{(t)}(G', \ms{v}')$ and (2)$F^{(t+1)}(G,G',\ms{v}, \ms{v}')\not = \emptyset$, $\forall  f \in F^{(t+1)}(G,G',\ms{v}, \ms{v}') $, $\forall y \in \ms{v}$, $\exists h_y:V(G) \rightarrow V(G')$, $\forall x \in V(G)$, 
$\bm{mwl}_k^{(t)}(G, \ms{v}[x/\text{idx}_{\ms{v}}(y)]) = \bm{mwl}_k^{(t)}(G', \ms{v}'[h_y(x)/\text{idx}_{\ms{v}'}(f(y))]) $. By induction hypothesis, (1) and $F^{(t+1)}(G,G',\ms{v}, \ms{v}') \subseteq F^{(t)}(G,G',\ms{v}, \ms{v}')$ imply (a)  $\forall f\in F^{(t+1)}(G,G',\ms{v}, \ms{v}'), \bm{wl}_k^{(t)}(G, \tps{v}) = \bm{wl}_k^{(t)}(G', f(\tps{v})) $; (2) and $\forall y \in \ms{v}$, $f \in F^{(t)}(G,G',\ms{v}[x/\text{idx}_{\ms{v}}(y)], \ms{v}'[h_y(x)/\text{idx}_{\ms{v}'}(f(y))])$ imply (b) $\exists h$, $\forall x \in V(G)$, $\bm{wl}_k^{(t)}(G, \tps{v}[x/\text{idx}_{\tps{v}}(y)] ) = \bm{wl}_k^{(t)}(G', f(\tps{v})[h(x)/\text{idx}_{f(\tps{v})}(f(y))] )$, which can be rewritten as $\forall y\in \ms{v}$, $\mms{\bm{wl}_k^{(t)}(G, \tps{v}[x/\text{idx}_{\tps{v}}(y)] \ | \ x\in V(G))}$ = $\mms{\bm{wl}_k^{(t)}(G', f(\tps{v})[x/\text{idx}_{f(\tps{v})}(f(y))] \ | \ x\in V(G))}$. Now applying Eq.\eqref{eq:kwl} with (a) and (b), we can derive that $\forall f \in F^{(t+1)}(G,G',\ms{v}, \ms{v}')$, $\bm{wl}_k^{(t+1)}(G, \tps{v}) = \bm{wl}_k^{(t+1)}(G, f(\tps{v}))$. Hence the statement is correct for any $t$. 

Using Lemma.\ref{lm:2} and the conclusion ($\forall p \in \text{perm[k]}$, $\bm{wl}_k^{(t+1)}(G, p(\tps{v})) = \bm{wl}_k^{(t+1)}(G', p(f(\tps{v})))$) proved in the first part of the proof of Theorem.\ref{thm:kwl=kmwl}, we proved $\forall t,  \bm{mwl}_k^{(t)}(G, \ms{v}) = \bm{mwl}_k^{(t)}(G', \ms{v}') \Longrightarrow \bm{wl}_k^{(t)}(G, \ms{v}) = \bm{wl}_k^{(t)}(G', \ms{v}') $. 
\end{proof}

\subsection{Algorithm of extracting $(k,c)(\leq)$ sets and constructing the super-graph of \kcswl}\label{apdx:construct_supergraph}

Here we present the algorithm of constructing supernodes $(t+1,c)(\leq)$ sets from supernodes $(t,c)(\leq)$ sets, and constructing the bipartite graph between $(t,c)(\leq)$ sets and  $(t+1,c)(\leq)$ sets. Notice the algorithm we presented is just the pseudo code and we use for-loops to help presentation. The algorithm can easily be parallelized (removing for-loops) with using matrix operations.  

\begin{algorithm}[h]
\caption{Constructing supernodes $(t+1,c)(\leq)$ sets and the bipartite graph between $t$ and $t+1$ }\label{alg:kcset}
\begin{algorithmic}
\Require Input graph $G$, the list $O_{t, \leq c}$ containing all supernodes with $t$ nodes and $\leq c$ number of components, the list $N_{t,\leq c}$ with $N_{t,\leq c}[i]$ be the number of components of $G[O_{t, \leq c}[i]]$. 
\Ensure $O_{t+1, \leq c}$, $N_{t+1, \leq c}$, $B_{t}$ containing edges between $O_{t, \leq c}$ and $O_{t+1, \leq c}$
\State $O_{t+1, \leq c}, N_{t+1, \leq c}, B_{t} \gets [], [], []$ 
\ForAll{ $\s{s}, n$ in zip( $O_{t, \leq c}, N_{t,\leq c}$ )}
    \State $\mathcal{N}_1 \gets$ $1$-st hop neighbors of $G[\s{s}]$
    \State $\mathcal{N}_{>1} \gets$ $ (V(G)\setminus \s{s}) \setminus \mathcal{N}_1$
    \ForAll {$m$ in $\mathcal{N}_1$} \Comment{Number of components doesn't change.}
        \State $\s{q} \gets \s{s} + \{m\}$
        \State Append $\s{q}$ into $O_{t+1, \leq c}$, $n$ into  $N_{t+1, \leq c}$
        \State Append edge $(\s{s}, \s{q} )$ into $B_t$  \Comment{{\color{blue} We change the edge to $(\s{s}, \s{q}, m)$ for Algorithm.\ref{alg:components_graph}} }
    \EndFor
    \If{n < c} \Comment{Creating an additional component.}
        \ForAll {$m$ in $\mathcal{N}_{>1}$}
            \State $\s{q} \gets \s{s} + \{m\}$
            \State Append $\s{q}$ into $O_{t+1, \leq c}$, $n+1$ into  $N_{t+1, \leq c}$
            \State Append edge $(\s{s}, \s{q} )$ into $B_t$ \Comment{ {\color{blue} We change the edge to $(\s{s}, \s{q}, m)$ for Algorithm.\ref{alg:components_graph}}}
        \EndFor
    \EndIf
\EndFor
\State Remove repeated elements inside $O_{t+1, \leq c}$, with corresponding mask $M$
\State Apply mask $M$ to  $N_{t+1, \leq c}$
\State \Return $O_{t+1, \leq c}$, $N_{t+1, \leq c}$, $B_{t}$ 

\end{algorithmic}
\end{algorithm}
To get the full supergraph $S_{k,c\text{-swl}}$, we need to apply Algorithm.\ref{alg:kcset} $k$$-$$1$ times sequentially. 

\subsection{Algorithm of connecting supernodes to their connected components}\label{apdx:inference_graph}

In Sec.\ref{ssec:supernode_init} we proved that for a supernode with multi-component induced subgraph (call it multi-component supernode), the color initialization can be greatly sped up given knowing its each component's corresponding supernode. Hence we need to build the bipartite graph (call it components graph) between single-component supernodes and multi-component supernodes. We present the algorithm of constructing these kind of connections sequentially. That is, given knowing the connections between single-component supernodes and all multi-component supernodes with $\leq t$ number of nodes, to build the connections between single-component supernodes and all multi-component supernodes with $\leq t$$+$$1$ number of nodes. To build the full bipartite graph for $(k,c)(\leq)$ sets, we need to conduct the algorithm $k$$-$$1$ times sequentially. 

\begin{algorithm}[h!]
\caption{Constructing the bipartite graph between all single-component supernodes and all multi-component supernodes with $\leq t$$+$$1$ number of nodes)}\label{alg:components_graph}
\begin{algorithmic}
\Require $\{B_i\}_{i=1}^t$, $\{O_{i, \leq c}\}_{i=1}^{t+1}$, $ \{ N_{i, \leq c} \}_{i=1}^{t+1} $ from Algorithm.\ref{alg:kcset} (with blue lines applied), dictionary $D_{\leq t}$ with each key being a multi-component supernode with $\leq t$ nodes, and value being a list of its corresponding single-component supernodes. 
\Ensure $D_{\leq t+1}$
\State $D_{\leq t+1} \gets  D_{\leq t} $
\ForAll{$\s{q}$ in $O_{t+1, \leq c}$}
   \State Get all edges $T=\{(\s{s}_i, \s{q}), m_i\}$ in $B_t$ with $\s{q}$ inside, and let $j \gets \arg \max_{i} N_{t,\leq c}[\s{s}_i]$ 
   \If{ $N_{t,\leq c}[\s{s}_j] ==N_{t+1,\leq c}[\s{q}] $} \Comment{No increasing of components}
        \State $D_{\leq t+1}[\s{q}] \gets []$
            \ForAll{$\s{p}$ in $D_{\leq t}[\s{s}_j]$}
                \State Assume $k=|\s{p}|$, i.e. the number of nodes inside.
                \State Traverse $B_{k}$ to find the supernode $(\s{p},m_j)$
                \If{Found it and $N_{k+1,\leq c}[(\s{p},m_j)] == 1 $}
                    \State Append $(\s{p},m_j)$ to $D_{\leq t+1}[\s{q}]$
                \Else
                    \State Append $\s{p}$ to $D_{\leq t+1}[\s{q}]$
                \EndIf
            \EndFor
   \Else \Comment{$N_{t,\leq c}[\s{s}_j] = N_{t+1,\leq c}[\s{q}] -1 $}
        \State $D_{\leq t+1}[\s{q}] \gets D_{\leq t}[\s{s}_j] $ + $[m_j]$
   \EndIf

\EndFor

\State \Return $D_{\leq t+1} $
\end{algorithmic}
\end{algorithm}

For the clear of presentation we use for-loops inside Algorithm.\ref{alg:components_graph}, while in practice we use matrix operations to eliminate for-loops.

\subsection{Dataset Details}
\label{ssec:data}
\begin{table}[H]
% \revise{
% \footnotesize 
\fontsize{8.7}{9.2}\selectfont
 \setlength{\tabcolsep}{1pt}
    \caption{Dataset statistics.}\label{tab:data-semantics}
    \centering
    \begin{tabular}{llcccc}
    \toprule
    Dataset & Task & \# Cls./Tasks & \# Graphs & Avg. \# Nodes & Avg. \# Edges  \\
    \midrule  
    EXP \cite{abboud2020surprising}  & Distinguish 1-WL failed graphs & 2  & 1200 & 44.4 & 110.2\\
    SR25 \cite{balcilar2021breaking} & Distinguish 3-WL failed graphs & 15 & 15   & 25   & 300\\
    \midrule
    CountingSub. \cite{chen2020can} & Regress num. of substructures   & 4& 1500 / 1000 / 2500& 18.8 & 62.6\\
    GraphProp. \cite{pna}   & Regress global graph properties & 3& 5120 / 640 / 1280 & 19.5 & 101.1 \\
    \midrule
    ZINC-12K \cite{dwivedi2020benchmarking}  & Regress molecular property    & 1  & 10000 / 1000 / 1000& 23.1 & 49.8\\
    QM9 \cite{wu2018moleculenet}   & Regress molecular properties       &  19\footnotemark & 130831  & 18.0 & 37.3 \\
    \bottomrule
    \end{tabular}
% }
\end{table}

\footnotetext{We use the version of the dataset from PyG \cite{Fey/Lenssen/2019}, and it contains 19 tasks. }

\subsection{Experimental Setup}
\label{apdx:experimental-setup}

Due to limited time and resource, we highly restrict the hyperparameters and fix most of hyperparameters the same across all models and baselines to ensure a fair comparison. This means the performance of \methods reported in the paper is not the best performance given that we didn't tune much hyperparameters. Nevertheless the performance still reflects the theory and designs proposed in the paper, and we postpone studying the SOTA performance of \methods to future work. To be clear, we fix batch size to 128, the hidden size to 128, the number of layers of Base GNN to 4, and the number of layers of \methods (the number of iterations of \kcswl) to be 2 (we will do ablation study over it later). This hyperparameters configuration is used for all datasets. We have run the baseline GINE over many datasets, and we tune the number of layers from [4,6] and keep other hyperparameters the same as \methods. For all other baselines, we took the performance reported in \cite{zhao2021stars}. 

For all datasets except QM9, we follow the same configuration used in \cite{zhao2021stars}. For QM9, we use the dataset from PyG and conduct regression over all 19 targets simultaneously. To balance the scale of each target, we preprocess the dataset by standardizing every target to a Gaussian distribution with mean 0 and standard derivation 1. The dataset is randomly split with ratio 80\%/10\%/10\% to train/validation/test sets (with a fixed random state so that all runs and models use the same split). For every graph in QM9, it contains 3d positional coordinates for all nodes, and we use them to augment edge features by using the absolute difference between the coordinates of two nodes on an edge for all models. Notice that our goal is not to achieve SOTA performance on QM9 but mainly to verify our theory and effectiveness of designs. 

We use Batch Normalization and ReLU activation in all models. We use Adam optimizer with learning rate 0.001 in all experiments for optimization. We repeat all experiments three times (for random initialization) to calculate mean and standard derivation. All experiments are conducted on V100 and RTX-A6000 GPUs.

\clearpage

\subsection{Graph Property Dataset Results}
\label{apdx:graph-property}
\begin{table}[h!]
\caption{Train and Test performances of \methods on regressing graph properties by varying $k$ and $c$. }
    \label{tab:graph_property}
    {\scalebox{0.75}{
    \begin{tabular}{rr|llllll}
    \toprule
\multicolumn{1}{l}{}  & \multicolumn{1}{l}{}  & \multicolumn{6}{c}{Regressing Graph Properties ($\log_{10}$(MSE))}                                                                                                                                                                          \\\midrule
\multicolumn{1}{c}{}  & \multicolumn{1}{c}{}  & \multicolumn{2}{c}{Is Connected}                         & \multicolumn{2}{c}{Diameter}                      & \multicolumn{2}{c}{Radius}                                                  \\
\midrule
\multicolumn{1}{l}{k} & \multicolumn{1}{l}{c} & \multicolumn{1}{c}{Train} & \multicolumn{1}{c}{Test} & \multicolumn{1}{c}{Train} & \multicolumn{1}{c}{Test} & \multicolumn{1}{c}{Train} & \multicolumn{1}{c}{Test} \\
\midrule
2                     & 1                     & -4.2266 ± 0.1222           & -2.9577 ± 0.1295          & -4.0347 ± 0.0468           & -3.6322 ± 0.0458          & -4.4690 ± 0.0348           & -4.9436 ± 0.0277                \\
3                     & 1                     & -4.2360 ± 0.1854           & -3.4631 ± 0.6392          & -4.0228 ± 0.1256           & -3.7885 ± 0.0589          & -4.4762 ± 0.1176           & -5.0245 ± 0.0881                   \\
4                     & 1                     & -4.7776 ± 0.0386           & -4.9941 ± 0.0913          & -4.1396 ± 0.0442           & -4.0122 ± 0.0071          & -4.2837 ± 0.5880           & -4.1528 ± 0.9383                 \\
2                     & 2                     & -4.6623 ± 0.3170           & -4.7848 ± 0.3150          & -4.0802 ± 0.1654           & -3.8962 ± 0.0124          & -4.5362 ± 0.2012           & -5.1603 ± 0.0610                   \\
3                     & 2                     & -4.2601 ± 0.3192           & -4.4547 ± 1.1715          & -4.3235 ± 0.3050           & -3.9905 ± 0.0799          & -4.6766 ± 0.1797           & -4.9836 ± 0.0658                  \\
4                     & 2                     & -4.8489 ± 0.1354           & -5.4667 ± 0.2125          & -4.5033 ± 0.1610           & -3.9495 ± 0.3202          & -4.4130 ± 0.2686           & -4.1432 ± 0.4405                   \\
\bottomrule
\end{tabular}
}}
\end{table}

\subsection{Ablation Study over Number of Layers}
\label{apdx:ablation}
\begin{table}[h!]
\caption{Ablation study of \methods over number of bidirectional message passing layers (L) on ZINC dataset.}
    \label{tab:ablation}
    {\scalebox{0.75}{
    \begin{tabular}{rr|llllll}
    \toprule
\multicolumn{2}{c|}{L}  & \multicolumn{2}{c}{2}                         & \multicolumn{2}{c}{4}                      & \multicolumn{2}{c}{6}                                                  \\
\midrule
k & c & \multicolumn{1}{c}{Train} & \multicolumn{1}{c}{Test} & \multicolumn{1}{c}{Train} & \multicolumn{1}{c}{Test} & \multicolumn{1}{c}{Train} & \multicolumn{1}{c}{Test} \\
\midrule
2                     & 1                     & 0.1381 ± 0.0240           & 0.2345 ± 0.0131          & 0.1135 ± 0.0418           & 0.1921 ± 0.0133          & 0.0712 ± 0.0015           & 0.1729 ± 0.0134                \\
3                     & 1                     & 0.1172 ± 0.0063           & 0.2252 ± 0.0030          & 0.0792 ± 0.0190           & 0.1657 ± 0.0035          & 0.0692 ± 0.0118           & 0.1679 ± 0.0061                   \\
4                     & 1                     & 0.0693 ± 0.0111           & 0.1636 ± 0.0052          & 0.0700 ± 0.0085           & 0.1566 ± 0.0101          & 0.0768 ± 0.0116           & 0.1572 ± 0.0051                 \\
\bottomrule
\end{tabular}
}}
\end{table}
The table shows that increasing the number of bidirectional message passing ($t$ in Eq.\eqref{eq:bmp1} and  Eq.\eqref{eq:bmp2} ) always increase the performance, which is aligning with the fact that increasing number of layers always increases expressivity.

\subsection{Computational Footprint on QM9}
\begin{figure}[h!]
	\vspace{-0.1in}
	\centering
	\begin{subfigure}{0.5\textwidth}
	    \centering
	    \includegraphics[width=6cm]{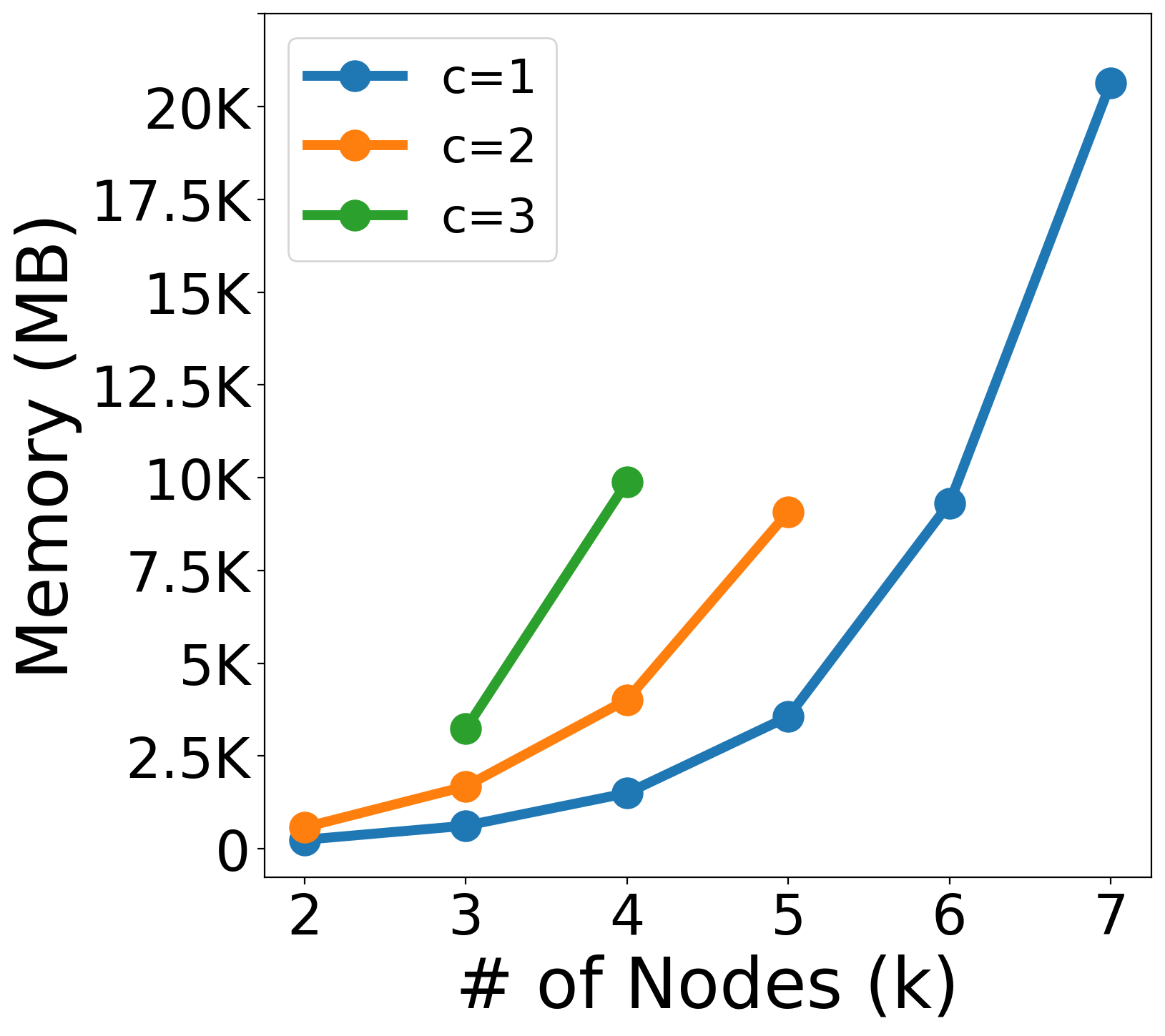}
	    \caption{Memory usage}
	    \label{fig:qm9_memory}
	\end{subfigure}%
	\begin{subfigure}{0.5\textwidth}
	    \centering
	    \includegraphics[width=5.85cm]{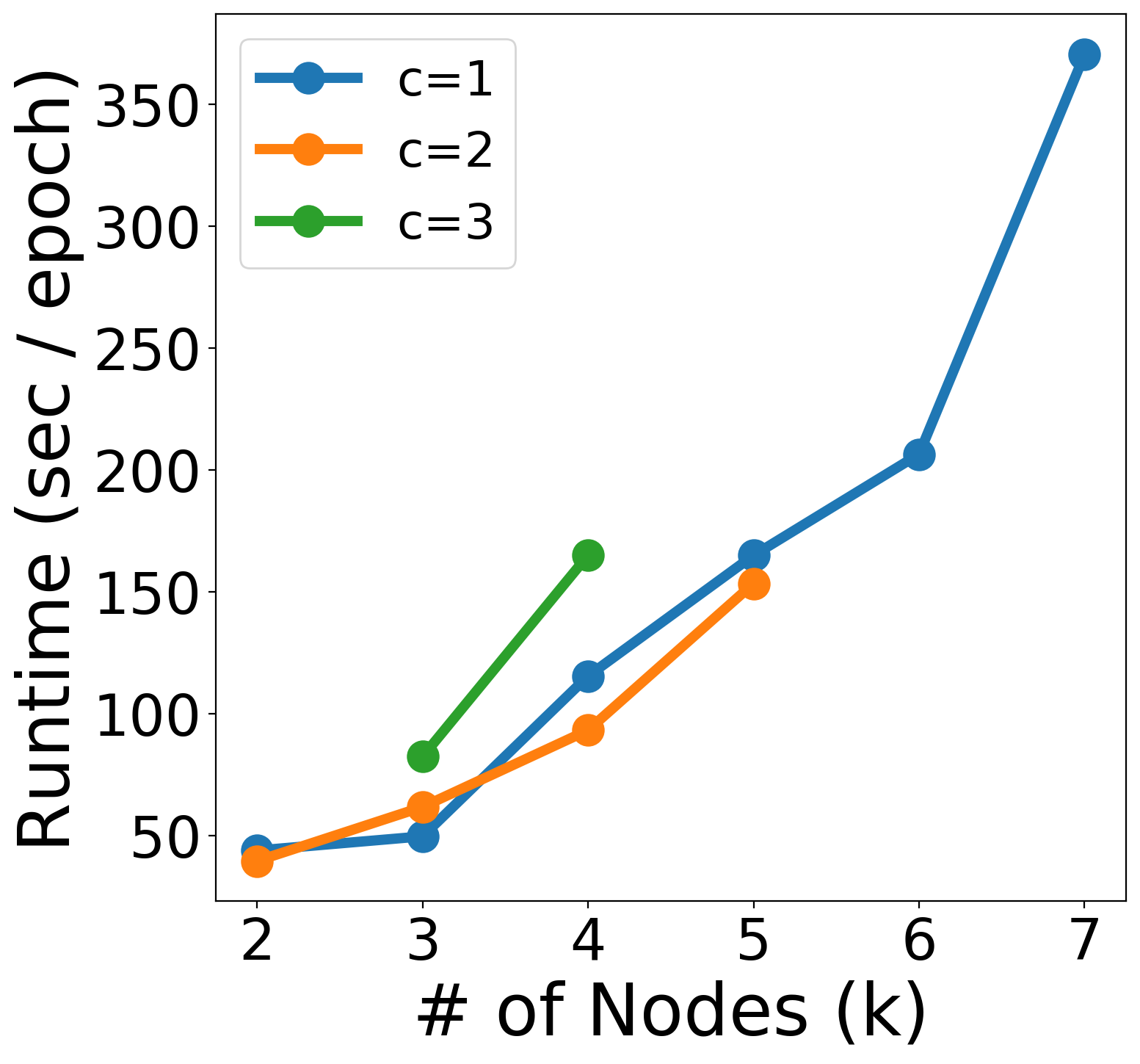}
	    \caption{Runtime}
	    \label{fig:qm9_runtime}
	\end{subfigure}
	\caption{\method's computational footprint scales with both $k$ and $c$ in terms of memory (a) and runtime (b).  Solid blue, orange and green lines track scaling as $k$ increases, when running \method on the QM9 dataset with $c=1$, 2 and 3 respectively.}
	\label{fig:qm9_scaling}
\end{figure}

\clearpage
\subsection{Discussion of k-Bipartite Message Passing}
\label{apdx:k_bipartite_discussion}

\begin{figure}[h!]
    \centering
    \includegraphics[width=\textwidth]{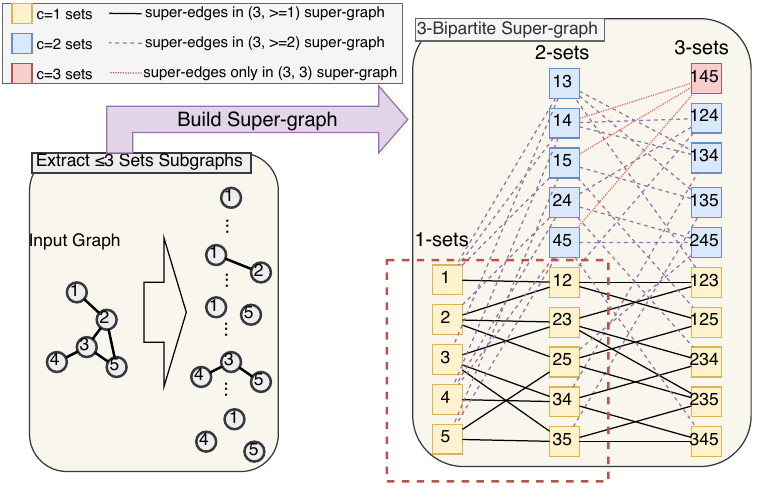}
    \caption{k-bipartite message passing recovers many well-known GNNs: message passing based GNNs \cite{gilmer2017neural}, edge-enhanced GNNs \cite{battaglia2018relational}, and line-graph GNNs \cite{chen2018supervised,choudhary2021atomistic}. The line graph is marked with red dash frame. }
    \label{fig:line_graph}
\end{figure}

Interestingly, the k-bipartite bidirectional message passing is very general and its modification covers many well known GNNs. Considering only using 1-sets and 2-sets with single connected components (the red frame inside Figure \ref{fig:line_graph}), and initializing their embeddings with their original nodes and edges representations, then it covers the follows
\begin{itemize} 
    \item Message passing based GNNs \cite{gilmer2017neural}. By using sequential message passing defined in Eq.\eqref{eq:bmp1} and Eq.\eqref{eq:bmp2}, and performing forward step first then backward step for all 1-sets. 
    \item Line graph based GNNs \cite{chen2018supervised, choudhary2021atomistic}. By using sequential message passing defined in Eq.\eqref{eq:bmp1} and Eq.\eqref{eq:bmp2}, and performing backward step first then forward step for all 2-sets. 
    \item Relational Graph Networks \cite{battaglia2018relational} or edge-enhanced GNN \citep{choudhary2021atomistic}. By using performing  bidirectional sequential message passing on all 1-sets and 2-sets. 
\end{itemize}

% \clearpage
% \section*{Checklist}
% \input{99-Checklist}

\end{document}